\documentclass{article}

\usepackage{multirow}
\usepackage{amsmath}
\usepackage{amssymb}
\usepackage{amsthm}
\usepackage{thm-restate}
\usepackage[mathscr]{euscript}
\usepackage{graphicx}
\usepackage{url}
\usepackage{bbm}
\usepackage[normalem]{ulem}
\useunder{\uline}{\ul}{}
\usepackage{makecell}
\usepackage{arydshln}
\usepackage{blkarray}
\usepackage{subfiles}
\usepackage{enumerate}
\usepackage{natbib}
\usepackage[section]{placeins}
\usepackage{graphicx}
\usepackage[hidelinks]{hyperref}

\usepackage{appendix}

\usepackage{tikz}
\usetikzlibrary{decorations.pathreplacing}
\newcommand{\tikzmark}[1]{\tikz[overlay,remember picture] \node[baseline] (#1) {};}
\tikzset{My Node Style/.style={midway, right, xshift=3.0ex, align=left, font=\small, draw=none, thin, text=black}}
\newcommand\VerticalBrace[4][]{%
\begin{tikzpicture}[overlay,remember picture]
  \draw[decorate,decoration={brace, amplitude=1.5ex, mirror}, #1] 
    ([yshift=1ex]#2.north east)  -- ([yshift=-1ex]#3.south east)
        node[My Node Style] {#4};
\end{tikzpicture}
}

\usepackage{fancyhdr}
\declaretheoremstyle[
  bodyfont=\normalfont,
  spaceabove=1em plus 0.75em minus 0.25em,
  spacebelow=2em plus 0.75em minus 0.25em,
  qed={$\triangle$},
]{exmpstyle2}
\newtheorem{theorem}{Theorem}
\newtheorem{corollary}{Corollary}
\newtheorem{lemma}{Lemma}
\newtheorem{remark}{Remark}
\theoremstyle{definition}\newtheorem{definition}{Definition}
\declaretheorem[
  style=exmpstyle2,
  title=Example,
  refname={example,examples},
  Refname={Example,Examples}
]{example}
\def\indep{\perp \!\!\! \perp}

\graphicspath{ {./images/} }

\newcommand{\myvec}[1]{\underline{\boldsymbol{{#1}}}}
\newcommand{\mymatrix}[1]{\boldsymbol{#1}}
\newcommand{\smeq}{\mkern1.5mu{=}}
\newcommand{\subheader}[1]{\vspace{0.15cm}\noindent\textit{#1}.}
\newcommand{\stress}[1]{\textit{#1}}
\renewcommand{\quote}[1]{`#1'}
\newcommand{\mymatrixdelta}{\mymatrix{\Delta}}

\newcommand\numberthis{\addtocounter{equation}{1}\tag{\theequation}}
\newcommand{\permutmat}{\mymatrix{\mathsf{\Pi}}}

\newcommand{\blind}{1}
\def\spacingset#1{\renewcommand{\baselinestretch}{#1}\small\normalsize\everydisplay{\def\arraystretch{0.5}}} 
\allowdisplaybreaks
\spacingset{1.9}

\begin{document}

\pagenumbering{roman}

\spacingset{1}

\if1\blind
{
  \title{\bf Domain Latent Class Models}
  \author{Jesse Bowers
  \thanks{We would like to thank Theren Williams, Eric Wayman, and Dr. Kristen Lee for feedback on writing style.
  }
  \hspace{.2cm}\\
    Department of Statistics, University of Illinois Urbana Champaign\\
    and \\
    Steve Culpepper \\
    Department of Statistics, University of Illinois Urbana Champaign}
  \maketitle
} \fi

\if0\blind
{
  \bigskip
  \bigskip
  \bigskip
  \begin{center}
    {\LARGE\bf Domain Latent Class Models}
\end{center}
  \medskip
} \fi

\bigskip
\begin{abstract}
Latent Class Models (LCMs) are used to cluster multivariate categorical data (e.g. group participants based on survey responses).  Traditional LCMs assume a property called conditional independence. This assumption can be restrictive, leading to model misspecification and overparameterization. To combat this problem, we developed a novel Bayesian model called a Domain Latent Class Model (DLCM), which permits conditional dependence. We verify identifiability of DLCMs. We also demonstrate the effectiveness of DLCMs in both simulations and real-world applications. Compared to traditional LCMs, DLCMs are effective in applications with time series, overlapping items, and structural zeroes.
\end{abstract}

\noindent%
{\it Keywords:}  LCM, Latent Variable, Bayes, Clustering, Categorical Data Analysis

\vfill

\newpage

\spacingset{1.9}

\begingroup
\setcounter{tocdepth}{2}
\renewcommand*{\addvspace}[1]{}
\tableofcontents
\endgroup
\newpage

\pagenumbering{arabic}

\section{Introduction}

\subsection{Problem Statement}

Latent Class Modeling (LCM) is a clustering technique for multivariate categorical data. LCMs are of interest in many areas including social, behavioral, health sciences, and record linkage. A common use is to group respondents based on their responses to a multiple choice survey and to interpret each of those groups.

Traditional LCMs break the respondents into $C \in \mathbb{N}$ groups called latent classes, and assume that respondents answer each question independently, conditional on class membership. Suppose a survey contains $J$ multiple choice questions (items). Let the $i$'th person's response to item $j$ be denoted $X_{ij} \in \mathbb{Z}_{Q_{j}} := \{0,1,\cdots,Q_{j}-1\}$. Let $\rho_{cjx_j}=P(X_{i j}= x_{j} |c_{i}=c)$ be the probability that members of class $c$ report $X_j=x_j$ for item $j$. LCMs assume that, given class membership, elements of the multivariate response vector $\myvec{X}_i=(X_{i0},\dots,X_{i,J-1})$ are conditionally independent. Consequently, if person $i$ belongs to class $c_{i}=c$ then the probability of observing $\myvec{X}_i=\myvec{x}_i$ is given by:
\begin{align}
P(\myvec{X}_{i}=\myvec{x}|c_{i}=c,\rho)  = \prod_{j=0}^{J-1}  \prod_{q=0}^{Q_{j}-1} \rho_{c j q}^{I(q=x_{j})}
\end{align}
where $I(\cdot)$ is the indicator function. Nominally the prior probability of the $i$'th subject belonging to class $c$ is $P(c_{i}=c|\myvec{\pi}) = \pi_{c}$. Therefore, the responses to our survey follow the distribution:
\begin{align}
& P(\myvec{X}_{i}=\myvec{x}|\rho,\myvec{\pi})
= \sum_{c=0}^{C-1} \pi_{c} \prod_{j=0}^{J-1} \prod_{q=0}^{Q_{j}-1} \rho_{c j q}^{I(q=x_{j})}.
\end{align}
One challenge in traditional LCMs is the assumption of conditional independence. In practice, it is sometimes inappropriate to assume that, for a member of a class, each question is answered independently. For instance, two items might overlap, asking similar questions in different ways.  Locally dependent questions also appear in time series data. Questions within the same time point may exhibit local dependence. Conversely, if the same question is asked across time points, then there may be local dependence between responses to the same question. This temporal dependence is notably present in pre-post testing with paired items.

\subsection{Contribution to Past Work}

A classical approach to address local dependence is via diagnostics and manual adjustments. One might fit a traditional LCM, check for local dependence, and tweak the model until the dependence disappears. There are a number of methods for detecting local dependence. Some classical methods include chi-squared tests and Fisher exact tests \citep{Agresti2018}. When dependence is found there are at least two techniques available to eliminate it.

The first approach is to increase the number of classes $C$ \citep{Bartholomew2011}. With more and more groups composed of smaller and smaller populations, the groups become increasingly homogeneous and local dependence decreases. In principal, with enough latent classes local dependence disappears entirely. In a later illustrative example, we show how doubling the amount of classes accounts for the local dependence caused by two dependent questions (Example~\ref{example:motivating}). In general, local dependence disappears no later than $C = \prod_{j=0}^{J-1} Q_{j}$ classes, where there is one class for every possible response pattern. The weakness of removing local dependence by increasing the number of classes is that it tends to overfit. Furthermore, a large number of classes can be hard to interpret. Considering that a main objective of LCMs is to provide a parsimonious interpretation of data, increasing the number of classes to deal with local dependence is not always attractive. Given that the correct number of classes is not known, it is also easy to choose too few classes resulting in model misspecification.

The second approach is the `Joint Variable' approach. The idea here is to transform the data itself to remove dependence. Suppose a pair of items are conditionally dependent. Those items correspond with a common \emph{domain} and could be merged into a `joint' variable \citep{Goodman1974} as demonstrated next in Example \ref{ex:jointvarex}.

\begin{example}\label{ex:jointvarex}
Consider a case with two binary items for $n=6$ respondents. If the two binary items are put into the same domain they might be recoded as follows: $(0,0)=0$, $(0,1)=1$, $(1,0)=2$, and $(1,1)=3$.
\begin{align*}
& \begin{blockarray}{cc}
\multicolumn{2}{c}{Original \quote{Items}} \\
\text{Item 1 Value} & \text{Item 2 Value} \\
\begin{block}{[cc]}
1            & 1            \\
0            & 1            \\
0            & 0            \\
1            & 1            \\
0            & 1            \\
1            & 0    \\
\end{block} \\
\end{blockarray}
\Rightarrow
\begin{blockarray}{c}
\\
\text{Recoded Joint Variable} \\
\begin{block}{[c]}
3 \\ 2 \\ 0 \\ 3 \\ 2 \\ 1 \\
\end{block} \\
\end{blockarray}
\end{align*}
\end{example}

When this technique is applied, paired items are removed and replaced with the joint variable in the dataset. This effectively removes the dependence by generating a new variable with one value for each possible response to the grouped items.

The obstacle with either aforementioned correction is that they are manual, iterative, and rely on personal judgement. The manual nature requires time and effort. The iterative process means that early decisions will be made based on a biased model and personal judgement can be difficult to reproduce. These are issues comparable to those tackled in the area of variable selection. More recent research in the area relies more on algorithms and less on human judgement.

We propose a model called a Domain Latent Class Model (DLCM). DLCM is a Bayesian model which extends the joint variable approach. It is an exploratory algorithm which identifies locally dependent items and groups dependent items together into joint variables. 

Additional work considered more sophisticated procedures for handling conditional dependence in latent class models. There are at least four general approaches for solving local dependence. The first type of models are hierarchical models. These are typically described as a tree with the latent class up top, intermediary latent variables in the middle, and the observed responses at the end. Latent Tree Models \citep{Chen2012} and Bayesian Pyramids \citep{gu2021} are two examples of this. The second type of models are mixture factor models \citep{Cagnone2012,Daggy2014}. For instance, \cite{Cagnone2012} developed a mixture of factor models to describe heterogeneity in multivariate binary response data. Another setting is from record linkage where there may be records from two different data sources with no common key, but several common items. For example, \cite{Daggy2014} fits a mixture model to classify all pairs of items into one of two classes: matching records and unmatching records. Dependence between paired items is handled with a single factor, mixture model, and a cumulative link. The third type of models are log-linear models. Here for every possible vector of item responses, we predict the number of matching observations.  These log-linear models can either be used directly \citep{Uebersax2009} or in a mixture \citep{Daggy2014}. \cite{Daggy2014} work is applied to record linkage. Cross terms are included in the log-linear model to account for dependence between known pairs of items. The fourth class of models use joint variables. \cite{Marbac2014} proposes a Conditional Modes Model (CMM). They use a Metropolis algorithm to search for conditionally dependent blocks of items and convert them to joint variables. Our Dependent Latent Class Models (DLCMs) also use a Metropolis algorithm to search for dependent items, but differs from Marbac's CMMs in several ways discussed below.

We offer several contributions to existing research. First, our method can be viewed as an exploratory version of the joint variable approach that does not require a priori knowledge (as in record linkage) or sequential analyses to formulate. Second we provide rigorous identifiability conditions. Not every author provides identifiability conditions (e.g. \cite{Chen2012}). In particular we establish more complete and general identifiability conditions versus Marbac's CMMs. Third, our model includes regularization which prefers simpler models with less assumed conditional dependence. Marbac uses a uniform prior over the distribution of joint variables and Monte Carlo simulations reported below provide evidence that our approach more accurately uncovers the joint variable dependence structure. Finally our model offers strong flexibility. Compared to mixture factor models, our joint variables do not require a latent linear relationship between items. We also, optionally, allow different classes from having different joint variables and therefore different local dependence structures. This differs from Marbac's CMMs and Daggy's mixed log-linear models which are structured to have the same dependence for different classes. Hierarchical models also offer strong flexibility, but can be difficult to interpret. Latent tree models in particular suffer from the fact that any intermediary node can be interpreted as the head of tree. This leads to many competing interpretations for the same model. We show in real world applications that our DLCMs offer clear interpretations.

The remainder of the paper is organized as follows. In section 2 we introduce the notation necessary to formalize DLCMs. Thoughout the paper we customarily use examples to illustrate definitions. In section 3, we establish sufficient conditions for generic identifiability of DLCMs. In sections 4 and 5, we discuss the full conditional distribution of the DLCM parameters and the prior for domains.  Section 6 describes the MCMC algorithm for approximating the posterior of our parameters.  In section 7 we validate the accuracy of DLCMs in simulation studies on artificial data. Section 8 showcases the power of DLCMs in real world applications. In section 9 we provide closing thoughts. We provide an R package for running the DLCM. This is available on Github\if1\blind{: \url{https://github.com/jessebowers/dependentLCM}}\fi.

\section{Domain Notation}

Within a class, we segment items into conditionally independent groups called domains. A domain is a set of items which will be combined to form one joint variable. Conditional on being in class $c_{i}=c$, items within the same domain are dependent. However, items of one domain are conditionally independent, given class, of the items of all other domains. Given that items in the same domain are transformed into a joint variable, every possible response pattern to the grouped items is given an individual probability. With a Metropolis within Gibbs Markov chain Monte Carlo (MCMC) process, our DLCM actively searches for a promising way to recode our data to capture conditional independence.

Within any class $c$, the items are partitioned into $D$ disjoint sets: $J(c,1),\cdots,J(c,D) \subseteq \mathbb{Z}_{J} := \{0,1,\cdots,J-1\}$. We allow for empty $J(c,d)$ with $D$ typically, but not necessarily, much larger than our number of items $J$. Let $J_{(k)}(c, d)$ refer to the item with the $k$'th smallest label in domain $(c, d)$, and in general let $S_{(k)}$ be the $k$'th smallest element of set $S$.

The DLCM domains are subject to certain restrictions. In the most restricted case, all classes are required to group items the same way and have the same domains: $J(c,d)=J(c',d)$ for all $c,c',d$. This is called a homogeneous DLCM. Conditional on a fixed domain structure, a homogeneous DLCM can be thought of as transforming our dataset $\mymatrix{X}$ by merging dependent items, and then applying traditional LCM onto the transformed dataset. In the least restrictive case, each class is allowed to have different domains and group items differently. This is called a heterogeneous DLCM. In this case the recoding of responses vary from class to class.

The responses for items in the same domain and class need to be modeled jointly. We call the series of responses in a domain a \quote{pattern}: $X_{i,J(c, d)}$. It is convenient to express these patterns as integer values. Let $r_{i c d}$ be the integer referring to the $i$'ths person's responses to the questions in domain $(c,d)$. The multivariate responses to a domain are transformed to an integer using a mapping vector $\myvec{V}(S)=[V_{1}(S),\cdots,V_{J}(S)]^{\top}$ which takes set $S\subseteq \mathbb{Z}_{J}$ and produces a vector in $(\mathbb{N}\cup \{0\})^{J}$. Specifically, $r_{icd}$ is defined as 
\begin{equation}
r_{i c d} :=  \myvec{V}(J(c,d))^{\top}\myvec{X}_{i}  \in \mathbb{Z}_{R_{c d}}
\end{equation}
where element $j$ of $\myvec{V}(S)$ is defined as
\begin{equation}
V_{j}(S) := \left\{\begin{array}{ll}
0 & j \notin S \\
1 & j = S_{(1)} \\
\prod_{w=1}^{m-1} Q_{S_{(w)}} & j = S_{(m)}, m>1 \\
\end{array}\right. 
\end{equation}
and
\begin{equation}
R_{c d} :=  \prod\limits_{j \in J(c,d)} Q_{j}
\end{equation}
is the total number of patterns for domain $(c,d)$. For a given class, patterns can also be expressed as a vector $\myvec{r}_{i c}$ with values $[\myvec{r}_{i c}]_{d} := r_{i c d}$.

Variable $\delta_{j c}$ allows one to look up what domain the $j$'th item belongs to in class $c$. By definition $j \in J(c, \delta_{j c})$ and $J(c,d) = \{j: \delta_{j c} = d\}$. For a given class, the $\delta_{j c}$ can be expressed as a vector: $\myvec{\delta}_{c} := [\delta_{0,c}, \cdots, \delta_{J-1,c}]^{\top}$. It can also be expressed as a $J\times C$ matrix, $\mymatrixdelta$, with elements $\mymatrixdelta_{j c} = \delta_{j c}$. Matrix $\mymatrixdelta$ completely specifies how the items are grouped into domains across all classes. We call $\mymatrixdelta$ the \emph{domain structure}.

We particularly care about nonempty domains and the patterns corresponding to these domains. Let set $\mathscr{D}_{c}:=\{d:|J(c,d)|>0\} \subseteq \mathbb{Z}_{D}$ identify the nonempty domains in class $c$.

To illustrate this notation consider the following example:

\begin{example}\label{ex:notation}
Suppose we have a dataset with  $J=6$ items and $C=2$ classes. The first three items are binary ($Q_{j}=2$), items four and five have five options ($Q_{j}=5$), and the last item is binary ($Q_{j}=2$). We show a possible domain structure $\mymatrixdelta$ in Figure~\ref{fig:exampleNotation1}a. The resulting domain items $J(c,d)$ are shown in Figure~\ref{fig:exampleNotation1}b and a visualization of the domains is shown in Figure~\ref{fig:exampleNotation1}c.

\begin{figure*}[!b]
\begin{tabular}{ccc}
$\mymatrixdelta
= \begin{blockarray}{ccc}
\myvec{\delta}_0 & \myvec{\delta}_1 & \\
\begin{block}{[cc]l}
\delta_{0,0} & \delta_{0,1} & j\smeq0 \\
\delta_{1,0} & \delta_{1,1} & 1 \\
\delta_{2,0} & \delta_{2,1} & 2 \\
\delta_{3,0} & \delta_{3,1} & 3 \\
\delta_{4,0} & \delta_{4,1} & 4 \\
\delta_{5,0} & \delta_{5,1} & 5 \\
\end{block}
\end{blockarray}
 \smeq \begin{blockarray}{cc}
 \\
\begin{block}{[cc]}
6 & 2 \\
6 & 7 \\
6 & 7 \\
4 & 2 \\
8 & 6 \\
8 & 2 \\
\end{block}
\end{blockarray}$
&
$\begin{array}{rl}
J(c,d)\\
J(0, 4) & =\{3\} \\ 
J(0, 6) & =\{0, 1, 2\} \\ 
J(0, 8) & =\{4, 5\} \\ 
J(1, 2) & =\{0, 3, 5\} \\ 
J(1, 6) & =\{4\} \\ 
J(1, 7) & =\{1, 2\} \\ 
\end{array}$
&
\begin{minipage}{.3\textwidth}
\includegraphics[height=4cm]{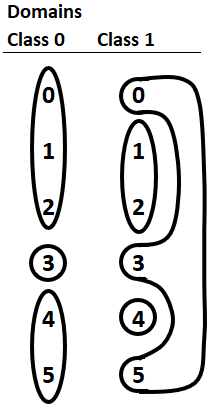}
\end{minipage}
\\
(a) & (b) & (c)
\end{tabular}
\caption{Domains in Example~\ref{ex:notation}.}
\label{fig:exampleNotation1}
\end{figure*}

In general, for domain $(c,d)=(1,2)$ with $J(1,2)=\{0,3,5\}$, there are $R_{12} = (2)(5)(2) = 20$ possible patterns with $r_{i12}\in\{0,1,\dots,19\}$. For example, for an individual who responded $\myvec{X}_{i} = [0, 0, 0, 2, 1, 1]^{\top}$ the corresponding pattern $r_{i12}$ is given by:
\begin{align*}
r_{i,c=1,d=2} & = \myvec{V}(J(c\smeq1, d\smeq2))^{\top}\myvec{X}_{i}\\
& = [0, 0, 0, 2, 1, 1] 
\left[\begin{array}{c} 1 \\0 \\ 0 \\ 2 \\ 0 \\ 2\cdot 5 \end{array}\right]
= 14.
\end{align*}

\end{example}

\subsection{Pattern Probabilities}

Patterns $r_{i c d}$ and $r_{i c d'}$ are assumed to be conditionally independent for $d\neq d'$. For subject $i$ in class $c$, a domain pattern has probability:
\begin{gather}
P(r_{i c d} = r|c_{i}=c,\theta_{c d r},\myvec{\delta}_{c}) = \theta_{c d r} \\
P(\myvec{r}_{i c}|c_{i}=c,\theta_{c},\myvec{\delta}_{c})
= \prod_{d \in \mathscr{D}_{c}}\prod_{r=0}^{R_{c d}-1} \theta_{c d r}^{I(r=r_{ic d})}
\end{gather}
An empty domain $d$ has $r_{i c d}=0$ and $\theta_{c d 0} = 1$ everywhere, and is generally omitted. All probabilities for a given class can be represented as $\theta_{c}$, and 
probabilities for a domain $(c,d)$ can also be expressed as vector $\myvec{\theta}_{c d} = [\theta_{c d 0}, \cdots \theta_{c,d,R_{c d-1}}]^{\top}$. For a given class $c$ with domains $\myvec{\delta}_{c}$ there is a one to one relationship between responses $\myvec{X}_{i}$ and patterns $\myvec{r}_{i c}$. It follows that:
\begin{gather}
P(\myvec{X}_{i}|c_{i}=c,\theta_{c},\myvec{\delta}_{c}) 
= P(\myvec{r}_{i c}|c_{i}=c,\theta_{c},\myvec{\delta}_{c}).
\end{gather}
When marginalized across class, the response probabilities are:
\begin{gather}
P(\myvec{X}_{i}|\theta,\mymatrixdelta,\myvec{\pi})
= \sum_{c=0}^{C} \pi_{c} P(\myvec{r}_{i c}|c_{i}=c,\theta_{c},\myvec{\delta}_{c})
\end{gather}
with $\pi_{c}$ representing the prior probability that subject $i$ is in class $c$. 

The following example demonstrates the pattern probabilities of a class in practice:

\addtocounter{example}{-1}
\begin{example}[continued]
Continuing the previous example, suppose class 1 had the following probabilities:
\begin{align*}
\myvec{\theta}_{1 2} 
& = \frac{1}{100}[4, 3, 5, 5, 6, 4, 4, 5, 3, 7, 5, 7, 6, 6, 9, 5, 5, 4, 3, 4]^{\top}\\
\myvec{\theta}_{1 6}  & = [0.20,0.28,0.18,0.25,0.09]^{\top}
\\
\myvec{\theta}_{1 7}  & = [0.26, 0.20, 0.25, 0.29]^{\top}
\end{align*}
The probability that subject $i$ in class 1 responds $\myvec{X}_{i} = [0, 0, 0, 2, 1, 1]^{\top}$ is given by:
\begin{align*}
& P(\myvec{X}_{i} = [0, 0, 0, 2, 1, 1]^{\top}|c_{i}=1,\theta_{1},\myvec{\delta}_{1})\\
& = P([r_{i 1 2},r_{i 1 6},r_{i 1 7}]=[14,1,0]|c_{i}=1,\theta_{1},\myvec{\delta}_{1}) \\
& = \theta_{1,2,14} \theta_{1,6,1} \theta_{1,7,0} \\
& = (0.09)(0.28)(0.26) \approx 0.0066.
\end{align*}
\end{example}
When looking at the joint distribution of $r_{i c d}$ and $r_{i c d'}$ it is useful to use Kronecker products. This allows us to consider a vector of probabilities rather than considering one pattern value at a time.

\begin{definition}
[Kronecker product] Let vectors $\myvec{Y}_{l}$ and $\myvec{Y}_{l}'$ be of size $m$ and $m'$ respectively. Their Kronecker product is given by:
\begin{align}
& \myvec{Y} \otimes \myvec{Y}'
 = \left[\begin{array}{c} Y_{1} \myvec{Y}' \\ Y_{2} \myvec{Y}' \\ \cdots \\ Y_{m} \myvec{Y}'\end{array}\right]
= \left[\begin{array}{c} Y_{1} Y'_{1} \\ \cdots  \\ Y_{1} Y'_{m'} \\ Y_{2} Y'_{1} \\ \cdots \\ Y_{2} Y'_{m'} \\ \cdots \\ \cdots \\ Y_{m} Y_{m'}'\end{array}\right]
\in \mathbb{R}^{m m'}.
\end{align}
We also define the column-wise Kronecker product, sometimes called the Khatri–Rao product:
\begin{align}
& \left[\myvec{Y}_{1},
\cdots,
\myvec{Y}_{k}
\right]
\otimes^{*} \left[\myvec{Y}_{1}',
\cdots,
\myvec{Y}_{k}'
\right]\\
&  = \left[
(\myvec{Y}_{1} \otimes \myvec{Y}_{1}'),
\cdots,
(\myvec{Y}_{k}\otimes\myvec{Y}_{k}')
\right] 
\end{align}
\end{definition}

An immediate use of the Kronecker products is to describe the joint distribution of $r_{i,c,d\in S}$ for $S \subseteq \mathbb{Z}_{D}$. There exists some permutation matrix $\permutmat$ where:
\begin{align*}
& P(r_{i,c,d\in S} | c_{i}=c, \theta, \myvec{\delta}_{c})
= P(\myvec{V}(\bigcup_{d \in S} J(c,d))^{\top}\myvec{X}_{i}| c_{i}=c, \theta, \myvec{\delta}_{c})
\\
& \left[\begin{array}{lll}
P(\myvec{V}(\bigcup_{d \in S} J(c,d))^{\top}\myvec{X}_{i} & = 0 & | c_{i}=c, \theta, \myvec{\delta}_{c}) \\
P(\myvec{V}(\bigcup_{d \in S} J(c,d))^{\top}\myvec{X}_{i} & = 1 & | c_{i}=c, \theta, \myvec{\delta}_{c}) \\
\cdots \\
P(\myvec{V}(\bigcup_{d \in S} J(c,d))^{\top}\myvec{X}_{i} & = \prod_{d \in S} R_{c d} - 1 & | c_{i}=c, \theta, \myvec{\delta}_{c})
\end{array}\right]
= \permutmat \bigotimes_{d \in S} \myvec{\theta}_{c d}
\end{align*}
That is, up to reordering, the Kronecker product $\bigotimes_{d \in S} \myvec{\theta}_{c d}$ describes the distribution of $r_{i,c,d\in S}$.

We now introduce a property called Kronecker Separability. 

\begin{definition}\label{def:kroneckerseparable}
Domain $(c,d)$ with probabilities $\myvec{\theta}_{c,d}$ is Kronecker separable if these probabilities could be formed from two groups of independent items. 

Loosely speaking, $\myvec{\theta}_{c d}$ is Kronecker separable if it can be expressed as a Kronecker product of two probability vectors. If there exist a bipartition of items $J_{0} \bigsqcup J_{1} = J(c,d)$ with probabilities $\myvec{\theta}^{(0)}$, $\myvec{\theta}^{(1)}$ where $\myvec{\theta}^{(0)} \otimes \myvec{\theta}^{(1)}$ equals $\myvec{\theta}_{c d}$ then $\myvec{\theta}_{c d}$ is Kronecker separable. When establishing equality, we force the terms of $\myvec{\theta}^{(0)} \otimes \myvec{\theta}^{(1)}$ to be reordered to match the patterns of $r_{i c d}$.
\end{definition}

\begin{example}
Suppose $J(c, d) := \{0,1\}$ with $Q_{1} = Q_{2} := 2$. If $\myvec{\theta}_{c d} = [0.25, 0.25, 0.25, 0.25]^{\top} = [0.5,0.5] \otimes [0.5,0.5]$ then $\myvec{\theta}_{c d}$ is Kronecker separable. If $\myvec{\theta}_{c d} = [0.5,0,0.5,0]^{\top}$ then both items are dependent and this is not Kronecker separable. 
\end{example}

Kronecker separability is important because if a domain is Kronecker separable then it can be split into two domains without changing the distribution of $\myvec{X}$. Conversely if a domain is not Kronecker separable then splitting the domain would change the distribution.

\section{Identifiability}\label{section:identifiability}

An important issue for mixture models is identifiability. Without identifiability there may be many choices of parameters $\omega:=(\myvec{\pi},\theta, \mymatrixdelta) \in \Theta$ which fit our responses equally well, making inference problematic. We define identifiability as follows:

\begin{definition}\qquad
\begin{enumerate}
\item Unique distribution: A specific choice of parameters $\omega:=(\myvec{\pi},\theta, \mymatrixdelta) \in \Theta$ has a unique distribution if no other choice of $\omega \in \Theta$ produces the same distribution of $\mymatrix{X}|\omega$. That is, $\omega$ has a unique distribution if for all $\omega'$ we have $\mymatrix{X}|\omega' \overset{d}{=} \mymatrix{X}|\omega$ only when $\omega' = \omega$. We consider $\omega$ and $\omega'$ to be the same when they have the same values up to relabeling of class and domain identifiers $\delta_{j c}$.
\item Strict identifiability: A model is strictly identifiable if every choice of $\omega \in \Theta$ has a unique distribution.
\item Generic Identifiability: A model is generically identifiable if only a measure zero subset of $\omega$ does not have a unique distribution. This is with respect to the standard Lebesgue measure.
\end{enumerate}
\end{definition}

Identifiability, especially strict identifiability, can fail for categorical mixture models \citep{CarreiraPerpin2000}. \citet{Allman2009} provides some useful results for determining generic identifiability for traditional LCM. These are based on algebraic results from \citet{Kruskal1977}. Our proofs are inspired by Allman, but generally work directly with Kruskal's theorem. For details see online Appendix~\ref{appendix:identifiability}.

\begin{restatable}[]{theorem}{restateidentifiability}\label{thm:identifiability}
[Identifiability] A DLCM is generically identifiable if the following conditions are met:

\begin{itemize}
\item Domain structure $\mymatrixdelta$ is restricted. For each allowed $\mymatrixdelta$ there must be a tripartition of items $J_{0},J_{1},J_{2}$ which fulfill the following. First the partitioned items must be conditionally independent for all classes: $\mymatrix{X}_{\cdot,J_{0}} \indep \mymatrix{X}_{\cdot,J_{1}} \indep \mymatrix{X}_{\cdot,J_{2}}$. Second the following inequality must hold:
\begin{align}
& min(\kappa_{0},C)+min(\kappa_{1},C)+min(\kappa_{2},C) \geq 2 C + 2 \label{eq:identifiabilityKruskal}\\
& \kappa_{k} := \prod_{j \in J_{k}} Q_{j}
\end{align}

\item Probabilities $\theta$ are restricted such that every $\theta_{c d}$ is Kronecker inseperable.

\item $\pi_{c}>0$ for all $c$

\end{itemize}
\end{restatable}

\section{DLCM Distributions}

In this section we introduce the DLCM Bayes priors and the full conditional distributions. By design these distributions have a lot in common with the traditional LCM. 

Our DLCM Bayes parameters use the following priors:
\begin{align}
c_{i}|\myvec{\pi} & \sim Cat(\myvec{\pi}) \\ 
\myvec{\pi} & \sim Dirichlet(\myvec{\alpha}^{(c)}) \\
\myvec{\theta}_{c d}|\mymatrixdelta & \sim Dirichlet(\alpha^{(\theta)} \myvec{1}_{R_{c d}})
= Dirichlet([\alpha^{(\theta)},\cdots,\alpha^{(\theta)}]^{\top})
\label{eq:thetaPrior}
\end{align}

To ensure identifiability \eqref{eq:thetaPrior} is restricted so $\myvec{\theta}_{c d}$ is not Kronecker separable, removing a measure zero space. These priors produce the following posteriors:

\begin{restatable}[]{theorem}{restateposterior}\label{thm:posterior}
The DLCM full conditional distributions for $c_i$, $\myvec{\pi}$, and $\myvec{\theta}_{cd}$ are:
\begin{align}
P(c_{i}=c|\myvec{X}_{i},\myvec{\pi},\mymatrix{\theta},\mymatrixdelta)
&= \frac{\pi_{c} p(\myvec{r}_{c,i}|c_{i}=c,\theta_{c},\myvec{\delta}_{c})}{\sum\limits_{c'=0}^{C-1} \pi_{c'} p(\myvec{r}_{c,i}|c_{i}=c',\theta_{c'},\myvec{\delta}_{c'})} 
\\
(\myvec{\pi}|\mymatrix{X},\theta, \mymatrixdelta,\myvec{c})
&=(\myvec{\pi}|\myvec{c})
\sim Dirichlet(\myvec{\alpha}^{(c)}+\myvec{n}^{(c)}) 
\\
(\myvec{\theta}_{c d}|\mymatrix{X},\myvec{\pi}, \mymatrixdelta,\myvec{c})
&=(\myvec{\theta}_{c d}|\mymatrix{r}_{c,\cdot,d},\myvec{c},\myvec{\delta}_{c})\nonumber\\
& \sim Dirichlet(\alpha^{(\theta)} \myvec{1}_{R_{c d}} + \myvec{n}^{(\theta)}_{c d}) 
\end{align}
where $\myvec{1}_{k} := [1,\cdots,1]^{\top}$ is a $k$-vector of ones and $n^{(c)}_{c} := \sum_{i=0}^{n-1} I(c_{i}=c)$ and $n^{(\theta)}_{c d r} := \sum_{i=0}^{n-1} I(r_{i c d} = r, c_{i}=c)$ are elements of $\myvec{n}^{(c)}$ and $\myvec{n}_{cd}^{(\theta)}$, respectively.

\end{restatable}

\begin{proof}
The proof of this theorem is standard, following similar lines to traditional LCM. Details can be found in online Appendix~\ref{appendix:posteriors}.
\end{proof}

These posteriors are similar to traditional LCMs. For comparison see Equations \eqref{eq:tlcm.cPosterior}, \eqref{eq:tlcm.piPosterior}, \eqref{eq:tlcm.rhoPosterior} in online Appendix~\ref{appendix:posteriors}. This has an important consequence for homogeneous DLCMs. For a homogeneous domain structure let $\mymatrix{r}$ be a $n \times D$ matrix with rows $\myvec{r}_{i, c=0}$. Conditional on this domain structure $\mymatrixdelta$, the homogeneous posteriors are equivalent to traditional LCM posteriors on $\mymatrix{r}$ in place of $\mymatrix{X}$. In other words, for fixed $\mymatrixdelta$ we transform $\mymatrix{X}$ into $\mymatrix{r}$ by coding locally dependent items into new single-item domain patterns. Then we apply a traditional LCM to this transformed dataset. Homogeneous DLCMs stochastically \quote{search} for a promising transformation of $\mymatrix{X}$ which supports good fit. Heterogeneous DLCMs further extend this idea allowing different classes to code responses differently.

\section{Domain Prior}\label{section:domainPrior}

In this section we examine the prior distribution of the domain structure $\mymatrixdelta$. We propose two types of priors: the \quote{bucket} prior, and the \quote{pattern adjusted} prior. Both priors work well in simulation studies with the bucket  prior working somewhat better for smaller samples, and the pattern adjusted prior performing somewhat better for larger samples. The support of $\mymatrixdelta$ is subject to certain restrictions. Namely $\mymatrixdelta$ is restricted by the identifiability conditions given by the Kruskal inequality \eqref{eq:identifiabilityKruskal} in Theorem~\ref{thm:identifiability}. The other restrictions are discussed in the two subsections below. For simplicity, we start with the simplest case: homogeneous DLCMs with bucket  priors.

\subsection{Homogeneous Domain with Bucket  Prior}

In homogeneous DLCMs every class has the same domains: $\myvec{\delta}_{c}=\myvec{\delta}_{c'}$ for all $c,c'\in \mathbb{Z}_{C}$. Therefore, we choose some representative class $c$ and fully explain the domain structure $\mymatrixdelta$ from $\myvec{\delta}_{c}$. In this subsection, we introduce our homogeneous domain bucket  prior and show that it encourages parsimony by means of less complex domain structures.

Within the support, every choice of $\myvec{\delta}_{c} \in \mathbb{Z}_{D}^{J}$ is has equal prior probability.  If we neglect restrictions for a moment, this prior has a certain interpretation. An individual item would be equally likely to be in any of the $D$ domains in class $c$. In this way, class $c$'s domains are analogous to $J$ numbered balls (items) distributed randomly among $D$ buckets (domains). We also have a restriction $\mathsf{MaxItems} \in \{1,\cdots,J\}$ which limits the greatest number of items which can be in any individual domain. Neither of these restrictions depend on the specific labeling of $\mymatrixdelta$, even in the heterogeneous case (see online Appendix~\ref{appendix:domainPrior}).

Although every allowable choice of $\myvec{\delta}_{c}$ is equally likely, not every partition of items $\{J(c,d): d\in\mathbb{Z}_{D}\}$ is equally likely. Consider the following example:

\begin{example}
Consider a homogeneous DLCM on $J=2$ items. Neglect identifiability restrictions for simplicity. If $D=1$, then the two items will always be in the same domain. If $D=2$ then the items are equally likely to be put together in the same domain versus put into separate domains. Suppose we have more domains: $D=20$. Then every choice of $(\delta_{0,0},\delta_{0,1}) \in \mathbb{Z}_{20}^{2}$  is equally likely, but only $20/200$ cases put both items in the same domain. Generally, the more domains $D$ there are, the less likely a pair of items will appear in the same domain.
\end{example}

When we apply restrictions we reduce the support, and then scale up all probabilities by a common normalizing constant. In general we have the following prior probabilities:

\begin{restatable}[]{theorem}{restatedomainPrior}\label{thm:domainPrior}
For a homogeneous DLCM, suppose $P(\mymatrixdelta)>0$. Up to proportionality, the bucket  prior probability of class $c$'s domains is given by:
\begin{align}
& P(\myvec{\delta}_{c}) \propto 1 \\
& P(\{J(c,d):d \in \mathbb{Z}_{D}\}) \propto \frac{D!}{(D-|\mathscr{D}_{c}|)!D^{J}}
\\
& P(\{|J(c,d)|:d \in \mathbb{Z}_{D}\})
\propto \frac{J!}{\prod_{k=0}^{D-1} |J(c,d)|!
\ \prod_{k=1}^{J} |\{d:|J(c,d)|=k\}|!}
\ 
\frac{D!}{(D-|\mathscr{D}_{c}|)!
\ D^{J}}
\end{align}
where $\mathscr{D}_{c}:=\{d:|J(c,d)|>0\}$ identifies the nonempty domains in class $c$.
\end{restatable}

\begin{proof}
Follows a standard combinatoral argument found in online Appendix~\ref{appendix:domainPrior}
\end{proof}

From a practical standpoint, we want the simplest domain structures to have the highest prior probability. In the spirit of Occam's Razor, this would bias our DLCM towards the best-fitting parsimonious models. Generally speaking, the larger $D$ the less likely a priori any two items will end up in the same domain. We formalize this idea with the following corollary.

\begin{restatable}[]{corollary}{restatedomainPriorIneq}\label{thm:domainPriorIneq}
For any $q$ and $J$, if $D$ is such that:
\begin{align}
& D \geq J + \frac{q}{2} J(J-1)-1
\label{eq:priorineq}
\end{align}
Then we have the following inequality under a bucket  prior:
\begin{align}
& P(\{|J(c,d)|:d \in \mathscr{D}_{c}\} = \{1,1,\cdots,1\})
\geq q  P(\{|J(c,d)|:d \in \mathscr{D}_{c}\} = \{2,1,1,\cdots,1\})
\label{eq:priorineq2}
\end{align}
If the terms of \eqref{eq:priorineq2} are nonzero, then exact equality in \eqref{eq:priorineq} provides exact equality \eqref{eq:priorineq2}. 
\end{restatable}

If $D$ is large enough to satisfy \eqref{eq:priorineq} with $q = 1$, then the most likely value of $\{|J(\delta,c)|:\delta \in \mathscr{D}_{c}\}$ puts each item in its own separate domain. When every item is in its own domain, we have conditional independence and DLCM is equivalent to traditional LCM - the simplest form.

By default in our distributable package we use $q = 2$ and $D = J^{2}-1$. This makes our simplest form twice as likely as the next simplest. We use these defaults in both our simulations and real world applications described in Sections \ref{section:simStudies} and \ref{section:applications}. We next employ a toy example to demonstrate how the aforementioned choice of defaults translates to the most common domain structures a priori.

\begin{example}\label{ex:domainprior}
Suppose we have $J=20$ binary items and set $D = J^{2}-1$. We fit with a homogeneous DLCM with $C=3$ classes. The most common domain structures a priori are given in Table~\ref{table:exDomainPrior}.

\begin{table}[!htbp]
\begin{tabular}{lrrr}
\multicolumn{4}{c}{Most common domain structures in Example~\ref{ex:domainprior}}\\
$S_{1}:=\{|J(c,d)|:d \in \mathscr{D}_{c}\}$ & $P(S_{1})$ & \# of $S_{2}:=\{J(c,d):d \in \mathbb{Z}_{D}\}$ & $\ln P(S_{2})$ \\\hline
1x20 & 61.6\% & 1 & -0.48 \\
2,1x18 & 30.8\% & 190 & -6.42 \\
3,1x17 & 0.5\% & 1,140 & -12.37 \\
2,2,1x16 & 6.2\% & 14,535 & -12.37 \\
3,2,1x15 & 0.2\% & 155,040 & -18.31 \\
2,2,2,1x14 & 0.6\% & 581,400 & -18.31 \\
4,1x16 & $<$0.1\% & 4,845 & -18.31 \\
$\cdots$ & $<$0.1\% & $\cdots$ & $\leq -24.26$ \\
\end{tabular}
\caption{The most common domain structures with $J=20$ binary items, $D=J^{2}-1$, and a bucket  domain prior. Let $S_{1}$ represent the size of each domain in a domain structure: $S_{1}:=\{|J(c,d)|:d \in \mathscr{D}_{c}\}$. Additionally let $S_{2}$ represent how items are partitioned: $S_{2}:=\{J(c,d):d \in \mathbb{Z}_{D}\}$.}
\label{table:exDomainPrior}
\end{table}

In this example, only domain structures with fewer than three non-empty domains do not satisfy sufficient conditions for generic identifiability. These domain structures fail the Kruskal condition given in \eqref{eq:identifiabilityKruskal}. If we chose $\myvec{\delta}_{c}$ at random without restriction there would only be a $8\mkern1.5mu{\times} 10^{-48}$ prior chance of choosing a non-identifiable domain structure. Finally, note that $\ln P(S_{2})$ in Table~\ref{table:exDomainPrior} represents a penalty term towards simpler domain structures.
\end{example}

We have discussed that the domain prior prioritizes simpler domains for large $D$. Another key question is when we prefer one domain structure over another. Suppose we have two sets of parameters $\omega=(\myvec{\pi},\theta, \mymatrixdelta)$ and $\omega'=(\myvec{\pi}',\theta', \mymatrixdelta')$. Suppose domain structure $\mymatrixdelta'$ dominates $\mymatrixdelta$. We say $\mymatrixdelta'$ dominates $\mymatrixdelta$ if every domain $J(c,d)$ in $\mymatrixdelta$ is the subset of some domain $J'(c,d')$ in $\mymatrixdelta'$. As a consequence $\mymatrixdelta$ is a special case of $\mymatrixdelta'$. For any $\theta$, there exists a Kronecker separable $\theta'$ where $\myvec{X}_{i}|\omega' \overset{d}{=} \myvec{X}_{i}|\omega$.

The prior also gives greater weight to simpler domain structures. We disallow Kronecker separability, but if $\theta'$ was almost Kronecker separable that would be allowed. There exists a Kronecker non-separable sequence $\theta_{t}'$ where $\myvec{X}_{i}|\omega_{t}' \overset{d}{\rightarrow} \myvec{X}_{i}|\omega$ on $t$. So we can choose $\omega_{t}'$ where $\myvec{X}_{i}|\omega_{t}'$ is asymptotically close to $\myvec{X}_{i}|\omega$. Since $\mymatrixdelta'$ is formed by merging domains in $\mymatrixdelta$, we know that $\mymatrixdelta'$ has smaller prior probability:
\begin{align}
\frac{P(\{J(c,d):\in\mathbb{Z}_{D}\})}{P(\{J'(c,d):d\in\mathbb{Z}_{D}\})}
& = \frac{D!}{(D-|\mathscr{D}_{c}|)!D^{J}} \big/ \frac{D!}{(D-|\mathscr{D}_{c}'|)!D^{J}} \\
& = \prod_{k=|\mathscr{D}_{c}'|}^{|\mathscr{D}_{c}|} (D-k).
\end{align}

This gives a strong bias towards the simpler model.
As $t$ increases, $\omega_{t}'$ becomes approximately Kronecker separable. In the fully Kronecker separable case where $\myvec{X}_{i}|\omega' \overset{d}{=} \myvec{X}_{i}|\omega$ we have the following ratio:
\begin{align}
& \frac{P(\mymatrix{X},\myvec{\theta},\myvec{\pi}=\myvec{\pi}_{0},\{J(c,d):\in\mathbb{Z}_{D}\})}{P(\mymatrix{X},\myvec{\theta}',\myvec{\pi}'=\myvec{\pi}_{0},\{J'(c,d):\in\mathbb{Z}_{D}\})}
= \frac{P(\theta|\mymatrixdelta)}{P(\theta'|\mymatrixdelta')} \prod_{k=|\mathscr{D}_{c}'|}^{|\mathscr{D}_{c}|} (D-k).
\label{eq:priorRatio1}
\end{align}

In the following section, we introduce the \quote{pattern adjusted prior} which regularizes even further towards simpler models. This alternate prior effectively cancels out $P(\theta|\mymatrixdelta)$ in Equation \eqref{eq:priorRatio1}.

\subsection{Pattern Adjusted Priors}

The pattern adjusted domain prior is an extension of the bucket  domain prior with a stronger regularization. Let the domain prior be  $g(\mymatrixdelta)=P(\mymatrixdelta)=$ with $g_{p}(\mymatrixdelta)$ referring to the bucket  prior and $g_{r}(\mymatrixdelta)$ referring to the pattern adjusted prior.  The pattern adjusted prior $g_{r}(\mymatrixdelta)$ is defined as:
\begin{align}
g_{r}(\mymatrixdelta)
&= \kappa \frac{g_{p}(\mymatrixdelta)}{\prod_{c,d} \Gamma(R_{c d})}
\label{eq:restrictiveDef}
\end{align}
with some normalizing constant $\kappa$ where $R_{c d}$ indicates the number of response patterns to domain $(c,d)$. Unlike its counterpart, pattern adjusted priors require that the the prior of $\theta$ is uniform Dirichlet with $\alpha^{(\theta)}=1$. Under this restriction, $P(\theta_{c d}|\mymatrixdelta)=\Gamma(R_{c d})$ for all $\theta_{c d}$. The joint distribution of $P(\theta,\mymatrixdelta)$ is,
\begin{align}
P(\theta,\mymatrixdelta)
= g(\mymatrixdelta) P(\theta|\mymatrixdelta)
= g(\mymatrixdelta)\prod_{c d} \Gamma(R_{c d}).
\end{align}
Note that $\Gamma(R_{c d})\Gamma(R_{c' d'}) < \Gamma(R_{c d}+R_{c' d'})$ in general. For this reason $P(\theta|\mymatrixdelta)$ is higher for more complex, dominating, domain structures. See the following example:

\begin{example}
Suppose we have $20$ binary items and one item with five categories. If each item is in their own domain then $P(\theta|\mymatrixdelta)=\Gamma^{20}(2)\Gamma(5)=24$. On the other hand if two particular binary items are put into the same domain then $P(\theta|\mymatrixdelta)=\Gamma^{18}(2)\Gamma(5)\Gamma(4)=6(24)$. If instead the  five-category item is combined with a particular binary item, the prior probability increases even further. We then have $P(\theta|\mymatrixdelta)=\Gamma^{19}(2)\Gamma(10)=(15120)(24)$.
\end{example}

For bucket  domain priors, the behavior of $P(\theta|\mymatrixdelta)$ can lead to counterinuitive results. Even though the bucket  prior $g_{p}(\mymatrixdelta)$ biases towards simpler domains, $P(\theta|\mymatrixdelta)$ biases towards more complicated domains. As a result jointly $P(\theta,\mymatrixdelta)$ can prioritize more complicated domains for every choice of $\theta$. Pattern adjusted domain structures on the other hand address this problem by cancelling out $P(\theta|\mymatrixdelta)$ as follows:
\begin{align}
P(\theta,\mymatrixdelta)
= g_{r}(\mymatrixdelta) P(\theta|\mymatrixdelta)
= \kappa\frac{g_{p}(\mymatrixdelta)}{\prod_{c d} \Gamma(R_{c d})}\prod_{c d} \Gamma(R_{c d})  
=\kappa g_{p}(\mymatrixdelta).
\end{align}
As a result, the ratio in Equation \eqref{eq:priorRatio1} becomes:
\begin{align}
& \frac{P(\mymatrix{X},\myvec{\theta},\myvec{\pi}=\myvec{\pi}_{0},\{J(c,d):\in\mathbb{Z}_{D}\})}{P(\mymatrix{X},\myvec{\theta}',\myvec{\pi}'=\myvec{\pi}_{0},\{J'(c,d):\in\mathbb{Z}_{D}\})}
= \prod_{k=|\mathscr{D}_{c}'|}^{|\mathscr{D}_{c}|} (D-k).
\label{eq:priorRatio2}
\end{align}
Recall the issue caused by (near) Kronecker separability. For nearly any choice of $\mymatrixdelta$ there is another $\mymatrixdelta'$ which dominates it. Using near Kronecker non-separability, for a given choice of $\omega = (\theta, \myvec{\pi}, \mymatrixdelta)$ there exists a $\omega' = (\theta', \myvec{\pi}', \mymatrixdelta')$ where $\myvec{X}_{i}|\omega' \overset{d}{\approx} \myvec{X}_{i}|\omega$. In other words, both choices fit the data almost equally well. This similar goodness of fit could pose a problem because we seek to choose between $\mymatrixdelta$ and $\mymatrixdelta'$. For this reason our domain priors bias towards simpler models. For our pattern adjusted prior in particular, the simpler model will be much more common than any single dominating alternative. This is valuable because the simpler model eliminates unnecessary dependencies.

In Section~\ref{section:simStudies} Simulation Studies,  we compare different domain priors. Both the bucket  and pattern adjusted priors perform well.

\subsection{Heterogenous Domain Prior}

For both bucket  and pattern adjusted priors, the prior for heterogeneous domain structures are constructed similarly. Consider a heterogeneous DLCM. Before applying restrictions, each $\myvec{\delta}_{c}$ and $\myvec{\delta}_{c'}$ are independent for $c\neq c'$. In other words, every $\mymatrixdelta\in\mathbb{Z}_{D}^{J\times C}$ in the support has equal prior probability. When we apply restrictions we reduce the support and then scale up all probabilities by a normalizing constant. This implies that the probabilities of each  $\myvec{\delta}_{c}$ are proportional to the probabilities given without restriction. Therefore, within the support, the prior probability of $\mymatrixdelta$ is proportional to the product of independent probabilities. For bucket  priors these probabilities are given in Theorem~\ref{thm:domainPrior}, and for pattern adjusted priors these are given by Equation \eqref{eq:restrictiveDef}.

In partially heterogeneous domain structures some, but not all, of the classes are restricted to have the same domains. More specifically, there is a hyperparameter $\myvec{\mathscr{E}} \in \mathbb{Z}_{C}^{C}$.  If $\mathscr{E}_{c} = \mathscr{E}_{c'}$ then classes $c,c'$ have the same domains: $\delta_{j c}=\delta_{j c'}$ $\forall j$. For each group of classes with the same domains, we choose a representative class $c$. Before applying restrictions, the domains $\myvec{\delta}_{c}$ are independent. In this way the partially heterogeneous domain prior is analogous to the heterogeneous domain prior if you consider just the representative classes.

\section{MCMC}\label{section:mcmc}

In this section we discuss our algorithm to approximate the posterior distribution of our parameters $\omega | \mymatrix{X}$. This is done by way of MCMC sampling with a Metropolis-Hastings within Gibbs sampler. This sampler generates $\mathsf{MaxItr}$ observations ostensibly from $\omega | \mymatrix{X}$. We denote the $t$'th iteration of the MCMC parameters as $\omega^{(t)}:=(\myvec{\pi}^{(t)},\theta^{(t)}, \mymatrixdelta^{(t)})$. The MCMC steps are as follows:

\begin{enumerate}
\item Use Metropolis-Hastings to sample $\mymatrixdelta^{(t)}$ collapsed on $\theta$. This step is repeated a specified number of times ($\mathsf{nDomainIters}$ times).
\item Sample $\theta^{(t)}$ with Gibbs using full conditional distribution as given in Theorem~\ref{thm:posterior}.
\item Sample $\myvec{\pi}^{(t)}$ with Gibbs using full conditional distribution as given in Theorem~\ref{thm:posterior}.
\item Sample $\myvec{c}^{(t)}$ with Gibbs using full conditional distribution as given in Theorem~\ref{thm:posterior}.
\item[ ]Increment $t$ and return to step 1 and repeat until $\mathsf{MaxItr}$ iterations have been reached.
\end{enumerate}

The Gibbs steps sample one parameter conditional on the others. For instance, the next value of $\theta^{(t+1)}$ is generated by the distribution of $\theta^{(t+1)}|\mymatrixdelta^{(t)},\myvec{\pi}^{(t)},\myvec{c}^{(t)},\mymatrix{X}$ given in Theorem~\ref{thm:posterior}. For more information on Gibbs sampling see \citet{Gelfand2000}.

The Metropolis-Hastings step for the domain structure requires a dedicated discussion.

\subsection{Domain Metropolis-Hastings Step, Concepts}

A Metropolis algorithm works roughly by proposing a new value of $\mymatrixdelta$. Depending on how likely the proposed value is, the algorithm will either stay at its current value or move to the new value with some probability. \citet{Chib1995} provide some useful details.

The proposal is done by taking two domains at random and mixing items between them. Every item in $J(c,d_{1})\cup J(c,d_{2})$ is equally likely to end up in either domain, up to some small restrictions such as identifiability. Since the mixing procedure allows for any partition of $J(c,d_{1})\cup J(c,d_{2})$, it can correct potentially large problems with a candidate domain structure and escape many local minima. Consider the following examples. 

\begin{example}
Suppose the true domain structure of a dataset includes the domains $J_{\text{truth}}(c,0) = A = \{a_{0},a_{1},a_{2},a_{3}\}$ and $J_{\text{truth}}(c,1) = B=\{b_{0},b_{1},b_{2},b_{3}\}$, and $J_{\text{truth}}(c,2) = E=\{e_{0},e_{1}, e_{2}, e_{3}\}$.
\begin{itemize}
\item Suppose our MCMC algorithm is at $J(c,0)= \{a_{1},a_{2}\}$ and $J(c,1)= \{a_{3},a_{4}\}$. If $J(c,0)$ and $J(c,1)$ are mixed, one possible proposal is $J(c,0)=A$ and $J(c,1)=\emptyset$. Notice the importance of moving all of the items at once. If only a single item was moved the goodness of fit might have actually worsened.  A proposal of $\{a_{1}\}$ and $\{a_{2},a_{3},a_{4}\}$ would be worse if the predictions of $a_{1}$ were poor enough.
\item Suppose our MCMC algorithm is at $J(c,0) = A \cup B$. If $J(c,0)$ is mixed with $J(c,d)=\emptyset$ then a possible proposal is $J(c,0)=A$ and $J(c,1)=B$. This splits the domain into its independent parts without losing the conditional dependence within each $A$ and $B$.
\item Suppose our MCMC algorithm is at $J(c,0)=\{a_{0},a_{1},b_{0},b_{1}\}$ and $J(c,1)=\{a_{2},a_{3},b_{2},b_{3}\}$. If $J(c,0)$ and $J(c,1)$ are mixed, one possible proposal is $J(c,0)=A$ and $J(c,1)=B$. Notice the importance of moving all of the items at once. If only a single item was moved the goodness of fit might have actually worsened.  A proposal of $\{a_{0},b_{0},b_{1}\}$ and $\{a_{1},a_{2},a_{3},b_{2},b_{3}\}$ would be worse if the predictions of $a_{1}$ were poor enough.
\end{itemize}

In practice a particular proposal might not operate as described above. A different pair of domains might be mixed. Even if the described domains are mixed, the items might be partitioned differently. However since these proposals can occur, they indeed will occur over sufficiently many iterations.
\end{example}

\subsection{Domain Metropolis-Hastings Step, Details}

In this subsection we explicitly define the steps used to propose and evaluate a new domain structure.

Although our algorithm generates new values of $\mymatrixdelta^{(t)}$, we do not care about the specific labelings of domains. The algorithm treats the domain labels strictly as an operational tool. Rather, we are interested in how items are partitioned into domains: $\{J(c,d):c=c_{0},d \in \mathscr{D}_{c_{0}}\}$. When we consider probabilities, for instance the prior probability of $\mymatrixdelta$, we consider the probability of the given partition rather than specific labeling of $\mymatrixdelta$.

Recall that the restriction $\myvec{\mathscr{E}}$ forces certain classes to have the same values of $\myvec{\delta}_{c}$. We loop through each group of equivalent classes, and run our Metropolis algorithm $\mathsf{nDomainIters}$ times for each group. By default we set $\mathsf{nDomainIters}=J$ in our models.

For a group of equivalent classes, fix some representative class $c$. We start by describing how a new value of $\mymatrixdelta$ is proposed. We will take $\myvec{\delta}_{c}^{(t)}$ and build a proposed $\myvec{\delta}_{c}'$. First a nonempty domain $d_{1} \in \mathscr{D}_{c}$ is chosen entirely at random. Next a second distinct domain $(c,d_{2})$ is chosen as follows. If $J(c,d_{1})$ has only a single item, then $J(c,d_{2})$ is chosen at random among nonempty domains. If $J(c,d_{1})$ contains multiple items, then $J(c,d_{2})$ will be an empty domain with probability $p_{\text{\tiny empty}}$ and a random nonempty domain with probability $1-p_{\text{\tiny empty}}$. Since we are indifferent to the labeling of domains, when selecting an empty domain the first empty domain from $\mathbb{Z}_{D}-\mathscr{D}_{c}$ is chosen.

With domains $(c,d_{1})$ and $(c,d_{2})$ chosen, items are now mixed between these domains. Every item in $J(c,d_{1})\cup J(c,d_{2})$ is equally likely to end up in either domain, up to two restrictions. First, we require that the proposal changes the partitioning of items: $\{J(c,d_{1}),J(c,d_{2})\}$. Second, we require that the $\mathsf{MaxItems}$ constraint is respected. If these conditions are not met, the items are randomly divided again until a valid proposal is reached. This swapping of items fully characterizes the proposed $\myvec{\delta}_{c}'$.

For the chosen proposal we calculate the proportional probability of proposing $\myvec{\delta}^{(t)} \rightarrow \myvec{\delta}'$ and $\myvec{\delta}' \rightarrow \myvec{\delta}^{(t)}$. The calculations of these proposal probabilities can be found in online Appendix~\ref{appendix:mcmc}.

Next we evaluate if $\mymatrixdelta'$ is identifiable with respect to the Kruskal condition \eqref{eq:identifiabilityKruskal}. If it is not identifiable then $\mymatrixdelta'$ has zero prior probability, the proposal $\mymatrixdelta'$ is rejected, and the domain structure does not change $\mymatrixdelta^{(t+1)}=\mymatrixdelta^{(t)}$. 

From Theorem~\ref{thm:identifiability}, we know that if the Kruskal inequality $M := min(\kappa_{0},C)+min(\kappa_{1},C)+min(\kappa_{2},C) \geq 2C + 2$ holds then we have generic identifiability. We evaluate this inequality by way of a greedy deterministic algorithm. We start with three empty partitions. Groups of conditionally independent items (denoted \quote{pooled} domains in appendix) are then taken one at a time and added (greedily) to the partition which incrementally maximizes the value of $M$. To ensure that the pooled domains are always evaluated in the same order, the pooled domain with the most number of patterns is processed first, and subsequent domains are processed in descending order. If after all pooled domains are processed $M < 2C + 2$ then $\mymatrixdelta'$ is possibly non-identifiable and the proposal is rejected. 

It is possible there is a different tripartition which permits the Kruskal inequality. However brute forcing every possible tripartition would be too expensive to evaluate. In fact, even if the Kruskal inequality did not hold for \stress{any} tripartition, this does not eliminate the possibility that $\mymatrixdelta'$ is generically identifiable. Theorem~\ref{thm:identifiability} provides sufficient rather than necessary conditions. By using this algorithm, we limit $\mymatrixdelta$ to domain structures which we can easily verify are identifiable.

The remainder of the $\mymatrixdelta$ Metropolis step proceeds as usual. Let $p_{f}$ be the \quote{forward} probability of proposing $\mymatrixdelta^{(t)}\rightarrow \mymatrixdelta^{(t+1)}$, and $p_{b}$ be the \quote{backward} probability of proposing $\mymatrixdelta^{(t+1)}\rightarrow \mymatrixdelta^{(t)}$. We build the ratio:
\begin{align}
a & = \frac{P(\{J'(c,d):d \in \mathscr{D}_{c_{0}}\})P(\mymatrix{X}|\myvec{\pi}',\myvec{c}',\mymatrixdelta')}{P(\{J^{(t)}(c,d):d \in \mathscr{D}_{c_{0}}\})P(\mymatrix{X}|\myvec{\pi}^{(t)},\myvec{c}^{(t)},\mymatrixdelta^{(t)})}\ \frac{p_{b}}{p_{f}}.
\end{align}
With probability $\min\{a,1\}$ we accept the proposal. That is, we generate a random value $u \sim \text{Uniform}(0,1)$. If $u \leq a$, then $\delta^{(t+1)} = \delta'$. Otherwise if $u > a$, then $\delta^{(t+1)} = \delta^{(t)}$.

\subsection{Heterogeneous and Partially Heterogeneous MCMC}

Running MCMC on heterogeneous and partially heterogeneous domain structures require some special considerations.

When we fit a heterogeneous DLCM the first $\mathsf{nHomoItrs}$ warmup iterations are done with homogeneous domain restrictions. This is done to move the process to a good initial value of $\mymatrixdelta$ before examining heterogeneity. By default we set $\mathsf{nHomoItrs}$ to 5\% of the total iterations or $1,000$ iterations, whichever is lower. Anecdotally we have found that these early homogeneous iterations help prevent the DLCM from getting stuck at local extrema. In particular, we found better recovery of truth (in simulation studies), and overall improved goodness of fit.

If poor starting conditions are given, partially heterogeneous DLCMs can be slow to converge to truth. Consider the following problematic example.

\begin{example}\label{ex:parthet}
Suppose we have three latent classes; the first two of which, in truth, have the same domains. Suppose our initial conditions correctly classify class membership $\myvec{c}$, but incorrectly specify restriction $\myvec{\mathscr{E}} = [0,1,1]^{\top}$. Here the last two classes are restricted to have the same domain structure rather than the first two. Since $\myvec{\mathscr{E}}$ is fixed, this would need to be corrected by changing $\myvec{c}$. Ideally, what we would like to see is label swapping where the members of the first and third classes switch places. However in LCMs label swapping occurs infrequently and is generally undesired as it can cause inference problems.
\end{example}

In real world applications we suggest the following procedure for fitting partially heterogeneous DLCMs. First, fit a fully heterogeneous DLCM. Second, hypothesize which classes should have the same domains. Third, fit a partially heterogeneous DLCM with classes $\myvec{c}$ seeded from the last model.

\section{Simulation Studies}\label{section:simStudies}

\subsection{Simulation Studies: Default Seeds}

We conducted Monte Carlo simulation studies to validate the accuracy of the proposed DLCM algorithms. Here, we will generate random datasets following a specific distribution. Then we see how well our models recover the true underlying distribution.

We considered three types of datasets. In the first case \stress{Traditional Data}, we generate data following the distribution of a traditional LCM. All items are conditionally independent given latent classes. The goal of the first simulation is to demonstrate that the DLCM accurately recovers conditionally independent domain structures. In the second case \stress{Homogeneous Data}, we generate data following the distribution of a homogeneous DLCM. Conditional dependence appears, but it is the same across classes. In the third case \stress{Heterogeneous Data}, we generate data following the distribution of a heterogeneous DLCM. Conditional dependence exists and differs across classes. For simplicity these simulated datasets use Bernoulli data, $C=2$ classes, and $J=24$ items. In Section \ref{section:appSociology} we provide a real world example of DLCMs on polytomous data and more classes.

Within each of the three scenarios for a given sample size $n$, we generate $100$ datasets. For each dataset, we fit a number of different models. We fit with traditional LCMs, homogeneous DLCMs, and heterogeneous DLCMs in turn. Each DLCM is fit with three different priors: uniform, bucket, and pattern adjusted.  The uniform prior assumes that every domain structure is equally likely, and serves as a baseline comparison similar to \cite{Marbac2014}. The bucket and pattern adjusted priors are our proposed regularizing priors discussed in Section~\ref{section:domainPrior}. More information on how the data was generated, how the models are tuned, and more detailed results can be found in online Appendix \ref{appendix:sims}.

We say that a simulation recovered the true domain structure if the most common domain structure across iterations matched the truth. This is shown in Table~\ref{table:simModeAccuracy}. In all cases, the baseline uniform prior does quite poor. We show good recovery for homogeneous DLCMs, heterogeneous DLCMs, bucket priors, and pattern adjusted priors. However heterogeneous models require larger sample sizes owing to their complexity, and restrictive priors require larger sample sizes owing to the strength of the prior. In Table~\ref{table:simFirstCorrect}, we see that the DLCMs typically reach the true domain structure early, often in the first $100$ MCMC iterations. Our simulations also ran quickly with $97\%$ of simulations with sample sizes of $n=1,000$ completing in $2$ minutes or less.

\begin{table}[]
\centerline{
\begin{tabular}{llllrrrrrr}
\multicolumn{4}{l}{Domain Structure Mode Accuracy}                                      & \multicolumn{6}{l}{\makecell[l]{\\Sample Size}}               \\
Data    & Model              & Prior         & Seed     & $n=100$   & 200   & 300   & 400   & 500   & 1000  \\\hline
Traditional    & Homogeneous   & Uniform & Default & 0\%  & 0\%   & 0\%   & 0\%   & 0\%   & 1\%   \\
Traditional    & Homogeneous   & Bucket         & Default & {96\%}  & 96\%  & 97\%  & 100\% & 99\%  & 100\% \\
Traditional    & Homogeneous   & Pattern Adjusted        & Default & {100\%} & 100\% & 100\% & 100\% & 100\% & 100\% \\
Traditional    & Heterogeneous & Uniform & Default & 0\%   & 0\%   & 0\%   & 0\%   & 0\%   & 0\%   \\
Traditional    & Heterogeneous & Bucket         & Default & {95\%}  & 94\%  & 96\%  & 97\%  & 99\%  & 100\% \\
Traditional    & Heterogeneous & Pattern Adjusted        & Default & {99\%}  & 99\%  & 100\% & 100\% & 100\% & 100\% \\
\\[-3ex]
Homogeneous   & Homogeneous   & Uniform & Default & 0\%   & 0\%   & 0\%   & 0\%   & 1\%   & 10\%  \\
Homogeneous   & Homogeneous   & Bucket         & Default & {96\%}  & 99\%  & 98\%  & 97\%  & 97\%  & 97\%  \\
Homogeneous   & Homogeneous   & Pattern Adjusted        & Default & 37\%  & {99\%}  & 97\%  & 98\%  & 99\%  & 98\%  \\
Homogeneous   & Heterogeneous & Uniform & Default & 0\%   & 0\%   & 0\%   & 0\%   & 0\%   & 0\%   \\
Homogeneous   & Heterogeneous & Bucket         & Default & 7\%   & 51\%  & {86\%}  & 94\%  & 93\%  & 98\%  \\
Homogeneous   & Heterogeneous & Pattern Adjusted        & Default & 0\%   & 4\%   & 34\%  & {78\%}  & 94\%  & 98\%  \\
Homogeneous   & Homogeneous   & Bucket           & Random & 93\%  & 99\%  & 98\% & 98\%  & 96\%  & 96\%  \\
Homogeneous   & Homogeneous   & Pattern Adjusted & Random & 35\%  & 97\%  & 97\% & 94\%  & 100\% & 97\%  \\
\\[-3ex]
Heterogeneous & Heterogeneous & Uniform & Default & 0\%   & 0\%   & 0\%   & 0\%   & 0\%   & 0\%   \\
Heterogeneous & Heterogeneous & Bucket         & Default & {80\%}  & 99\%  & 98\%  & 100\% & 96\%  & 99\%  \\
Heterogeneous & Heterogeneous & Pattern Adjusted        & Default & 8\%   & {100\%} & 99\%  & 100\% & 100\% & 100\% \\
Heterogeneous & Heterogeneous & Bucket           & Random & 79\%  & 98\%  & 98\% & 100\% & 96\%  & 100\% \\
Heterogeneous & Heterogeneous & Pattern Adjusted & Random & 10\%  & 100\% & 99\% & 100\% & 100\% & 100\%
\end{tabular}
}
\caption{From simulation studies with $C=2$ classes. Across 100 generated datasets, in what percent did the most common posterior domain structure match the truth?}
\label{table:simModeAccuracy}
\end{table}

\begin{table}[]
\centerline{
\begin{tabular}{llllrrrrrr}
\multicolumn{4}{l}{First Iteration at True Domain Structure}                                      & \multicolumn{6}{l}{\makecell[l]{\\Sample Size}}               \\
Data               & Model         & Prior       & Seed     & n=100 & 200 & 300   & 400 & 500   & 1000  \\\hline
Homogeneous   & Homogeneous   & Bucket           & Default & 31    & 28  & 34  & 32  & 34  & 26   \\
Homogeneous   & Homogeneous   & Pattern Adjusted & Default & 57    & 30  & 31  & 25  & 26  & 30   \\
Heterogeneous & Heterogeneous & Bucket           & Default & 415   & 329 & 317 & 314 & 314 & 315  \\
Heterogeneous & Heterogeneous & Pattern Adjusted & Default & 3,535 & 334 & 324 & 318 & 323 & 325  \\
\\[-3ex]
Homogeneous   & Homogeneous   & Bucket           & Random & 30    & 31  & 30  & 29  & 25  & 26   \\
Homogeneous   & Homogeneous   & Pattern Adjusted & Random & 49    & 31  & 27  & 25  & 28  & 25   \\
Heterogeneous & Heterogeneous & Bucket           & Random & 432   & 328 & 317 & 313 & 314 & 313  \\
Heterogeneous & Heterogeneous & Pattern Adjusted & Random & 3,549 & 342 & 327 & 322 & 320 & 317  \\
\end{tabular}
}
\caption{From simulation studies with $C=2$ classes. For each dataset, at what MCMC iteration did we first observe the true domain structure? Median is taken across the 100 simulated datsets. Form homogeneous DLCMs, iterations 1-1000 are warmup iterations. For heterogeneous DLCMs iterations 1-300 are warmup iterations using homogeneous assumptions, and iterations 301-1300 are warmup iterations using heterogeneous assumptions. For a heterogeneous DLCM on heterogeneous data, the first iteration where truth could possibly occur is iteration 301 when heterogeneous assumptions are introduced.}
\label{table:simFirstCorrect}
\end{table}

\subsection{Simulation Studies: Random Seeds}

In this subsection we examine DLCM performance when the starting conditions are not chosen favorably. We initialize domains at random using the uniform prior. In this way the initial domains will typically be far from truth. We also seed class membership using independent Bernouli variables. This will cause starting class membership to start far from the truth. In Tables \ref{table:simModeAccuracy} and \ref{table:simFirstCorrect} the Seed of \quote{random} indicates these alternate starting conditions are used. We see that performance using this versus the default seed are very similar.

\section{Real World Applications}\label{section:applications}

In this section we illustrate the effectiveness of DLCMs in three real world examples. We consider applications for datasets related to issues in education, pediatric health, and adolescent sociology. The education application examines pre-post testing and highlights how DLCMs can identify local dependence between two time-points. The pediatric medical application examines a time series and highlights how DLCMs can identify local dependence within each time-point. Finally the sociological example contains overlapping questions and highlights how DLCMs can identify structural zeros.

We fit the DLCMs to each application dataset with four chains, $2,000$ warmup iterations, $10,000$ main iterations, and $\mathsf{nHomoItrs}=600$. When updating classes $\myvec{c}$ our Gibbs step collapses on $\theta$ and $\myvec{\pi}$. Otherwise these examples were executed with the same hyperparameters as the simulation studies.

To measure MCMC convergence we use the Gelman-Rubin statistic. We jointly measure the convergence of class probabilities $\myvec{\pi}^{(t)}$, marginalized item probabilities $P(X_{ij}|c_{i}=c,\omega^{(t)})$, and log-likelihood $\ln P(\mymatrix{X}|\omega^{(t)})$.

\subsection{Education Application}\label{section:application-education}

In this experiment, participants' skill with probability theory was assessed. The study followed a pre-post design. First, participants are given a 12-question pre-test (questions B101-B112). Then participants were randomly given one of two treatments. Finally, participants are given a 12-question post-test with matched items (B201-B212). Each post-test question  mirrored a pre-test question with slightly different numbers or labels. For example, B105 from the pre-test matches with B205 in the post-test. All questions are listed in online Appendix~\ref{appendix:education}. This data was collected by Anselmi, et al. (2010) and is freely available in the `pks' R package \citep{Anselmi2010}. For each question we examined the Bernouli responses: 1 for correct and 0 for incorrect.

Subjects were eliminated if they responded too quickly, responded too slowly, or did not answer every question. This left $n=345$ participants considered. Both treatments were assessed within the same latent class model without differentiation. A potential goal was to find relationships between our latent classes and the treatments.

When models were compared, the homogeneous DLCM with bucket prior performed best. In Table~\ref{table:education-gof} we see that the homogeneous DLCM has both a higher likelihood (LPPD) and smaller model complexity (WAIC penalty) compared to the traditional model. The three class homogeneous DLCM shows good MCMC convergence with a multivariate Gelman-Rubin statistic of 1.02.

The homogeneous DLCM fits well with three classes. There are proficient students (80\% of participants), beginners (17\%), and students who performed worse in their post-test than their pre-test (3\%). Each latent class has a roughly even amount of subjects from each treatment. Thus, there is not a large effect size between our latent classes and these treatments. Information on these classes including response probabilities can be found in online Appendix~\ref{appendix:education}.

Table~\ref{table:education-domainshomo} reports the most frequently visited domain structures. The mode domain structure contains three pairs of pre/post items: \{b105,b205\}; \{b108,b208\}, \{b109,b209\}. For these paired items, the participants typically got a pair both right or both wrong. The homogeneous DLCM also grouped some pre-test items which were especially difficult: \{b104,b110,b111,b112\}. In this domain participants have heavy concentrations on `all right' and `all wrong', possibly owing to the skill level needed to solve these problems. See online Appendix~\ref{appendix:education} for details.

Compared to a traditional model, the homogeneous DLCM produced both a simpler and more accurate fit. The homogeneous DLCM was able to identify paired questions in the pre-post design and incorporate their conditional dependence into its model.

\begin{table}[!htbp]
\begin{tabular}{llcrrr}
Model Name     & Prior   & \# of Classes & LPPD   & WAIC Penalty & WAIC   \\\hline
Traditional LCM   &             & 7 & -2,503 & 127 & 5,260 \\
Homogeneous DLCM  & Bucket  & 3 & -2,502 & 83  & 5,170 \\
Homogeneous DLCM  & Pattern Adjusted & 5 & -2,503 & 106 & 5,218 \\
Heterogenous DLCM & Bucket  & 4 & -2,506 & 95  & 5,202 \\
Heterogenous DLCM & Pattern Adjusted & 5 & -2,493 & 120 & 5,226 \\
\end{tabular}
\caption{Education Application. Best models by type.}
\label{table:education-gof}
\end{table}

\begin{table}[!htbp]
\begin{tabular}{ll}
Domains (\{Domain1\}; \{Domain2\}; ...)                                                & \% of Iterations      \\\hline
\{b104,b110,b111,b112\}; \{b105,b205\}; \{b108,b208\}; \{b109,b209\} & 87.3\% \\
\{b104,b110,b111\}; \{b105,b205\}; \{b108,b208\}; \{b109,b209\} & 4.0\% \\
\{b104,b110,b111,b112\}; \{b105,b205\}; \{b108,b208\}; \{b109,b209\}; \{b106,b107\} & 2.5\%
\end{tabular}
\caption{Education Application. Most Common Homogeneous Domain Structures.}
\label{table:education-domainshomo}
\end{table}

\subsection{Medical Application}\label{section:application-medical}

This application is drawn from a Childhood Prevention Study (CAPS) by \citet{Mihrshahi2001} (available in the R randomLCA package). This is a longitudinal study which investigated children originally under two years of age. Every six months for two years these children were tracked for nighttime coughing, wheezing, itchy rashes, and flexural dermatitis. This creates four time periods where we indicate the absence (0) or presence (1) of each symptom. We removed any observations with missing data leaving $n=533$ subjects.

When models were compared, a homogeneous DLCM with bucket prior fit best. The homogeneous DLCM was effective with four classes. Traditional LCM performed worst requiring eight or more classes. The homogeneous DLCM has both less model complexity and improved fit compared to the traditional model (Table~\ref{table:medical-gof}). We have adequate convergence with a multivariate Gelman-Rubin statistic of $<1.025$.

Homogeneous DLCM provides four classes: bad lungs (30\% of subjects), bad skin (17\%), bad all symptoms (14\%), and good all symptoms (39\%). Symptoms generally improve as time goes on. See Table~\ref{table:medical-classes} in the supplemental appendix for symptom prevalence within each class.

The most common domain structure can be found in Table~\ref{table:medical-homodomains}. This domain structure pairs related symptoms for the same visit. The two lung issues are related: nighttime coughing and wheezing. Similarly, the two skin issues are related: itchy rashes and flexural dermatitis. Paired symptoms are typically comorbid. See online Appendix~\ref{appendix:medical} for details.

Overall, the homogeneous model was able to produce a simpler and more accurate model compared to the traditional LCM. In the context of longitudinal data, the homogenous DLCM was able to identify dependence between questions at the same time-point. 

\begin{table}[!htbp]
\begin{tabular}{llrrrr}
Model Name   & Prior     & \# of Classes & LPPD   & WAIC Penalty & WAIC  \\\hline
Traditional LCM   &             & 8 & -4,383 & 136 & 9,037 \\
Homogeneous DLCM  & Bucket  & 4 & -4,271 & 78  & 8,698 \\
Homogeneous DLCM  & Pattern Adjusted & 4 & -4,276 & 86  & 8,723 \\
Heterogenous DLCM & Bucket  & 4 & -4,286 & 94  & 8,760 \\
Heterogenous DLCM & Pattern Adjusted & 3 & -4,318 & 77  & 8,789
\end{tabular}
\caption{Medical Application. Goodness of fit for top models.}
\label{table:medical-gof}
\end{table}

\begin{table}[!htbp]
\scalebox{0.7}{
\begin{tabular}{ll}
Domains  & \% of Iterations \\\hline
\{IR.1,FD.1\}; \{IR.2,FD.2\}; \{IR.3,FD.3\}; \{IR.4,FD.4\}; \{NC.1,W.1\}; \{NC.3,W.3\}; \{NC.4,W.4\}; & 95.3\% \\
\{IR.1,FD.1\}; \{IR.2,FD.2\}; \{IR.3,FD.3\}; \{IR.4,FD.4\}; \{NC.1,W.1\}; \{NC.2,W.2\}; \{NC.3,W.3\}; \{NC.4,W.4\} & 4.7\% \\
All Others & $<0.1\%$ \\
\end{tabular}
}
\caption{Medical Application. Most Common Homogeneous Domains. NC=NightCough, W=Wheeze, IR=ItchyRash, FD=FlexDerma.}
\label{table:medical-homodomains}
\end{table}

\subsection{Sociology Application}\label{section:appSociology}

This application was drawn from a CDC Youth Risk Behavior Survey (YRBS) \citep{cdc2017}. We isolated thirteen questions about sexual violence and sexual risk (e.g. STDs) answered by high school women. Each response is polytomous with 2-5 possible responses depending on the question. We removed any observations with missing data leaving $n=1,295$ subjects.

When models were compared, heterogeneous DLCM with bucket prior fit best. For details see Table~\ref{table:sociology-gof}. A four class heterogeneous DLCM has good convergence with a multivariate Gelman-Rubin statistic of $<1.01$.

The heterogeneous DLCM identified three classes: class 0 \quote{Sexually Active} (27\% of participants), class 1 \quote{Not Sexually Active} (56\%), and class 2 \quote{At Risk} (17\%). The marginal response probabilities can be found in online Appendix~\ref{appendix:sociology}.

The most common domain structures can be found in Table~\ref{table:sociology-domainshet}. Questions Q20 and Q21 are two questions about frequency of sexual violence (see Table~\ref{table:sociology-selectquestions}). They form an almost triangular structure with $Q20 \geq Q21$. Questions Q64 and Q65 are overlapping questions about contraceptive use. For details see online Appendix~\ref{appendix:sociology}.

Class 0 \quote{At Risk} has the most complex local dependencies with domains \{Q19,Q20,Q21\} and \{Q64,Q65\}. Class 2 \quote{Sexually Active} is almost as complex with domains \{Q20,Q21\} and \{Q64,Q65\}. Dependence on Q19 was not important because $97\%$ of participants answered this question `no'. Class 1 \quote{Not Sexually Active} is the least complex with domain \{Q20,Q21\}. Responses in class 1 were very concentrated with 11/13 questions each having $\geq 95\%$ of their responses associated to a single value indicating they never had sex (online Appendix~\ref{appendix:sociology}).

The heterogeneous DLCM performed well in this example. In domain \{Q20,Q21\}, we recover near structural zeros. In domain \{Q64,Q65\}, we identify overlapping questions. Different classes have different domains because some classes are heavily concentrated on certain values, reducing the need to manage dependence in corresponding questions.

\begin{table}[!htbp]
\begin{tabular}{llcrrr}
Model Name     & Prior   & \# of Classes & LPPD   & WAIC Penalty & WAIC  \\\hline
Traditional LCM   &             & 7 & -5,690 & 92 & 11,565 \\
Homogeneous DLCM  & Bucket  & 3 & -5,687 & 52 & 11,479 \\
Homogeneous DLCM  & Pattern Adjusted & 3 & -5,687 & 52 & 11,479 \\
Heterogenous DLCM & Bucket  & 3 & -5,652 & 57 & 11,419 \\
Heterogenous DLCM & Pattern Adjusted & 5 & -5,662 & 66 & 11,456 \\
\end{tabular}
\caption{Sociology Application. Goodness of fit for best models.}
\label{table:sociology-gof}
\end{table}

\begin{table}[!htbp]
\begin{tabular}{rlr}
& Domains (\{Domain1\}; \{Domain2\}; ...) & \% of Iterations \\\hline
\makecell[l]{Class0: \\Class1: \\Class2: }
&\makecell[l]{
\{Q20,Q21\}; \{Q64,Q65\}; \\
\{Q20,Q21\};\\
\{Q19,Q20,Q21\}; \{Q64,Q65\};
}
& 76.4\%           \\\hline
\makecell[l]{Class0: \\Class1: \\Class2: }
& \makecell[l]{
\{Q20,Q21\}; \{Q64,Q65\}; \\
\{Q20,Q21\};\\
\{Q19,Q20,Q21\}; \{Q64,Q65\}; \{Q61,Q62\};
}
& 19.1\%            \\\hline

& All others
& $<5\%$            \\\hline
\end{tabular}
\caption{Sociology Application. Most common heterogeneous domain structures. Domains with a single item are omitted.}
\label{table:sociology-domainshet}
\end{table}

\begin{table}
\subheader{Selected Questions from Survey}
\begin{itemize}
\item[\tikzmark{top 1} Q19:] 
Have you ever been physically forced to have sexual intercourse when you did not want to?
\item[ Q20:] During the past 12 months, how many times did anyone force you to do sexual things that you did not want to do?
\item[\tikzmark{bottom 1} Q21:] During the past 12 months, how many times did someone you were dating or going out with force you to do sexual things that you did not want to do?
\item[\tikzmark{top 2} Q64:] The last time you had sexual intercourse, did you or your partner use a condom?
\item[\tikzmark{bottom 2} Q65:] The last time you had sexual intercourse, what one method did you or your partner use to prevent pregnancy?
\end{itemize}
\VerticalBrace[ultra thick, black]{top 1}{bottom 1}{}
\VerticalBrace[ultra thick, black]{top 2}{bottom 2}{}
\caption{Sociology Application. For other questions see online Appendix~\ref{appendix:sociology}.}
\label{table:sociology-selectquestions}
\end{table}

\section{Conclusion}

In the presence of conditional dependence, traditional LCMs tend to overfit with too many classes. Even with additional classes, traditional LCMs may suffer from model mis-specification leading to poor goodness of fit.

We proposed a Domain LCM (DLCM) model to account for these dependencies. The DLCM works by grouping together conditionally dependent items into conditionally independent domains. We verified the generic identifiability of this model. We also demonstrated the effectiveness of DLCMs in simulation studies and real world applications. In applications we demonstrated that DLCMs are particularly effective at analyzing time series data, pre-post testing, overlapping items, and structural zeros.

One avenue of future research is allowing the order of $\myvec{\mathscr{E}}$ to be stochastic. This would allow label swapping in $\myvec{\mathscr{E}}$ and potentially prevent the problems describe in Example~\ref{ex:parthet}.

\if1\blind
{
\section{Acknowledgements}

We would like to thank Theren Williams, Eric Wayman, and Dr. Kristen Lee for providing valuable feedback on writing style.
} \fi

\nocite{*}
\bibliographystyle{plainnat}

\newpage


\appendix

\section{Motivating Example}

\begin{example}\label{example:motivating}
 
Below we offer a simple example to motivate DLCMs in the context of (near) structural zeros.

Students are given a 7 question math exam. We record whether each student got the right answer (1) or a wrong answer (0). Questions 0-4 are distinct, but questions 5-6 are a two-part question. If you get question 5 wrong you only have a 5\% chance of getting question 6 right, regardless of your skill level.

There are good students (30\% of students), medium students (40\%), and poor students (30\%).
Good students get questions 0-5 right 80\% of the time independently. Medium and poor students get questions 0-5 right 50\% and 20\% of the time, respectively. If question 5 is right, students get question 6 right 80\%, 50\%, and 20\% of the time respectively for good, medium, and poor students.

The question of interest is whether we can  exactly reproduce the above data generating process with LCM models. For simplicity, we ignore our posteriors and ask whether there exists a specific choice of parameters which produce the above distribution. 

\subheader{Traditional LCM} To reproduce with a traditional LCM model you need six classes. Each type of student is split in two: one group for those who answered question 5 correctly and a second group who answered incorrectly.

\begin{table}[!htbp]
\begin{tabular}{lllllll}
Class: & Good A & Good B & Medium A & Medium B & Poor A & Poor B \\
 \hline
$\pi_{c}$ & 0.24 & 0.06 & 0.20 & 0.20 & 0.06 & 0.24 \\
$\rho_{j=0:4,c,q=1}$ & 0.80 & 0.80 & 0.5 & 0.50 & 0.20 & 0.20 \\
$\rho_{5,c,1}$ & 1.00 & 0.00 & 1.00 & 0.00 & 1.00 & 0.00 \\
$\rho_{6,c,1}$ & 0.80 & 0.05 & 0.50 & 0.05 & 0.20 & 0.05
\end{tabular}
\caption{Traditional LCM response probabilities by class. Parameter $\rho_{j,c,q}$ gives the probability $P(X_{ij}=q|c_{i}=c)$ in a traditional LCM. Subscript `$j=0:4$' describes each of the items $j \in \{0,\cdots,4\}$  in turn.}
\end{table}

\subheader{DLCM} To reproduce we put questions Q5 and Q6 into the same domain. We use a homogeneous domain structure with $J(d=5,c) = \{5, 6\}$, and $J(k,0)=\{k\}$ for $0 \leq k < 5$. With the domain structure as described, we reproduce the specified structure with 3 classes.

\begin{table}[!htbp]
\begin{tabular}{lllll}
Class: & Good & Medium & Poor & Pattern Description  \\\hline
$\pi_{c}$ & 0.30 & 0.40 & 0.30 &  \\
$\theta_{d=0:4,c,r=1}$ & 0.80 & 0.50 & 0.20 & $X_{i j} = 1$  \\
$\theta_{5,c,r=0}$ & 0.19 & 0.48 & 0.76 & $X_{i 5}=0$, $X_{i 6}=0$ \\
$\theta_{5,c,r=1}$ & 0.16 & 0.25 & 0.16 & $X_{i 5}=1$, $X_{i 6}=0$ \\
$\theta_{5,c,r=2}$ & 0.01 & 0.03 & 0.04 & $X_{i 5}=0$, $X_{i 6}=1$ \\
$\theta_{5,c,r=3}$ & 0.64 & 0.25 & 0.04 & $X_{i 5}=1$, $X_{i 6}=1$
\end{tabular}
\caption{Domain LCM response probabilities by class. Subscript `$d=0:4$' describes each of the domains $d \in \{0,\cdots,4\}$  in turn.}
\end{table}

The DLCM model was able to reproduce the same responses with half as many classes and close to half as many parameters (47 vs 26 ignoring $\mymatrixdelta$). Additionally, the DLCM is more interpretable owing to fewer classes, and provides a clear signal that questions 5 and 6 are related. The latter is important because in a more complicated survey it might not be obvious that two questions are related. Finally, note that this is a relatively simple example with just a few items, two of which are related and share the same domain. In a more intricate scenario a DLCM has the potential to provide stronger improvements compared to traditional LCM. We demonstrate these benefits later in simulation studies and real world applications.

\end{example}

\section{Identifiability}\label{appendix:identifiability}

\subsection{Pooled Domains}

Pooled domains are used to identify conditionally independent items in a DLCM. These are important for our identifiability arguments in Section~\ref{appendix:identifiability2}. Pooled domains are defined as follows:

\begin{definition}
For a fixed choice domain structure $\mymatrixdelta$, Pooled domains $\mathscr{P}_{k} \subseteq \mathbb{Z}_{J}$ are constructed so that the items of two distinct pooled domains are conditionally independent for all $\theta$. The pooled domain structure $\text{Pooled}(\mymatrixdelta) = \{\mathscr{P}_{0},\mathscr{P}_{1},\cdots,\mathscr{P}_{m}\}$ is the maximal set of pooled domains on $\mymatrixdelta$. That is, the pooled domain structure $\text{Pooled}(\mymatrixdelta)$ partitions our $J$ items into as many pooled domains as possible. 
\end{definition}

To illustrate this definition, consider the following example:

\begin{example}
Suppose a DLCM has domain structure:

\begin{align*}
& \mymatrixdelta = \begin{blockarray}{[cc]l}
1 & 1 & j\smeq0 \\
1 & 2 & 1 \\
2 & 2 & 2 \\
2 & 3 & 3 \\
4 & 4 & 4 \\
4 & 4 & 5 \\
5 & 5 & 6 \\
\end{blockarray}
\end{align*}

Here the pooled domains would be $\{0,1,2,3\}$, $\{4,5\}$, and $\{6\}$. For $\{0,1,2,3\}$ note that any one item is dependent on some other item for one or more classes. Following a similar argument, any nonempty bipartition of $\{0,1,2,3\}$ has conditional dependence between the two parts.
\end{example}

The pooled domain structure exists uniquely and can be constructed using Theorem~\ref{thm:pooledconstruction}:

\begin{theorem}\label{thm:pooledunique}
The pooled domain structure $\text{Pooled}(\mymatrix{\delta})$ exists uniquely.
\end{theorem}

\begin{proof}
Since $\text{Pooled}(\mymatrix{\delta}) = \{\mathbb{Z}\}$ is one possible partition, we know $\text{Pooled}(\mymatrix{\delta})$ exists. Suppose there were two possible pooled domain structures $W,W'\subset 2^{\mathbb{Z}_{J}}$ which partition $\mathbb{Z}_{J}$ into the largest number of pooled domains $|W|=|W'|$. If $W$ and $W'$ are distinct, then there exists some set $S' \in W'$ where $S' \notin W$. Take some $S \in W$ where $S' \cap S \neq \emptyset$. In $W$, items $S'\cap S$ and $S'-S$ are conditionally independent. However in $W'$ these are conditionally dependent. If $S'\cap S$ and $S'-S$ were conditionally independent, then there would exist a larger pooled domain structure where $W'$ is taken and $S$ is split into two pooled domains $S'\cap S$ and $S'-S$.
\end{proof}

\begin{restatable}[]{theorem}{restatepooledconstruction}\label{thm:pooledconstruction}
The following are equivalent:
\begin{enumerate}
\item $\mathscr{P} \in \text{Pooled}(\mymatrixdelta)$
\item $\mathscr{P}$ is formed by merging overlapping domains $J(c, d)$ until no overlaps remain. That is $\mathscr{P}$ is of the form:
\begin{itemize}
\item Union of Domains: $\mathscr{P} = \bigcup_{k=0}^{m} J(c_{k},d_{k})$
\item Overlapping: $J(c_{k},d_{k}) \cap \bigcup_{k'=0}^{k-1} J(c_{k'},d_{k'}) \neq \emptyset$ for all $k \in \mathbb{Z}_{m}$
\item Maximum Coverage: $J(c,d)\cap \mathscr{P} = \emptyset$ for $(c,d) \notin \{(c_{k},d_{k}):k\in\mathbb{Z}_{m}\}$
\end{itemize}
\item Build a graph with one node per item and an edge $j \leftrightarrow j'$ if $\delta_{j c} = \delta_{j' c}$ for any $c$. $\mathscr{P}$ is a maximally connected subgraph of this graph.
\end{enumerate}
\end{restatable}

\begin{proof}
Sketch of proof. If item $j_{0}$ is conditionally dependent with $j_{1}$ for some $\theta$ and $j_{1}$ is conditionally dependent with $j_{2}$ for some $\theta'$ then all three items must be in the same pooled domain. This causes us to put all all items from the same domain into the same pooled domain. Furthermore if two domains contain the same item, the items of both domains are put into the same pooled domain, the pooled domain of that overlapping item. This is repeated until no overlaps remain. Full proof is in online Appendix~\ref{appendix:identifiability}.

\subheader{$1 \rightarrow 2$:} Two partitions with overlapping items must be same partition. Let $\mathscr{P}$, $\mathscr{P}'$ be two pooled domains. If items $S \subseteq \mathscr{P}$ and $S' \subseteq \mathscr{P}'$ are such that $S \cap S' \neq \emptyset$ then $\mathscr{P} = \mathscr{P}'$. 

Items in the same domain $(c,d)$ have conditional dependence (for some $\theta_{c,d}$). Therefore, items $J(c,d)$ must all be in the same pooled domain. Consider the items formed by $S := \bigcup_{k}^{n} J(c_{k},d_{k})$ where $J(c_{k},d_{k}) \cap \bigcup_{k'}^{k} J(c_{k'},d_{k'}) \neq \emptyset$ for all $k \leq n$. Furthermore, let $S$ be as large as possible where $J(c,d)\cap S = \emptyset$ for any additional domain. Items $S$ are all in the same pooled domain denoted $\mathscr{P}$.

We claim $\mathscr{P} = S$. By contradiction suppose $S \subsetneq \mathscr{P}$. We know that $S$ and $\mathscr{P}-S$ are conditionally independent for all $\theta$ because all domains $J(c,d) \in S$ and $J(c',d')$ are disjoint. Therefore, we could make the domain structure larger by splitting $\mathscr{P}$ into $S$ and $\mathscr{P}-S$, a contradiction.

\subheader{$2 \rightarrow 3$:} Take a pooled domain $\mathscr{P}$. We know $\mathscr{P}$ is formed by overlapping domains: $\mathscr{P} = \bigcup_{k=0}^{m} J(c_{k},d_{k})$. We show all items in $\mathscr{P}$ are connected by induction. Base case: we know all items of $J(c_{0},d_{0})$ are connected. Inductive step: suppose all items of $\bigcup_{k=0}^{m'} J(c_{k},d_{k})$ are connected for some $m'<m$. Since $J(c_{k},d_{k}) \cap \bigcup_{k'=0}^{k} J(c_{k'},d_{k'})$, we know every item of $J(c_{m'+1},d_{m'+1})$ is connected to some item of $\bigcup_{k=0}^{m'} J(c_{k},d_{k})$. Therefore, all items in $\bigcup_{k=0}^{m'+1} J(c_{k},d_{k})$ are connected. Take any distinct pooled domains $\mathscr{P}$ and $\mathscr{P}'$. Since these domains are made of non-overlapping domains $J(c,d)$ we know there are no edges connecting these two pooled domains.

\subheader{$3 \rightarrow 1$:} By contradiction, suppose $\mathscr{P}$ is not formed by the maximally connected subgraph. If $\mathscr{P}$ is not closed, then there exists some $\theta$ where $\mathscr{P}$ is conditionally dependent on some outside item. This contradicts that $\mathscr{P}$ is a pooled domain. If $\mathscr{P}$ is closed, then it is formed by two or more maximimally connected subgraphs. This contradicts that $\text{Pooled}(\mymatrix{\delta})$ contains the maximum number of pooled domains. We could break $\mathscr{P}$ into multiple pooled domains using claim $1 \rightarrow 3$.
\end{proof}

\begin{corollary}
For any domain (c,d), the items $J(c,d)$ belong to exactly one pooled domain in $\text{Pooled}(\mymatrixdelta)$.
\end{corollary}

Let $\omega:=(\myvec{\pi},\theta, \mymatrixdelta)$ represent the DLCM parameters. Since pooled domains are conditionally independent for all $\theta$, we can write the probability of $\myvec{X}|\omega,c_{i}$ as:

\begin{gather}
P(\myvec{V}(\mathbb{Z}_{J})^{\top}\myvec{X}_{i}|\omega,c_{i}=c)
 = \prod_{\mathscr{P} \in \text{Pooled}(\mymatrixdelta)} P(\myvec{V}(\mathscr{P})^{\top}\myvec{X}_{i}|\omega, c_{i}=c)
 \label{eq:poolProb}.
\end{gather}

This result is notable because no matter the class membership $c_{i}=c$ we always multiply over the same pooled domains. In comparison, domains vary from class to class and a product over domains would similarly vary across classes.

Rather than look at an individual value of $\myvec{X}_{i}$, we want to extend \eqref{eq:poolProb} to a probability vector evaluating every possible value of $\myvec{X}_{i}$. To this end we define the following:

\begin{definition}
\begin{align}
\myvec{P}(S|C) 
& := \left[\begin{array}{ll}
P(\myvec{V}(S)^{\top}\myvec{X}_{i}=0& |\omega,c_{i}=c) \\
\cdots \\
P(\myvec{V}(S)^{\top}\myvec{X}_{i}= \prod\limits_{j \in S} Q_{j}-1& |\omega,c_{i}=c) \\
\end{array}\right]
\\
\myvec{P}(S) 
& := \left[\begin{array}{ll}
P(\myvec{V}(S)^{\top}\myvec{X}_{i}=0& |\omega) \\
\cdots \\
P(\myvec{V}(S)^{\top}\myvec{X}_{i}= \prod\limits_{j \in S} Q_{j}-1& |\omega) \\
\end{array}\right]
\end{align}
\end{definition}

We can now extend \eqref{eq:poolProb} as follows. There exists a permutation matrix $\permutmat$ where:

\begin{align}
\myvec{P}(\mathbb{Z}_{J}|c) 
& = \permutmat \bigotimes\limits_{\mathscr{P} \in \text{Pooled}(\mymatrixdelta)} \myvec{P}(\mathscr{P}|c)
\end{align}

Marginalizing this equation over classes, we then get the following. Below $\myvec{1}$ refers to a vector of 1's.

\begin{align}
\myvec{P}(\mathbb{Z}_{J}) 
& = [\myvec{P}(\mathbb{Z}_{J}|c=0),\cdots,\myvec{P}(\mathbb{Z}_{J}|c=C-1)] \myvec{\pi}
\nonumber\\
& = \permutmat \sideset{}{^*}{(\bigotimes}\limits_{\mathscr{P} \in \text{Pooled}(\mymatrixdelta)} \left[\myvec{P}(\mathscr{P}|c=0),\cdots,\myvec{P}(\mathscr{P}|c=C-1)\right])\ \myvec{\pi}
\nonumber\\
& = \permutmat \sideset{}{^*}{(\bigotimes}\limits_{\mathscr{P} \in \text{Pooled}(\mymatrixdelta)} \left[\pi_{0}^{1/|\text{Pooled}(\mymatrixdelta)|}\myvec{P}(\mathscr{P}|c=0),\cdots,\pi_{C}^{1/|\text{Pooled}(\mymatrixdelta)|}\myvec{P}(\mathscr{P}|c=C-1)\right]) \myvec{1}
\label{eq:pooledKronecker}
\end{align}

Equation \eqref{eq:pooledKronecker} expresses the full distribution of $\myvec{X}_{i}|\omega$ in terms of Kronecker products of pooled domains. This product is used in our identifiability arguments in the following section.

\subsection{Identifiability}\label{appendix:identifiability2}

In this section, we show sufficient conditions for generic identifiability of a DLCM. \citet{Allman2009} provides some useful results for determining generic identifiability for traditional LCM. These are based on algebraic results from \citet{Kruskal1977}. Our proofs are inspired by Allman, but generally work directly off of Kruskal's results. The main theorem from Kruskal we leverage is discussed next.

\begin{definition}
The Kruskal rank, $\text{rank}_{\kappa} \mymatrix{M}$, is the largest number $k$ where every set of $k$ columns of $\mymatrix{M}$ are linearly independent.
\end{definition}

\begin{theorem}\label{thm:Kruskal}
[Kruskal] Let vector $\myvec{M}$ be: 
\begin{align}
\myvec{M} 
& = [\mymatrix{A}^{(0)}\otimes^{*} \mymatrix{A}^{(1)}\otimes^{*} \mymatrix{A}^{(2)}]\ \myvec{1}
\label{eq:triprod}
\\
& = [\mymatrix{B}^{(0)}\otimes^{*} \mymatrix{B}^{(1)}\otimes^{*} \mymatrix{B}^{(2)}]\ \myvec{1}
\nonumber
\end{align}
where $\mymatrix{A}^{(k)}$ and $\mymatrix{B}^{(k)}$ are $m_{k} \times m$ matrices. If:
\begin{align}
& \text{rank}_{\kappa} \mymatrix{A}^{(0)}
+ \text{rank}_{\kappa} \mymatrix{A}^{(1)}
+ \text{rank}_{\kappa} \mymatrix{A}^{(2)}
\geq 2 m + 2.
\end{align}
Then $[\mymatrix{A}^{(0)},\mymatrix{A}^{(1)},\mymatrix{A}^{(2)}]$ equals $[\mymatrix{B}^{(0)},\mymatrix{B}^{(1)},\mymatrix{B}^{(2)}]$ up to simultaneous permutation and rescaling of columns.
\end{theorem}

Note the similarity between the distribution of $\mymatrix{X}|\omega$ in \eqref{eq:pooledKronecker} and the triproduct given in \eqref{eq:triprod}. Furthermore, note that the permutation matrix $\permutmat$ in \eqref{eq:pooledKronecker} depends only on the order $\mathscr{P}$ is evaluated in the Kronecker product. Therefore, we can freely change the order of $\mathscr{P}$ terms by making corresponding changes to $\permutmat$. This gives us:

\begin{corollary}\label{thm:individualIdentifiability}
Fix a choice of $\omega$ with $\mymatrixdelta = \mymatrixdelta'$. Take a tripartition of pooled domains $S_{0} \sqcup S_{1} \sqcup S_{2} = \text{Pooled}(\mymatrixdelta')$ and define:
\begin{align}
& \mymatrix{A}^{(k)} := \sideset{}{^*}\bigotimes\limits_{\mathscr{P} \in S_{k}} \left[\pi_{0}^{1/|\text{Pooled}(\mymatrixdelta)|}\myvec{P}(\mathscr{P}|c=0),\cdots,\pi_{C-1}^{1/|\text{Pooled}(\mymatrixdelta)|}\myvec{P}(\mathscr{P}|c=C-1)\right]
\end{align}
In the special case where $S_{k}$ is empty let $\mymatrix{A}^{(k)}$ be a $1 \times 1$ matrix of all 1's. If there exists a tripartition where:
\begin{align}
\text{rank}_{\kappa} \mymatrix{A}_{0}
+ \text{rank}_{\kappa} \mymatrix{A}_{1}
+ \text{rank}_{\kappa} \mymatrix{A}_{2}
\geq 2 C + 2
\label{eq:ineqranks}
\end{align}
 then $\omega$ is has a unique distribution on $\theta' = \{\omega \in \theta:\mymatrixdelta=\mymatrixdelta'\}$.
\end{corollary}

\begin{proof}
Using Equation~\eqref{eq:pooledKronecker}, there exists a permutation matrix $\permutmat$ where:
\begin{align}
& \permutmat\ (\mymatrix{A}^{(0)} \otimes \mymatrix{A}^{(1)} \otimes \mymatrix{A}^{(2)})\ \myvec{1} = \myvec{P}(\mathbb{Z}_{j})
\label{eq:amatrix}
\end{align}
For this choice of $\mymatrixdelta'$, every possible $(\myvec{\pi},\theta)$ can be written in the above form. Therefore if we show that this decomposition is unique, we have a unique distribution.

Suppose inequality \eqref{eq:ineqranks} holds. We apply Kruskal's theorem and consider all matrices of the same dimension as $\mymatrix{A}^{(0)},\mymatrix{A}^{(1)},\mymatrix{A}^{(2)}$. Up to rescaling and permutation of columns, we know that $[\mymatrix{A}^{(0)}, \mymatrix{A}^{(1)}, \mymatrix{A}^{(2)}]$ are the only matrices which produce $(\mymatrix{A}^{(0)} \otimes \mymatrix{A}^{(1)} \otimes \mymatrix{A}^{(2)}) \myvec{1}$. Permutation of columns amounts to label switching. Rescaling of columns is disallowed because probabilities $\myvec{P}(\mathscr{P}|c)$ must sum to one.
\end{proof}

For each $\mymatrixdelta$, we will now proceed to show that Corollary~\ref{thm:individualIdentifiability} holds almost everywhere.

\begin{restatable}[]{lemma}{restatefullRank}\label{thm:fullRank}
Take a specific tripartition of pooled domains $S_{k}$ with corresponding probabilities $\mymatrix{A}^{(k)}$ as given in Corollary~\ref{thm:individualIdentifiability}. The Kruskal rank of $\mymatrix{A}^{(k)}$ is
\begin{align}
& \text{rank}_{\kappa} \mymatrix{A}^{(k)} = \min(C,\kappa_{k})\\
& \kappa_{k} := \prod_{j \in \bigcup\limits_{\mathscr{P} \in S_{k}} \mathscr{P}} Q_{j}
\end{align}
Except in a measure zero space.
\end{restatable}

\begin{proof}
Largely follows arguments given by \citet{Allman2009}.

Let $m := \min(C,\kappa_{k})$. We know that $\text{rank}_{\kappa} \mymatrix{A}^{(k)} = m$ if and only if there exists a maximal minor which is nonzero. This means there exists a $m \times m$ submatrix $\mymatrix{M}$ of  $\mymatrix{A}^{(k)}$ which has a nonzero determinant. This determinant amounts to a polynomial of the cells in $\mymatrix{M}$. Each cell of $\mymatrix{M}$ is itself a simple product of  formed by $\pi_{c}$ and corresponding values of $\theta_{c d r}$ (see \eqref{eq:amatrix}). So the minor $|\mymatrix{M}|$ is a polynomial of our parameters $\theta$ and $\myvec{\pi}$. If this polynomial is not zero everywhere, then the roots of this polynomial are a proper subset of measure zero. When we are not at a root, $\text{rank}_{\kappa} \mymatrix{A}^{(k)} = m$. Therefore if for a given $\mymatrix{\delta}$, there exists a choice of $\theta$, $\myvec{\pi}$ where $\text{rank}_{\kappa} \mymatrix{A}^{(k)} = m$ then  $\text{rank}_{\kappa} \mymatrix{A}^{(k)} = m$ almost everywhere. For more details see \citet{Allman2009}.

Take any choice of $\mymatrix{\delta}$ and $S_{k}$. Set $\pi = 1/C$. Consider $\mymatrix{A}^{(k)}$. From \eqref{eq:amatrix}, we see that each row of $\mymatrix{A}^{(k)}$ represents one pattern of items $\bigcup_{\mathscr{P} \in S_{k}} \mathscr{P}$. Specifically, each column gives a vector of conditional probabilities $\myvec{P}(\bigcup_{\mathscr{P} \in S_{k}} \mathscr{P}|c)$, up to permutation of rows. For any class we can choose $\theta_{c}$ such that the first pattern $r'$ has probability one, and the other patterns have probability zero. This is done taking each domain $d$ in class $c$ and setting $\theta_{c d r}=1$ if that pattern matches $r'$ and zero otherwise. In this way let the first pattern of class 0 have probability 1, the second pattern of class 1 have probability 1, and so on until the first $m \times m$ cells are diagonal. Clearly for this choice of $\myvec{\pi}, \theta$ we have  $\text{rank}_{\kappa} \mymatrix{A}^{(k)} = m$. It follows that $\text{rank}_{\kappa} \mymatrix{A}^{(k)} = m$ almost everywhere.
\end{proof}

\begin{corollary}\label{thm:condidentifiability}
Conditional on domain structure $\mymatrixdelta$, a DLCM is generically identifiable if there exists a tripartition of pooled domains $S_{0},S_{1},S_{2}$ where

\begin{align}
& min(\kappa_{0},C)+min(\kappa_{1},C)+min(\kappa_{2},C) \geq 2 C + 2 \\
& \kappa_{k} := \prod_{j \in \bigcup\limits_{\mathscr{P} \in S_{k}} \mathscr{P}} Q_{j}
\end{align}
\end{corollary}

\begin{proof}
By Lemma~\ref{thm:fullRank} we know that Corollary~\ref{thm:individualIdentifiability} holds almost everywhere. Therefore for a fixed choice of $\mymatrixdelta$ almost every choice of $\omega$ has a unique distribution.
\end{proof}

Corollary~\ref{thm:condidentifiability} is very specific and requires us to fix a choice of $\mymatrixdelta$. To generalize this result to unconditional $\mymatrixdelta$, we need to introduce a restriction on $\theta$. We restrict $\theta$ by enforcing as simple a structure as possible. Roughly speaking, the more domains in a DLCM the less conditional dependence exists and the simpler the model. If a domain $(c,d)$ can be split into two domains without changing the distribution of $X|\omega$ then it should be split. To formalize this we define Kronecker separability described in Definition~\ref{def:kroneckerseparable}.

We restrict $\theta_{c d}$ to not be Kronecker separable, or Kronecker inseparable. Practically, if $\theta_{c d}$ were Kronecker separable we would prefer to break domain (c,d) into several smaller domains of locally independent items. We now finish our general identifiability argument.

\begin{restatable}[]{theorem}{restatekroneckerzero}\label{thm:kroneckerzero}
The subspace where $\myvec{\theta}_{c,d}$ is Kronecker separable is measure zero with respect to the standard Lebesgue measure.
\end{restatable}

\begin{proof}
Suppose $\myvec{\theta}_{c,d}$ is Kronecker separability with conditionally independent bipartition $J_{0}\sqcup J_{1} = J(c,d)$.  Notationally for the k'th bipartition define:

\begin{itemize}
\item[ ]Pattern id: $r^{(k)} = \myvec{V}(J_{k})^{\top} \myvec{X}$ for some response pattern $\myvec{X}$
\item[ ]Pattern Probabilities: $\theta^{(k)}_{r^{(k)}} = P(\myvec{V}(J_{k})^{\top} \myvec{X}_{i} = r^{(k)}|\theta,c_{i}=c) $
\item[ ]Number of patterns: $R^{(k)} = \prod_{j \in J_{k}} Q_{j}$
\end{itemize}

Without loss of generality, relabel our items such that $\max{J_{0}} < \min{J_{1}}$. Under this relabeling, we know pattern IDs $r^{(0)},r^{(1)}$ refer to pattern ID $r^{(0)}+R^{(0)}r^{(1)}$ of domain (c,d). From independence we have that:
\begin{align*}
& \theta_{c,d,r^{(0)}+R^{(0)}r^{(1)}}
= \theta^{(1)}_{r^{(1)}}\theta^{(0)}_{r^{(0)}} \\
& \myvec{\theta}_{c,d} = \myvec{\theta}^{(0)} \otimes \myvec{\theta}^{(1)}
\end{align*}
where $\otimes$ is the kronecker product.

Before kronocker separability, $\myvec{\theta}_{c,d}$ spans the subspace of $\mathbb{R}_{>0}^{R_{c d}}$ where $\sum_{r} \myvec{\theta}_{c d r} = 1$. This gives $R_{c d}-1$ free parameters. Similarly, $\theta^{(k)}$ spans the subspace of $\mathbb{R}_{>0}^{R^{(k)}}$ where $\sum_{r^{(k)}} \theta^{(k)}_{r^{(k)}}=1$ with $R^{(k)}$ free parameters. This implies that $\myvec{\theta}_{c,d}$ is on a space of size $R_{c d}-1$ before kronecker separability, but under kronecker separability separability $\myvec{\theta}_{c,d}$ is on a space of size $R^{(0)}+R^{(1)}-2$.
\end{proof}

\begin{corollary}
Under the restriction that $\theta$ is not Kronecker separable, lemma \ref{thm:fullRank} still holds.
\end{corollary}

\begin{proof}
Let $\Theta$ be the unrestricted parameter space. Let $A$ be the event that $\text{rank}_{\kappa} \mymatrix{A}^{(k)}$ is full rank, and $B$ be the event that $\theta$ is Kronecker separable. From theorem \ref{thm:kroneckerzero} and lemma \ref{thm:fullRank} we know that $1-P(A)=P(B)=0$. It follows that $P(A|\neg B) \geq P(A \cap \neg B) \geq P(A) - P(B) = 1-0$.
\end{proof}

\begin{theorem}\label{thm:identifiability2}
[Identifiability] A DLCM is generically identifiable if the following conditions are met:

\begin{itemize}
\item Domain structure $\mymatrixdelta$ is restricted. For each allowed $\mymatrixdelta$ there must be a tripartition of items $J_{0},J_{1},J_{2}$ where all items of the same pooled domain are partitioned together and:

\begin{align}
& min(\kappa_{0},C)+min(\kappa_{1},C)+min(\kappa_{2},C) \geq 2 C + 2 \\
& \kappa_{k} := \prod_{j \in J_{k}} Q_{j}
\end{align}

\item Probabilities $\theta$ are restricted such that every $\theta_{c d}$ is Kronecker nonseperable.

\item $\pi_{c}>0$ for all $c$

\end{itemize}

\end{theorem}

\begin{proof}\qquad

\subheader{Claim} Fix domain structures where $\mymatrixdelta \neq \mymatrixdelta'$ up to relabeling of classes and domain identifiers. Disallow Kronecker separability. Parameters $\omega = (\myvec{\pi},\theta, \mymatrixdelta)$ and $\omega' = (\myvec{\pi}',\theta', \mymatrixdelta')$ produce different distributions of $\myvec{X}_{i}$.

\subheader{Proof} The domain structures $\mymatrixdelta \neq \mymatrixdelta'$ are distinct. This means there are domains $(c,d)$ from $\mymatrixdelta$ and $(c,d')$ from $\mymatrixdelta'$ where $J(c,d) \neq J(c,d')$. Without loss of generality, assume $J(c,d) \cap J'(c,d')$ is a strict subset of $J(c,d)$. Under $\omega'$ we know that items $J(c,d) \cap J'(c,d')$ and $J(c,d)-J'(c,d')$ are conditionally independent. Therefore, $X|\omega$ is the same as $X|\omega'$ only if $\myvec{\theta}_{c d}$ is Kronecker separable. Kronecker separablility is disallowed.

\subheader{Claim} Under the given restrictions, we have generic identifiability.

\subheader{Proof} Every choice of domain structure $\mymatrixdelta$ contains distinct distributions of $X|\omega$. Therefore, $\omega$ has a unique distribution in $\Theta$ if and only if it has a unique distribution conditional on $\mymatrixdelta$. 

Let $S_{\mymatrixdelta'}$ be the set of $\omega$ with $\mymatrixdelta = \mymatrixdelta'$ where $\omega$ is does not have a unique distribution. Conditional on $\mymatrixdelta$, we have generic identifiability (Theorem~\ref{thm:condidentifiability}). Therefore each $S_{\mymatrixdelta}$ is measure zero: $P(S_{\mymatrixdelta}) = P(\mymatrixdelta) P(S_{\mymatrixdelta}|\mymatrixdelta) = P(\mymatrixdelta)(0)$. Since there are finitely many $\mymatrixdelta \in \mathbb{Z}_{D}^{J \times C}$, we know the union $\bigsqcup_{\mymatrixdelta} S_{\mymatrixdelta}$ is measure zero: $P(\bigsqcup_{\mymatrixdelta} S_{\mymatrixdelta}) = \sum_{\mymatrixdelta} P(S_{\mymatrixdelta}) = 0$.
\end{proof}

\restateidentifiability*

\begin{proof}
Follows immediately from last theorem.
\end{proof}

To ensure identifiability of our DLCM model we  apply the restrictions given in Theorem~\ref{thm:identifiability}. Unless otherwise stated, these restrictions are applied both to our parameter priors and our sigma algebra. Notably the Kronecker non-separable restriction to $\myvec{\theta}_{cd}$ only removes a measure zero space (see appendix Section~\ref{appendix:posteriors}); therefore this restriction is mild.

\section{DLCM Posteriors}\label{appendix:posteriors}

\begin{remark}\label{remark:beta}
Recall the following properties of the Beta function. 

The Beta function is defined as:
\begin{align*}
\beta(\myvec{v}) = \frac{\prod_{k} \Gamma(v_{k})}{\Gamma(\sum_{k} v_{k})}
\end{align*}
Since $\sum \pi_{c} = 1$, we have the following integral following the Dirichlet distribution:
\begin{align*}
\int \prod_{c=0}^{C-1} \pi_{c}^{\alpha_{c}^{(c)}-1} d\myvec{\pi} = \beta(\myvec{\alpha}^{(c)})
\end{align*}
Furthermore let $\myvec{w} = [0,\cdots,0,1,0,\cdots,0]^{\top}$ with a value of one in the $c'$ component. Then we have the following proportion:
\begin{align*}
\frac{\beta(\myvec{v}+\myvec{w})}{\beta(\myvec{v})}
& 
= \frac{\Gamma(\sum v_{k})}{\Gamma(1+\sum v_{k})} \frac{\Gamma(v_{c'}+1) \prod_{k\neq c'} \Gamma(v_{k})}{\prod_{k} \Gamma(v_{k})}
= \frac{v_{c'}}{\sum v_{k}}
\end{align*}

\end{remark}

\restateposterior*

\begin{proof}
Our responses and priors jointly follow:
\begin{align*}
P(\myvec{c},\myvec{\pi},\theta,\mymatrix{X},\mymatrixdelta)
& = P(\myvec{\pi})P(\mymatrixdelta) \left[\prod_{c,d=0}^{C-1,D-1} P(\myvec{\theta}_{c,d}|\mymatrixdelta) \right] \prod_{i} \left[P(c_{i}|\myvec{\pi})P(\myvec{X}_{i}|c_{i},\theta,\mymatrixdelta)\right].
\end{align*}

Going forward we assume a fixed value of $\mymatrixdelta$ and omit it for simplicity. With $\mymatrixdelta$ fixed, we also use patterns $r_{i d}$ in place of responses $x_{i j}$.

The conditional distribution of class $c$ is:
\begin{align*}
P(c_{i}=c|\myvec{\pi},\theta,\myvec{r}_{i})
& =\frac{P(c_{i}=c,\myvec{r}_{i}|\myvec{\pi},\theta)}{\sum_{c'=0}^{C-1} P(c_{i}=c',\myvec{r}_{i}|\myvec{\pi},\theta)}
\\
& =\frac{P(c_{i}=c|\myvec{\pi})
P(\myvec{r_{i}}|c_{i}=c,\theta)}
{\sum_{c'=0}^{C-1} P(c_{i}=c'|\myvec{\pi}) P(\myvec{r_{i}}|c_{i}=c',\theta)}
\\
& =\frac{\pi_{c} \prod_{d=0}^{D-1}\prod_{r=0}^{R_{d}-1} \theta_{c d r}^{I(r_{i d} = r)}}
{\sum_{c'=0}^{C-1} \pi_{c'} \prod_{d=0}^{D-1}\prod_{r=0}^{R_{d}-1} \theta_{c' d r}^{I(r_{i d} = r)}}
.\\
\end{align*}

We have the following conditional distribution of $\myvec{\pi}$
\begin{align*}
P(\myvec{\pi}=\myvec{t},\myvec{c})
& = P(\pi)P(\myvec{c}|\myvec{\pi})
\\
& = \left[\frac{1}{\beta(\myvec{\alpha}^{(c)}) }\prod_{c=0}^{C-1}\pi_{c}^{\alpha^{(c)}_{c}-1}\right] \left[\prod_{i=0}^{N-1} \prod_{c=0}^{C-1} \pi_{c}^{I(c_{i}=c)}\right]
\\
& = \left[\frac{1}{\beta(\myvec{\alpha}^{(c)}) }\prod_{c=0}^{C-1}\pi_{c}^{\alpha^{(c)}_{c}-1}\right] \left[\prod_{c=0}^{C-1}\prod_{i=0}^{N-1} \pi_{c}^{I(c_{i}=c)}\right]
\\
& = \left[\frac{1}{\beta(\myvec{\alpha}^{(c)}) }\prod_{c=0}^{C-1}\pi_{c}^{\alpha^{(c)}_{c}-1}\right] \left[\prod_{c=0}^{C-1} \pi_{c}^{\sum_{i=0}^{N-1} I(c_{i}=c)}\right]
\\
& = \left[\frac{1}{\beta(\myvec{\alpha}^{(c)}) }\prod_{c=0}^{C-1}\pi_{c}^{\alpha^{(c)}_{c}-1}\right] \left[\prod_{c=0}^{C-1} \pi_{c}^{n^{(c)}_{c}}\right]
\\
& = \frac{1}{\beta(\myvec{\alpha}^{(c)})} \prod_{c=0}^{C-1}\pi_{c}^{(n^{(c)}_{c}+\alpha^{(c)}_{c})-1}
\\ 
P(\myvec{\pi}=\myvec{t}|\myvec{c},\theta,\myvec{X}_{i})
& = P(\myvec{\pi}=\myvec{t}|\myvec{c})
\\
& = \frac{P(\myvec{\pi}=\myvec{t},\myvec{c})}{\int P(\myvec{\pi},\myvec{c}) d \myvec{\pi}}
\\
& = \frac{\frac{1}{\beta(\myvec{\alpha}^{(c)})} \prod_{c=0}^{C-1}t_{c}^{(n^{(c)}_{c}+\alpha^{(c)}_{c})-1}}{\int \frac{1}{\beta(\myvec{\alpha}^{(c)})} \prod_{c=0}^{C-1}\pi_{c}^{(n^{(c)}_{c}+\alpha^{(c)}_{c})-1} d \myvec{\pi}}
\\
& = \frac{\prod_{c=0}^{C-1}t_{c}^{(n^{(c)}_{c}+\alpha^{(c)}_{c})-1}}
{\beta(\myvec{n}^{(c)}+\myvec{\alpha}^{(c)}) \int \frac{1}{\beta(\myvec{n}^{(c)}+\myvec{\alpha}^{(c)})} \prod_{c=0}^{C-1}\pi_{c}^{(n^{(c)}_{c}+\alpha^{(c)}_{c})-1} d \myvec{\pi}}
\\
& = \frac{1}
{\beta(\myvec{n}^{(c)}+\myvec{\alpha}^{(c)})} \prod_{c=0}^{C-1}t_{c}^{(n^{(c)}_{c}+\alpha^{(c)}_{c})-1}
\\
\end{align*}

The conditional distribution of $\mymatrix{\theta}_{d}$ is:
\begin{align*}
P(\mymatrix{\theta}_{d},\myvec{c},\myvec{r}_{d})
& = \left[\prod_{c=0}^{C-1} P(\myvec{\theta}_{cd})\right]\left[\prod_{i=0}^{N-1}P(r_{i d}|c_{i},\myvec{\theta}_{cd})\right]
\\
& = \left[\prod_{c=0}^{C-1} \frac{1}{\beta(\myvec{\alpha}_{d}^{(\theta)})}\prod_{r=0}^{R_{d}-1} \theta_{c d r}^{\alpha_{d r}^{(\theta)}-1} \right] 
\left[\prod_{i=0}^{N-1} \prod_{c=0}^{C-1}\prod_{r=0}^{R_{d}-1} \theta_{c d r}^{I(c_{i}=c)I(r_{i d}=r)}\right]
\\
& = \left[\prod_{c=0}^{C-1} \frac{1}{\beta(\myvec{\alpha}_{d}^{(\theta)})}\prod_{r=0}^{R_{d}-1} \theta_{c d r}^{\alpha_{d r}^{(\theta)}-1} \right] 
\left[\prod_{c=0}^{C-1}\prod_{r=0}^{R_{d}-1}\prod_{i=0}^{N-1} \theta_{c d r}^{I(c_{i}=c)I(r_{i d}=r)}\right]
\\
& = \left[\prod_{c=0}^{C-1} \frac{1}{\beta(\myvec{\alpha}_{d}^{(\theta)})}\prod_{r=0}^{R_{d}-1} \theta_{c d r}^{\alpha_{d r}^{(\theta)}-1} \right] 
\left[\prod_{c=0}^{C-1}\prod_{r=0}^{R_{d}-1} \theta_{c d r}^{\sum_{i=0}^{N-1} I(c_{i}=c)I(r_{i d}=r)}\right]
\\
& = \left[\prod_{c=0}^{C-1} \frac{1}{\beta(\myvec{\alpha}_{d}^{(\theta)})}\prod_{r=0}^{R_{d}-1} \theta_{c d r}^{\alpha_{d r}^{(\theta)}-1} \right] 
\left[\prod_{c=0}^{C-1}\prod_{r=0}^{R_{d}-1} \theta_{c d r}^{n^{(\theta)}_{c d r}}\right]
\\
& = \prod_{c=0}^{C-1} \left[ \frac{1}{\beta(\myvec{\alpha}_{d}^{(\theta)})}\prod_{r=0}^{R_{d}-1} \theta_{c d r}^{(n^{(\theta)}_{c d r} + \alpha_{d r}^{(\theta)})-1} \right]
\\
\\ 
P(\mymatrix{\theta}_{d}=\mymatrix{t}_{d}|\myvec{c},\myvec{r}_{d})
& = \frac{P(\theta_{d}=\mymatrix{t}_{d},\myvec{c},\myvec{r}_{d})}
{\int P(\theta_{d},\myvec{c},\myvec{r}_{d}) d \theta_{d}}
\\
& = \frac{\prod_{c=0}^{C-1} \left[ \frac{1}{\beta(\myvec{\alpha}_{d}^{(\theta)})}\prod_{r=0}^{R_{d}-1} t_{c d r}^{(n^{(\theta)}_{c d r} + \alpha_{d r}^{(\theta)})-1} \right]}
{\int \prod_{c=0}^{C-1} \left[ \frac{1}{\beta(\myvec{\alpha}_{d}^{(\theta)})}\prod_{r=0}^{R_{d}-1} \theta_{c d r}^{(n^{(\theta)}_{c d r} + \alpha_{d r}^{(\theta)})-1} \right] d \theta_{d}}
\\
& = \frac{\prod_{c=0}^{C-1}  \prod_{r=0}^{R_{d}-1} t_{c d r}^{(n^{(\theta)}_{c d r} + \alpha_{d r}^{(\theta)})-1} }
{ \prod_{c=0}^{C-1} \int \prod_{r=0}^{R_{d}-1} \theta_{c d r}^{(n^{(\theta)}_{c d r} + \alpha_{d r}^{(\theta)})-1}  d \myvec{\theta}_{c d}}
\\
& = \frac{\prod_{c=0}^{C-1} \left[ \frac{1}{\beta(\myvec{n}^{(\theta)}_{c d}+\myvec{\alpha}_{d}^{(\theta)})} \prod_{r=0}^{R_{d}-1} t_{c d r}^{(n^{(\theta)}_{c d r} + \alpha_{d r}^{(\theta)})-1} \right]}
{ \prod_{c=0}^{C-1} \int \frac{1}{\beta(\myvec{n}^{(\theta)}_{c d}+\myvec{\alpha}_{d}^{(\theta)})} \prod_{r=0}^{R_{d}-1} \theta_{c d r}^{(n^{(\theta)}_{c d r} + \alpha_{d r}^{(\theta)})-1}  d \myvec{\theta}_{c d}}
\\
& = \prod_{c=0}^{C-1} \left[ \frac{1}{\beta(\myvec{n}^{(\theta)}_{c d}+\myvec{\alpha}_{d}^{(\theta)})} \prod_{r=0}^{R_{d}-1} t_{c d r}^{(n^{(\theta)}_{c d r} + \alpha_{d r}^{(\theta)})-1}\right]
\\
\end{align*}

\end{proof}

\begin{theorem}
For a fixed domain structure $\mymatrixdelta$, if we collapse on $\theta$ and $\pi$ then the conditional distribution of $c_{i}$ is:
\begin{align*}
P(c_{i}=c'|\myvec{c}_{(i)},\mymatrix{r},\mymatrixdelta)
&=\frac{(n_{c',(i)}^{(c)} + \alpha_{c'}^{(c)}) \prod\limits_{d\in \mathscr{D}_{c'}} (n^{(\theta)}_{c'dr_{icd},(i)} + \alpha^{(\theta)}_{c'dr_{icd}})}{\sum_{c''} (n_{c'',(i)}^{(c)} + \alpha_{c''}^{(c)}) \prod\limits_{d\in \mathscr{D}_{c''}} (n^{(\theta)}_{c''dr_{icd},(i)} + \alpha^{(\theta)}_{c''dr_{icd}})}\\
&= \frac{P(\myvec{r}_{i}|c_{i}=c',\myvec{\pi}=E[\myvec{\pi}|\myvec{c}_{i}],\theta = E[\theta|\myvec{c}_{i},\mymatrix{r}_{(i)}])}{\sum_{c''} P(\myvec{r}_{i}|c_{i}=c'',\myvec{\pi}=E[\myvec{\pi}|\myvec{c}_{i}],\theta = E[\theta|\myvec{c}_{i},\mymatrix{r}_{(i)}])} \end{align*}

\end{theorem}

\begin{proof}
Fixed domain structure $\mymatrixdelta$ and omit it for simplicity. We start by finding the joint distribution of $P(\myvec{c},\myvec{\pi},\theta,\mymatrix{r})$:
\begin{align*}
P(\myvec{\pi}) P(\myvec{c}|\myvec{\pi})
& = \left[\frac{1}{\beta(\myvec{\alpha}_{c}^{(c)})} \prod_{c=0}^{C-1} \pi_{c}^{\alpha_{c}^{(c)}-1} \right]
\left[\prod_{i=0}^{N-1} \prod_{c=0}^{C-1} \pi_{c}^{I(c_{i}=c)}\right]\\
& = \frac{1}{\beta(\myvec{\alpha}_{c}^{(c)})} \left[\prod_{c=0}^{C-1} \pi_{c}^{n^{(c)}_{c}+\alpha_{c}^{(c)}-1} \right]\\
P(\theta)P(\mymatrix{r}|\myvec{c},\theta)
& = \left[\prod_{c,d=0}^{C-1,D-1} P(\myvec{\theta}_{c,d}) \right] \prod_{i} P(\myvec{r}_{i}|c_{i},\theta) \\
& = \left[\prod_{c=0,d=0}^{C-1,D-1} \frac{1}{\beta(\myvec{\alpha}_{d}^{(\theta)})}\prod_{r=0}^{R_{d}-1} \theta_{c d r}^{\alpha_{d r}^{(\theta)}-1} \right] 
\left[\prod_{i=0}^{N-1} \prod_{c=0,d=0}^{C-1,D-1}\prod_{r=0}^{R_{d}-1} \theta_{c d r}^{I(c_{i}=c)I(r_{i d}=r)}\right]\\
& = \left[\prod_{c=0,d=0}^{C-1,D-1} \frac{1}{\beta(\myvec{\alpha}_{c d}^{(\theta)})} \right] 
\left[ \prod_{c=0,d=0}^{C-1,D-1}\prod_{r=0}^{R_{d}-1} \theta_{c d r}^{n^{(\theta)}_{cdr}+\alpha_{c d r}^{(\theta)}-1}\right]\\
P(\myvec{c},\myvec{\pi},\theta,\mymatrix{r})
& = P(\myvec{\pi}) P(\myvec{c}|\myvec{\pi})P(\theta)P(\mymatrix{r}|\myvec{c},\theta) \\
& = \frac{1}{\beta(\myvec{\alpha}_{c}^{(c)})} \left[\prod_{c=0}^{C-1} \pi_{c}^{n^{(c)}_{c}+\alpha_{c}^{(c)}-1} \right]
\left[\prod_{c=0,d=0}^{C-1,D-1} \frac{1}{\beta(\myvec{\alpha}_{c d}^{(\theta)})} \right] 
\left[ \prod_{c=0,d=0}^{C-1,D-1}\prod_{r=0}^{R_{d}-1} \theta_{c d r}^{n^{(\theta)}_{cdr}+\alpha_{c d r}^{(\theta)}-1}\right] \\
\end{align*}

Next we collapse over $\myvec{\pi}$ and $\theta$ we get the following joint distribution of $\myvec{c},\myvec{r}$.
\begin{align*}
P(\myvec{c})
&= \int P(\myvec{\pi})P(\myvec{c}|\myvec{\pi}) d\myvec{\pi} \\
&= \frac{1}{\beta(\myvec{\alpha}_{c}^{(c)})} \int \prod_{c=0}^{C-1} \pi_{c}^{n^{(c)}_{c}+\alpha_{c}^{(c)}-1} d\myvec{\pi}\\
&= \frac{\beta(\myvec{n}^{(c)} + \myvec{\alpha}^{(c)})}{\beta(\myvec{\alpha}_{c}^{(c)})}
\\
P(\mymatrix{r}|\myvec{c})
& = \int P(\theta)P(\mymatrix{r}|\myvec{c},\theta)d\theta\\
&=\left[\prod_{c=0,d=0}^{C-1,D-1} \frac{1}{\beta(\myvec{\alpha}_{c d}^{(\theta)})} \right] 
\prod_{c=0,d=0}^{C-1,D-1}
\left[
\int
\prod_{r=0}^{R_{d}-1} \theta_{c d r}^{n^{(\theta)}_{cdr}+\alpha_{c d r}^{(\theta)}-1}
d\myvec{\theta}_{cd}
\right]\\
& =\prod_{c=0,d=0}^{C-1,D-1} \frac{\beta(\myvec{n}^{(\theta)}_{cd} + \myvec{\alpha}^{(\theta)}_{cd})}{\beta(\myvec{\alpha}_{c d}^{(\theta)})}
\\
P(\myvec{c},\myvec{r})
& =  P(\myvec{c})P(\mymatrix{r}|\myvec{c}) \\
& = \frac{\beta(\myvec{n}^{(c)} + \myvec{\alpha}^{(c)})}{\beta(\myvec{\alpha}_{c}^{(c)})} \prod_{c=0,d=0}^{C-1,D-1} \frac{\beta(\myvec{n}^{(\theta)}_{cd} + \myvec{\alpha}^{(\theta)}_{cd})}{\beta(\myvec{\alpha}_{c d}^{(\theta)})}
.\\
\end{align*}

Note that using the properties of the Beta function (e.g. Remark~\ref{remark:beta}) we can derive the following useful proportion. This will allow us to simplify a later expression. Let the subscript $(i)$ indicate that the $i$'th observation is omitted. For example $\myvec{c}_{(i)}$ indicates the class of all observations except observation $i$. Without loss of generality suppose $c_{i}=c'$. We have:

\begin{align*}
\frac{P(\myvec{c},\myvec{r})}{P(\myvec{c}_{(i)},\myvec{r}_{(i)})}
& = \left.\left[
\frac{\beta(\myvec{n}^{(c)} + \myvec{\alpha}^{(c)})}{\beta(\myvec{\alpha}_{c}^{(c)})} \prod_{c=0,d=0}^{C-1,D-1} \frac{\beta(\myvec{n}^{(\theta)}_{cd} + \myvec{\alpha}^{(\theta)}_{cd})}{\beta(\myvec{\alpha}_{c d}^{(\theta)})}
\right]
\middle/
\left[
\frac{\beta(\myvec{n}_{(i)}^{(c)} + \myvec{\alpha}^{(c)})}{\beta(\myvec{\alpha}_{c}^{(c)})} \prod_{c=0,d=0}^{C-1,D-1} \frac{\beta(\myvec{n}^{(\theta)}_{cd,(i)} + \myvec{\alpha}^{(\theta)}_{cd})}{\beta(\myvec{\alpha}_{c d}^{(\theta)})}
\right]\right.\\
& = 
\frac{\beta(\myvec{n}^{(c)} + \myvec{\alpha}^{(c)})}{\beta(\myvec{n}_{(i)}^{(c)} + \myvec{\alpha}^{(c)})} \prod_{c=0,d=0}^{C-1,D-1} \frac{\beta(\myvec{n}^{(\theta)}_{cd} + \myvec{\alpha}^{(\theta)}_{cd})}{\beta(\myvec{n}^{(\theta)}_{cd,(i)} + \myvec{\alpha}^{(\theta)}_{cd})}
\\
& = 
\frac{n_{c',(i)}^{(c)} + \alpha_{c'}^{(c)}}{||\myvec{n}_{(i)}^{(c)} + \myvec{\alpha}^{(c)}||_{L1}} \prod_{d\in \mathscr{D}_{c'}} \frac{n^{(\theta)}_{c'dr_{ic'd},(i)} + \alpha^{(\theta)}_{c'dr_{ic'd}}}{||\myvec{n}^{(\theta)}_{c'd,(i)} + \myvec{\alpha}^{(\theta)}_{c'd}||_{L1}}
\label{eq:pprop1}\numberthis
\\
& = 
E[\pi_{c'}|\myvec{c}_{(i)}] \prod_{d\in \mathscr{D}_{c'}} E[\theta_{c'dr_{ic'd}}|\mymatrix{r}_{(i)},\myvec{c}_{(i)}]
\label{eq:pprop2}\numberthis
\\
& = 
P(\myvec{r}_{i}|c_{i}=c',\myvec{\pi}=E[\myvec{\pi}|\myvec{c}_{i}],\theta = E[\theta|\myvec{c}_{i},\mymatrix{r}_{(i)}])
\label{eq:pprop3}\numberthis
\\
\end{align*}

We now consider the distribution of $c_{i}|\myvec{c}_{(i)},\myvec{r}_{i}$, where $\myvec{c}_{(i)}$ indicates the class of all observations except observation $i$. The final three expressions below follow from substitutions of \eqref{eq:pprop1}, \eqref{eq:pprop2}, and \eqref{eq:pprop3} respectively,

\begin{align*}
P(c_{i}=c'|\myvec{c}_{(i)},\mymatrix{r})
&=\frac{P(c_{i}=c',\myvec{c}_{(i)},\mymatrix{r})}{\sum_{c''} P(c_{i}=c'',\myvec{c}_{(i)},\mymatrix{r})}\\
&=\frac{P(c_{i}=c',\myvec{c}_{(i)},\mymatrix{r})/P(\myvec{c}_{(i)},\myvec{r}_{(i)})}{\sum_{c''} P(c_{i}=c'',\myvec{c}_{(i)},\mymatrix{r})/P(\myvec{c}_{(i)},\myvec{r}_{(i)})}\\
&=\frac{(n_{c',(i)}^{(c)} + \alpha_{c'}^{(c)}) \prod\limits_{d\in \mathscr{D}_{c'}} (n^{(\theta)}_{c'dr_{icd},(i)} + \alpha^{(\theta)}_{c'dr_{icd}})}{\sum_{c''} (n_{c'',(i)}^{(c)} + \alpha_{c''}^{(c)}) \prod\limits_{d\in \mathscr{D}_{c''}} (n^{(\theta)}_{c''dr_{icd},(i)} + \alpha^{(\theta)}_{c''dr_{icd}})} 
\\
&=\frac{E[\pi_{c'}|\myvec{c}_{(i)}] \prod_{d\in \mathscr{D}_{c'}} E[\theta_{c'dr_{ic'd}}|\mymatrix{r}_{(i)},\myvec{c}_{(i)}]}{\sum_{c''} E[\pi_{c''}|\myvec{c}_{(i)}] \prod_{d\in \mathscr{D}_{c''}} E[\theta_{c''dr_{ic''d}}|\mymatrix{r}_{(i)},\myvec{c}_{(i)}]} 
\\
& = \frac{P(\myvec{r}_{i}|c_{i}=c',\myvec{\pi}=E[\myvec{\pi}|\myvec{c}_{i}],\theta = E[\theta|\myvec{c}_{i},\mymatrix{r}_{(i)}])}{\sum_{c''} P(\myvec{r}_{i}|c_{i}=c'',\myvec{\pi}=E[\myvec{\pi}|\myvec{c}_{i}],\theta = E[\theta|\myvec{c}_{i},\mymatrix{r}_{(i)}])} 
\\
\end{align*}

\end{proof}

\begin{remark}
For traditional LCM the priors are:
\begin{gather}
\text{Priors:} \nonumber\\
c_{i}|\myvec{\pi} \sim Cat(\myvec{\pi}) \label{eq:tlcm.cPrior}\\
\myvec{\pi} \sim Dirichlet(\myvec{\alpha}^{(c)}) \label{eq:tlcm.piPrior}\\
\myvec{\theta}_{cj} \sim Dirichlet(\myvec{\alpha}^{(\theta)}_{j}) \label{eq:tlcm.rhoPrior}
\end{gather}
and the full conditional distributions are:
\begin{align}
& \text{Posteriors:}& \nonumber\\
& P(c_{i}=c|\myvec{\pi},\mymatrix{\theta},\myvec{X}_{i}) = \frac{\pi_{c} P(\myvec{X}_{i}|\mymatrix{\theta},C_{i}=c)}{\sum\limits_{c'=0}^{C-1} \pi_{c'} P(\myvec{X}_{i}|\mymatrix{\theta},C_{i}=c')} 
\label{eq:tlcm.cPosterior}\\
& \myvec{\pi}|\myvec{c},\mymatrix{X} \sim Dirichlet(\myvec{\alpha}^{(c)}+ \myvec{n}^{(c)}) 
\label{eq:tlcm.piPosterior}\\
& \myvec{\theta}_{cj}|\myvec{c},\mymatrix{X} \sim Dirichlet(\myvec{\alpha}^{(\theta)}_{j}+\myvec{n}^{(\theta)}_{c j}) 
\label{eq:tlcm.rhoPosterior}\\
\nonumber\\
& n^{(c)}_{c} := \sum_{i=0}^{n-1} I(c_{i}=c) \\
& n^{(\theta)}_{cjq} := \sum_{i=0}^{n-1} I(X_{ij}=q,c_{i}=c)
\end{align}
These are well studied. For instance see \citet{Li2018} for a recent overview.
\end{remark}

\begin{theorem}\label{thm:marginalTheta}
Fix a domain $(c,d)$. Suppose there are $m$ subjects which belong to class $c$. Without loss of generality assume subjects $i \in \{0,\cdots,m-1\}$ all belong to class $c$. Let $r_{icd}$ represent the response pattern given by by subject $i$ to domain $(c,d)$. Let $\myvec{r}_{cd}=[r_{i=0,c,d},\cdots,r_{m-1,c,d}]^{\top}$ be the vector of all responses to this domain. Marginalized over $\theta$, the probability of $\myvec{r}_{cd}$ is given by:
\begin{equation}
P(\myvec{r}_{cd}=\myvec{r}|\myvec{c},\mymatrixdelta)
= \frac{\beta(\alpha^{(\theta)}\myvec{1}+\myvec{n}_{c d})}{\beta(\alpha^{(\theta)}\myvec{1})}
\end{equation}
with the following definitions:
\begin{align*}
\beta(\myvec{\alpha}) & := \frac{\prod_{k} \Gamma(\alpha_{k})}{\Gamma(\sum_{k} \alpha_{k})} \\
n^{(\theta)}_{c d r} & := \sum_{i=0}^{n-1} I(r_{i c d} = r, c_{i}=c) \\
n^{(\theta)}_{c d} & := [n^{(\theta)}_{c d 0},\ n^{(\theta)}_{c d 1},\ \cdots,\ n^{(\theta)}_{c,d,R_{c d}-1}]^{\top} \\
\end{align*}
\end{theorem}

\begin{proof}
\begin{align*}
P(\myvec{r}_{cd}=\myvec{r}|\myvec{c},\mymatrixdelta)
& = \int P(\myvec{\theta}_{cd}|\mymatrixdelta)P(\myvec{r}_{cd}=\myvec{r}|\myvec{c},\myvec{\theta}_{cd}) d\myvec{\theta}_{cd}
\\
& = \int 
\left[\frac{1}{\beta(\alpha^{(\theta)}\myvec{1})} \prod_{r'} \theta_{c d r'}^{\alpha^{(\theta)}-1}\right]
\left[
\prod_{i} \prod_{r'} \theta_{c d r'}^{I(r_{i c d}=r')}\right]
 d\myvec{\theta}_{cd}
\\
& = \frac{1}{\beta(\alpha^{(\theta)}\myvec{1})} \int 
\prod_{r'} \theta_{c d r'}^{\alpha^{(\theta)}+n^{(\theta)}_{c d r'}-1}
 d\myvec{\theta}_{cd}
\\
& = \frac{\beta(\alpha^{(\theta)}\myvec{1}+\myvec{n}_{c d})}{\beta(\alpha^{(\theta)}\myvec{1})} \int 
\frac{1}{\beta(\alpha^{(\theta)}\myvec{1}+\myvec{n}_{c d})}
\prod_{r'} \theta_{c d r'}^{\alpha^{(\theta)}+n^{(\theta)}_{c d r'}-1}
 d\myvec{\theta}_{cd}
\\
& = \frac{\beta(\alpha^{(\theta)}\myvec{1}+\myvec{n}_{c d})}{\beta(\alpha^{(\theta)}\myvec{1})}
\\
\end{align*}

\end{proof}

\section{Domain Prior}\label{appendix:domainPrior}

\begin{theorem}\label{thm:domainlabel}
Domain identifiability and $\mathsf{MaxItems}$ restrictions are a function of $\{J(c,d): c \in \mathbb{Z}_{C},d\in\mathbb{Z}_{D}\}$ and do not depend on the specific labeling of $\mymatrix{\delta}$.
\end{theorem}

\begin{proof}

From theorem \ref{thm:identifiability} we know that $\mymatrix{\delta}$ identifiability restrictions are based on pooled domains. From theorem \ref{thm:pooledconstruction}, we know that pooled domains are constructed based on items $J(c,d)$ without regard to the specific labeling of $(c,d)$. In this way identifiability restrictions do not depend on labeling of $\mymatrix{\delta}$.

Similarly $\mathsf{MaxItems}$ is a restriction on $|J(c,d)|$ without consideration of the label $(c,d)$.

\end{proof}

\restatedomainPrior*

\begin{proof}\qquad
\begin{enumerate}
\item Theorem's first equation: By definition every allowed choice of $\mymatrix{\delta}$ is equally likely.
\item Theorem's second equation: There are $_{D}\mathscr{P}_{|\mathscr{D}_{c}|} = D! / (D-|\mathscr{D}_{c}|)!$ possible ways to permute our $|\mathscr{D}_{c}|$ groups on $D$ available domains. Each permutation fully specifies $\myvec{\delta}_{c}$ which has $D^{J}$ possibilities.
\item Theorem's third equation: Fix some choice of $\{|J(c,d)|:d \in \mathbb{Z}_{D}\}$. There are $\binom{J}{|J(c,0)|,\cdots,|J(c,D-1)|}$ ways we can split our $J$ items into groups of the size specified. This assumes each group is distinct. Our groups are not distinct and two groups of the same size can be swapped. As a result we must divide by $\prod_{k=1}^{J} |\{d:|J(c,d)|=k\}|!$. This results in $J! \big/ [\prod_{k=0}^{D-1} |J(c,d)|!\ \prod_{k=1}^{J} |\{d:|J(c,d)|=k\}|!]$ ways $\{J(c,d):d \in \mathbb{Z}_{D}\}$ can produce $\{|J(c,d)|:d \in \mathbb{Z}_{D}\}$. Each of these have probabilities as defined in the theorem's second equation.

\end{enumerate}
\end{proof}

\restatedomainPriorIneq*

\begin{proof} For some constant $a$:
\begin{align*}
P(\{|J(c,d)|:d \in \mathscr{D}_{c}\} = \{1,1,\cdots,1\}) 
& = a \frac{J!}{\prod_{k=0}^{D-1} |J(c,d)|!
\ \prod_{k=1}^{J} |\{d:|J(c,d)|=k\}|!}
\ 
\frac{D!}{(D-|\mathscr{D}_{c}|)!
\ D^{J}}
\\
& = a \frac{J!}{(1)
J!}
\ 
\frac{D!}{(D-J)!
\ D^{J}}
\\
P(\{|J(c,d)|:d \in \mathscr{D}_{c}\} = \{2,1,1,\cdots,1\}) 
& = a \frac{J!}{(2)
(J-2)!}
\ 
\frac{D!}{(D-(J-1))!
\ D^{J}}
\\
\frac{P(\{|J(c,d)|:d \in \mathscr{D}_{c}\} = \{1,1,\cdots,1\})}{P(\{|J(c,d)|:d \in \mathscr{D}_{c}\} = \{2,1,1,\cdots,1\})}
& = \frac{2(D-(J-1))}{J(J-1)} \geq q
\\
& D 
\geq \frac{q}{2} J(J-1) + (J-1)
\\
\end{align*}
\end{proof}

\section{MCMC}\label{appendix:mcmc}

Suppose we are proposing a change to domains $(c,d_{1})$ and $(c,d_{2})$. Let $p_{f}$ be the \stress{forward} probability of proposing $\mymatrix{\delta}^{(t)}\rightarrow \mymatrix{\delta}'$, and $p_{b}$ be the \stress{backward} probability of proposing $\mymatrix{\delta}' \rightarrow \mymatrix{\delta}^{(t)}$.

\begin{lemma}\label{thm:proposaldomains} Let $g(c,d_{1},d_{2}|\mymatrix{\delta})$ be the probability of choosing first domain $(c,d_{1})$ and then domain $(c,d_{2})$. 
\begin{enumerate}
\item The probability of choosing nonempty domain $(c,d_{1})$ first and then nonempty domain $(c,d_{2})$ is:
\begin{align}
g(c,d_{1},d_{2}|\mymatrix{\delta}) = & \frac{1 - p_{\text{\tiny empty}} I(|J(c,d_{1})| > 1)}{|\mathscr{D}_{c}|(|\mathscr{D}_{c}|-1)}
\label{eq:proposal1then2}
\end{align}

\item The probability of choosing two nonempty domains $(c,d_{1})$ and $(c,d_{2})$, in either order, is:
\begin{align}
g(c,d_{1},d_{2}|\mymatrix{\delta}) + g(c,d_{2},d_{1}|\mymatrix{\delta}) = & \frac{2 - p_{\text{\tiny empty}} I(|J(c,d_{1})| > 1) +  p_{\text{\tiny empty}} I(|J(c,d_{2})| > 1)}{|\mathscr{D}_{c}|(|\mathscr{D}_{c}|-1)}
\label{eq:proposal1or2}
\end{align}

\item The probability of choosing nonempty domain $(c,d_{1})$ and then the first empty domain is:

\begin{align}
g(c,d_{1},d_{2}|\mymatrix{\delta}) = & \frac{p_{\text{\tiny empty}}}{|\mathscr{D}_{c}|}
\label{eq:proposal1empty}
\end{align}

\end{enumerate}

\end{lemma}

\begin{proof} Simple probability exercise.

\subheader{Claim 1} The first domain is always nonempty. The probability of choosing $(c,d_{1})$ is $1/|\mathscr{D}_{c}|$. If $J(c,d_{1})$ only has a single element, then the second domain will always be nonempty. If $J(c,d_{1})$ has multiple elements then the second domain is nonempty with probability $1 - p_{\text{\tiny empty}}$. Therefore, the probability that the second chosen domain is nonempty is $1 - p_{\text{\tiny empty}} I(|J(c,d_{1})| > 1)$. If the second domain is nonempty, the probability of choosing $(c,d_{2})$ is $1/(|\mathscr{D}_{c}|-1)$. Therefore the probability of choosing first $(c,d_{1})$ and then $(c,d_{2})$ is \eqref{eq:proposal1then2}.

\subheader{Claim 2} Apply claim 1 twice and sum. First calculate the probability of $(c,d_{1})$ then $(c,d_{2})$. Next calculate the probability of $(c,d_{2})$ then $(c,d_{1})$. These events are disjoint and their probability can be summed.

\subheader{Claim 3} The probability of choosing $(c,d_{1})$ is $1/|\mathscr{D}_{c}|$. If $J(c,d_{1})$ has two or more items, the probability of choosing an empty domain is $p_{\text{\tiny empty}}$. If we choose an empty domain, we always pick the first empty domain. Equation \eqref{eq:proposal1empty} follows.

\end{proof}

\begin{lemma}\label{thm:proposalratiog}
Suppose we are proposing a change to domains $(c,d_{1})$ and $(c,d_{2})$. The ratio of proposal probabilities $p_{f}/p_{b}$ depends only on $g(\cdot)$ as defined in lemma~\ref{thm:proposaldomains}. That is:

\begin{align*}
\frac{p_{f}}{p_{b}}
& = \frac{g(c,d_{1},d_{2}|\mymatrix{\delta}^{(t)})+g(c,d_{2},d_{1}|\mymatrix{\delta}^{(t)})}{g(c,d_{1},d_{2}|\mymatrix{\delta}')+g(c,d_{2},d_{1}|\mymatrix{\delta}')}
\end{align*}

\end{lemma}

\begin{proof}
Take any candidate proposal. We assume this proposal is valid. The forward and backward probabilities are given by:
\begin{align*}
p_{f} & = [g(c,d_{1},d_{2}|\mymatrix{\delta}^{(t)})+g(c,d_{2},d_{1}|\mymatrix{\delta}^{(t)})] a_{f} \\
p_{b} & = [g(c,d_{1},d_{2}|\mymatrix{\delta}')+g(c,d_{2},d_{1}|\mymatrix{\delta}')] a_{b}
\end{align*}
where $a_{f}$ is the probability of partitioning the items $J(c,d_{1})\cup J(c,d_{2})$ as proposed. Conversely $a_{b}$ is the probability of partitioning $J'(c,d_{1})\cup J'(c,d_{2}) = J(c,d_{1})\cup J(c,d_{2})$ back into its original form. By construction every valid partition is equally likely. It follows that $a_{b}=a_{f}$. The result immediately follows.
\end{proof}

\begin{theorem}
[Proposal Probabilities]\qquad

\begin{enumerate}
\item If $J^{(t)}(c,d_{k})$ and $J'(c,d_{k})$ are each nonempty, then:

\begin{align}
\frac{p_{f}}{p_{b}} & = \frac{2 - p_{\text{\tiny empty}} I(|J(c,d_{1})| > 1) +  p_{\text{\tiny empty}} I(|J(c,d_{2})| > 1) }{2 - p_{\text{\tiny empty}} I(|J'(c,d_{1})| > 1) +  p_{\text{\tiny empty}} I(|J'(c,d_{2})| > 1) }
\end{align}

\item If $J^{(t)}(c,d_{k})$ are both nonempty, and one of $J'(c,d_{k})$ is empty, then we have the following proposal probabilities:

\begin{align}
\frac{p_{f}}{p_{b}} 
& = \frac{2 - p_{\text{\tiny empty}} I(|J(c,d_{1})| > 1) +  p_{\text{\tiny empty}} I(|J(c,d_{2})| > 1) }{p_{\text{\tiny empty}} |\mathscr{D}_{c}|}
\end{align}

\item If $J^{(t)}(c,d_{2})$ is empty, then we have the following proposal probabilities:

\begin{align}
\frac{p_{f}}{p_{b}} & = \frac{p_{\text{\tiny empty}}(|\mathscr{D}_{c}|+1)}{2 - p_{\text{\tiny empty}} I(|J'(c,d_{1})| > 1) +  p_{\text{\tiny empty}} I(|J'(c,d_{2})| > 1)}
\end{align}
\end{enumerate}

\end{theorem}

\begin{proof}
Follows by substituting the probabilities of lemma~\ref{thm:proposaldomains} into lemma~\ref{thm:proposalratiog}.
\end{proof}

\section{Simulation Studies}\label{appendix:sims}

There are three scenarios we consider Traditional Data, Homogeneous Data, and Heterogeneous Data. Within each of the three scenarios, we generate data of different sample sizes. We generate 100 datasets for each sample size $n\in \{100,200,300,400,500,1000\}$. For the largest sample size $n=1000$ we fit 35 different models to each of the 100 datasets. We fit traditional LCMs, homogeneous DLCMs, and heterogeneous DLCMs each with $C \in \{1,\cdots,5\}$ classes. For the DLCM models, we fit with three different priors: a bucket, pattern adjusted, and uniform. The uniform prior assumes that every domain structure is equally likely, and serves as a baseline comparison similar to \cite{Marbac2014}. For smaller sample sizes we fit with the same models, but restrict to $C=2$ classes.

When fitting any of these models we use the following specifications. Initial class membership is done by choosing $C$ random centers, and classifying each subject to the closest center based on $L_{1}$ distance. The first center is chosen at random from the unique rows of $\mymatrix{X}$. Subsequent centers are chosen one at a time to be the farthest point from previously chosen centers. This is an attempt to overdisperse the starting classes. We fit with a single MCMC chain with 1,000 warmup iterations followed by 5,000 main iterations. The hyperparameters we use are: $\myvec{\alpha}^{(c)}=1$, $\alpha^{(\theta)}=1$, $D=J^{2}-1$, $p_{\text{\tiny empty}} = 0.3$, $\mathsf{nDomainIters}=J$, $\mathsf{MaxItems}=10$. For heterogeneous DLCM we set $\mathsf{nHomoItrs}=300$. To improve runtime, we do not collapse on $\theta$ and $\myvec{\pi}$ when updating $c_{i}$.

Let $\omega^{(s,t)}$ be the parameter values from the $t$'th MCMC iteration of the $s$'th simulated dataset. Let $\hat{\mymatrixdelta}^{(s)}$ be the most common posterior domain structure in the $s$'th simulation. We consider the following metrics for domain accuracy:

\begin{itemize}
\item \subheader{Mode Accuracy} In what percent of simulated datasets did the mode posterior domain structure $\hat{\mymatrixdelta}^{(s)}$ equal the true domain structure?
\item \subheader{Within-Chain Accuracy} Within a given simulation, what percent of MCMC iterations did $\mymatrixdelta^{(s,t)}$ equaled the true domain structure?
\item \subheader{Within Chain Concentration} Within a given simulation, build the following ratio. Divide the number of iterations where $\mymatrixdelta^{(s,t)}$ equaled the true domain structure by the number of iterations for the overall second most common domain structure. If this ratio is $\gg 1$ then the true domain structure is much more common than the next most frequent domain structure.
\end{itemize}

We also have the following metrics for model fit:

\begin{itemize}
\item \subheader{LPPD} The log pointwise predictive density (LPPD) measures the predictive accuracy of the model (\cite{Gelman2013}). This serves a similar purpose to log likelihood in frequentist models. LPPD of the $s$'th simulation is:

\begin{align}
& \text{LPPD}_{s} = \sum_{i=1}^{n} \ln \left(\frac{1}{\text{nItr}} \sum_{t=1}^{\text{nItr}} P(\myvec{X}_{i}|\omega^{(s,t)})\right)
\end{align}

\item \subheader{WAIC Penalty} The Watanabe-Akaike information criterion (WAIC) penalty measures model complexity (\cite{Gelman2013}). The WAIC penalty for the s'th simulation is:

\begin{align}
\text{WAIC Penalty}_{s} 
&= 2 \sum_{i=1}^{n}  \left[\ln \left(\frac{1}{\text{nItr}} \sum_{t=1}^{\text{nItr}} P(\myvec{X}_{i}|\omega^{(s,t)})\right)\right.
\nonumber\\
& - \left.\frac{1}{\text{nItr}} \sum_{t=1}^{\text{nItr}} \ln P(\myvec{X}_{i}|\omega^{(s,t)})\right]
\end{align}

\item \subheader{WAIC} The Watanabe-Akaike information criterion (WAIC) is a measure of goodness of fit. It starts with a measure of model accuracy then penalizes model complexity.  The WAIC can be thought of as an approximation of leave one out cross validation (\cite{Gelman2013}). The lower the WAIC the better. For the $s$'th simulation, the WAIC is:

\begin{align}
& \text{WAIC}_{s} = -2 \text{LPPD}_{s} + 2 \text{WAIC Penalty}_{s}
\end{align}

\item \subheader{Times chosen as min WAIC} For each simulation, identify which fitted model optimizes (minimizes) WAIC. In how many simulations was this model chosen as best?

\end{itemize}

\FloatBarrier
\subsection{Simulations: Domain Accuracy}

\begin{table}[!htbp]
\begin{tabular}{lllllllll}
\multicolumn{3}{l}{Domain Within-Chain   Accuracy}           & \multicolumn{6}{l}{\makecell[l]{\\Sample Size}}            \\
Data               & Model              & Prior       & $n=100$  & 200  & 300  & 400   & 500   & 1000  \\\hline
Traditional Data   & Homogeneous DLCM   & Uniform     & 0\%  & 0\%  & 0\%  & 0\%   & 0\%   & 0\%   \\
Traditional Data   & Homogeneous DLCM   & Bucket  & 69\% & 77\% & 83\% & 87\%  & 88\%  & 94\%  \\
Traditional Data   & Homogeneous DLCM   & Pattern Adjusted & 99\% & 99\% & 99\% & 100\% & 100\% & 100\% \\
Traditional Data   & Heterogeneous DLCM & Uniform     & 0\%  & 0\%  & 0\%  & 0\%   & 0\%   & 0\%   \\
Traditional Data   & Heterogeneous DLCM & Bucket  & 42\% & 49\% & 55\% & 57\%  & 61\%  & 70\%  \\
Traditional Data   & Heterogeneous DLCM & Pattern Adjusted & 85\% & 87\% & 90\% & 91\%  & 92\%  & 94\%  \\
\\[-3ex]
Homogeneous Data   & Homogeneous DLCM   & Uniform     & 0\%  & 0\%  & 0\%  & 0\%   & 0\%   & 0\%   \\
Homogeneous Data   & Homogeneous DLCM   & Bucket  & 80\% & 88\% & 89\% & 90\%  & 91\%  & 94\%  \\
Homogeneous Data   & Homogeneous DLCM   & Pattern Adjusted & 37\% & 99\% & 96\% & 97\%  & 99\%  & 98\%  \\
Homogeneous Data   & Heterogeneous DLCM & Uniform     & 0\%  & 0\%  & 0\%  & 0\%   & 0\%   & 0\%   \\
Homogeneous Data   & Heterogeneous DLCM & Bucket  & 5\%  & 35\% & 62\% & 71\%  & 72\%  & 81\%  \\
Homogeneous Data   & Heterogeneous DLCM & Pattern Adjusted & 0\%  & 7\%  & 36\% & 73\%  & 89\%  & 95\%  \\
\\[-3ex]
Heterogeneous Data & Heterogeneous DLCM & Uniform     & 0\%  & 0\%  & 0\%  & 0\%   & 0\%   & 0\%   \\
Heterogeneous Data & Heterogeneous DLCM & Bucket  & 42\% & 66\% & 70\% & 73\%  & 72\%  & 80\%  \\
Heterogeneous Data & Heterogeneous DLCM & Pattern Adjusted & 7\%  & 93\% & 94\% & 95\%  & 94\%  & 96\% 
\end{tabular}
\caption{From simulation studies with $C=2$ classes. In what percent of MCMC iterations did the posterior domain structure match the true domain structure? Averaged across simulations.}
\label{table:simWithinChainAccuracy}
\end{table}

\begin{table}[!htbp]
\begin{tabular}{lllrrrrrr}
\multicolumn{3}{l}{Domain Within-Chain Concentration}           & \multicolumn{6}{l}{\makecell[l]{\\Sample Size}}            \\
Data               & Model              & Prior       & n=100 & 200   & 300   & 400   & 500   & 1000  \\\hline
Traditional Data   & Homogeneous DLCM   & Bucket  & 19.3  & 27.4  & 38.0  & 49.0   & 49.5   & 95.9   \\
Traditional Data   & Homogeneous DLCM   & Pattern Adjusted & 411.8 & 496.7 & 711.4 & 831.3  & 711.9  & 998.2  \\
Traditional Data   & Heterogeneous DLCM & Bucket  & 14.4  & 14.3  & 17.5  & 18.6   & 20.5   & 24.3   \\
Traditional Data   & Heterogeneous DLCM & Pattern Adjusted & 73.7  & 85.9  & 102.4 & 115.4  & 103.0  & 146.5  \\
\\[-3ex]
Homogeneous Data   & Homogeneous DLCM   & Bucket  & 34.3  & 46.4  & 55.3  & 68.0   & 102.1  & 180.3  \\
Homogeneous Data   & Homogeneous DLCM   & Pattern Adjusted & 0.3   & 997.3 & 914.5 & 1247.0 & 1456.0 & 4997.0 \\
Homogeneous Data   & Heterogeneous DLCM & Bucket  & 0.0   & 1.1   & 18.5  & 24.7   & 31.8   & 45.3   \\
Homogeneous Data   & Heterogeneous DLCM & Pattern Adjusted & 0.0   & 0.0   & 0.3   & 23.2   & 103.3  & 301.4  \\
\\[-3ex]
Heterogeneous Data & Heterogeneous DLCM & Bucket  & 6.4   & 23.2  & 27.8  & 24.4   & 28.6   & 31.4   \\
Heterogeneous Data & Heterogeneous DLCM & Pattern Adjusted & 0.0   & 155.6 & 172.5 & 154.1  & 159.4  & 211.0 
\end{tabular}
\caption{From simulation studies with $C=2$ classes. Ratio of the number of MCMC iterations on the correct domain structure divided by the number of iterations for the most common alternative. Median across simulations.}
\label{table:simDomainRatio}
\end{table}

\FloatBarrier
\subsection{Simulations: Goodness of Fit}

In this subsection, we examine how well each model fit the data using WAIC. Note Table~\ref{table:simWAICn}. For Traditional Data, notice that the goodness of fit is roughly equal for Traditional LCMs, Homogeneous DLCMs, and Heterogeneous DLCMs. This indicates that our more complicated models still perform well when Traditional LCM assumptions are satisfied. Note Tables \ref{table:simsWaicBest1} and \ref{table:simsWaicBest2}. For Traditional Data, the best model is split between Traditional LCM, Homogeneous DLCMS, and Heterogeneous DLCMs. For Homogeneous Data, the best model is split between Homogeneous DLCMS and Heterogeneous DLCMs. For Heterogeneous Data, the best models are all Heterogeneous DLCMs. This indicates that among models where assumptions are satisfied that the goodness of fit is often comparable between models.

\begin{table}[!htbp]
\begin{tabular}{lllrrrrr}
\multicolumn{3}{l}{WAIC}           & \multicolumn{5}{l}{\makecell[l]{\\Number of Classes}}            \\
Data               & Model              & Prior       & C=1 & 2      & 3      & 4      & 5   \\\hline
Traditional Data   & Traditional LCM    &             & 29,476   & 26,900 & 26,947 & 26,968 & 26,982 \\
Traditional Data   & Homogeneous DLCM   & niave       & 27,764   & 26,910 & 26,949 & 26,965 & 26,976 \\
Traditional Data   & Homogeneous DLCM   & Bucket  & 27,812   & 26,900 & 26,946 & 26,967 & 26,982 \\
Traditional Data   & Homogeneous DLCM   & Pattern Adjusted & 28,245   & 26,900 & 26,947 & 26,968 & 26,982 \\
Traditional Data   & Heterogeneous DLCM & niave       & 27,770   & 26,917 & 26,938 & 26,943 &        \\
Traditional Data   & Heterogeneous DLCM & Bucket  & 27,818   & 26,900 & 26,947 & 26,967 & 26,981 \\
Traditional Data   & Heterogeneous DLCM & Pattern Adjusted & 28,248   & 26,900 & 26,947 & 26,967 & 26,981 \\
\\[-3ex]
Homogeneous Data   & Traditional LCM    &             & 31,749   & 29,665 & 29,061 & 28,651 & 28,441 \\
Homogeneous Data   & Homogeneous DLCM   & niave       & 27,991   & 27,238 & 27,260 & 27,273 & 27,294 \\
Homogeneous Data   & Homogeneous DLCM   & Bucket  & 28,018   & 27,247 & 27,256 & 27,269 & 27,294 \\
Homogeneous Data   & Homogeneous DLCM   & Pattern Adjusted & 28,288   & 27,244 & 27,257 & 27,270 & 27,302 \\
Homogeneous Data   & Heterogeneous DLCM & niave       & 27,991   & 27,236 & 27,251 & 27,258 &        \\
Homogeneous Data   & Heterogeneous DLCM & Bucket  & 28,014   & 27,238 & 27,262 & 27,274 & 27,299 \\
Homogeneous Data   & Heterogeneous DLCM & Pattern Adjusted & 28,288   & 27,244 & 27,262 & 27,274 & 27,303 \\
\\[-3ex]
Heterogeneous Data & Traditional LCM    &             & 30,783   & 28,339 & 28,041 & 27,792 & 27,604 \\
Heterogeneous Data & Homogeneous DLCM   & niave       & 27,762   & 26,481 & 26,492 & 26,507 & 26,524 \\
Heterogeneous Data & Homogeneous DLCM   & Bucket  & 27,783   & 26,476 & 26,492 & 26,512 & 26,531 \\
Heterogeneous Data & Homogeneous DLCM   & Pattern Adjusted & 28,072   & 26,795 & 26,558 & 26,578 & 26,558 \\
Heterogeneous Data & Heterogeneous DLCM & niave       & 27,753   & 26,455 & 26,472 & 26,477 &        \\
Heterogeneous Data & Heterogeneous DLCM & Bucket  & 27,786   & 26,444 & 26,477 & 26,503 & 26,523 \\
Heterogeneous Data & Heterogeneous DLCM & Pattern Adjusted & 28,070   & 26,444 & 26,474 & 26,503 & 26,531   
\end{tabular}
\caption{From simulation studies with sample size $n=1000$. WAIC is averaged across simulations.}
\label{table:simWAICnclass}
\end{table}

\begin{table}[!htbp]
\begin{tabular}{lllrrrrrr}
\multicolumn{3}{l}{WAIC}           & \multicolumn{6}{l}{\makecell[l]{\\Sample Size}}            \\
Data               & Model                & Prior        & n=100    & 200      & 300      & 400      & 500      & 1000     \\\hline
Traditional Data   & Traditional LCM & & 2,738 & 5,410 & 8,091 & 10,787 & 13,472 & 26,900 \\
Traditional Data   & Homogeneous DLCM     & niave        & 2,744 & 5,419 & 8,100 & 10,798 & 13,482 & 26,910 \\
Traditional Data   & Homogeneous DLCM     & Bucket   & 2,738 & 5,410 & 8,091 & 10,788 & 13,472 & 26,900 \\
Traditional Data   & Homogeneous DLCM     & Pattern Adjusted  & 2,738 & 5,410 & 8,091 & 10,787 & 13,472 & 26,900 \\
Traditional Data   & Heterogeneous DLCM   & niave        & 2,742 & 5,419 & 8,104 & 10,801 & 13,487 & 26,917 \\
Traditional Data   & Heterogeneous DLCM   & Bucket   & 2,738 & 5,410 & 8,091 & 10,788 & 13,472 & 26,900 \\
Traditional Data   & Heterogeneous DLCM   & Pattern Adjusted  & 2,738 & 5,410 & 8,091 & 10,787 & 13,472 & 26,900 \\
\\[-3ex]
Homogeneous Data   & Traditional LCM & & 3,009 & 5,987 & 8,933 & 11,908 & 14,868 & 29,665 \\
Homogeneous Data   & Homogeneous DLCM     & niave        & 2,772 & 5,506 & 8,211 & 10,936 & 13,645 & 27,238 \\
Homogeneous Data   & Homogeneous DLCM     & Bucket   & 2,768 & 5,500 & 8,208 & 10,930 & 13,647 & 27,247 \\
Homogeneous Data   & Homogeneous DLCM     & Pattern Adjusted  & 2,795 & 5,501 & 8,217 & 10,933 & 13,640 & 27,244 \\
Homogeneous Data   & Heterogeneous DLCM   & niave        & 2,776 & 5,511 & 8,215 & 10,935 & 13,647 & 27,236 \\
Homogeneous Data   & Heterogeneous DLCM   & Bucket   & 2,782 & 5,505 & 8,211 & 10,931 & 13,647 & 27,238 \\
Homogeneous Data   & Heterogeneous DLCM   & Pattern Adjusted  & 2,806 & 5,515 & 8,226 & 10,940 & 13,637 & 27,244 \\
\\[-3ex]
Heterogeneous Data & Traditional LCM & & 2,871 & 5,701 & 8,537 & 11,350 & 14,180 & 28,339 \\
Heterogeneous Data & Homogeneous DLCM     & niave        & 2,723 & 5,382 & 8,015 & 10,632 & 13,274 & 26,481 \\
Heterogeneous Data & Homogeneous DLCM     & Bucket   & 2,727 & 5,399 & 8,023 & 10,627 & 13,269 & 26,476 \\
Heterogeneous Data & Homogeneous DLCM     & Pattern Adjusted  & 2,777 & 5,413 & 8,098 & 10,754 & 13,427 & 26,795 \\
Heterogeneous Data & Heterogeneous DLCM   & niave        & 2,685 & 5,334 & 7,983 & 10,606 & 13,247 & 26,455 \\
Heterogeneous Data & Heterogeneous DLCM   & Bucket   & 2,690 & 5,325 & 7,972 & 10,595 & 13,236 & 26,444 \\
Heterogeneous Data & Heterogeneous DLCM   & Pattern Adjusted  & 2,741 & 5,324 & 7,972 & 10,594 & 13,236 & 26,444
\end{tabular}
\caption{From simulation studies with $C=2$ classes. WAIC is averaged across simulations.}
\label{table:simWAICn}
\end{table}

\begin{table}[!htbp]
\begin{tabular}{lllrrrrr}
\multicolumn{3}{l}{Times Chosen as min WAIC}           & \multicolumn{5}{l}{\makecell[l]{\\Number of Classes}}            \\
Data               & Model                & Prior        & C=1 & 2 & 3 & 4 & 5 \\\hline
Traditional Data   & Traditional LCM & & 0  & 33 & 0  & 0  & 0  \\
Traditional Data   & Homogeneous DLCM     & Bucket   & 0  & 11 & 0  & 0  & 0  \\
Traditional Data   & Homogeneous DLCM     & Pattern Adjusted  & 0  & 25 & 0  & 0  & 0  \\
Traditional Data   & Heterogeneous DLCM   & Bucket   & 0  & 5  & 0  & 0  & 0  \\
Traditional Data   & Heterogeneous DLCM   & Pattern Adjusted  & 0  & 26 & 0  & 0  & 0  \\
\multicolumn{3}{r}{Subtotal:}                  & \multicolumn{5}{c}{100 simulated datasets}            \\
Homogeneous Data   & Traditional LCM & & 0  & 0  & 0  & 0  & 0  \\
Homogeneous Data   & Homogeneous DLCM     & Bucket   & 0  & 29 & 1  & 0  & 0  \\
Homogeneous Data   & Homogeneous DLCM     & Pattern Adjusted  & 0  & 36 & 0  & 0  & 0  \\
Homogeneous Data   & Heterogeneous DLCM   & Bucket   & 0  & 15 & 0  & 0  & 0  \\
Homogeneous Data   & Heterogeneous DLCM   & Pattern Adjusted  & 0  & 19 & 0  & 0  & 0  \\
\multicolumn{3}{r}{Subtotal:}                  & \multicolumn{5}{c}{100 simulated datasets}            \\
Heterogeneous Data & Traditional LCM & & 0  & 0  & 0  & 0  & 0  \\
Heterogeneous Data & Homogeneous DLCM     & Bucket   & 0  & 0  & 0  & 0  & 0  \\
Heterogeneous Data & Homogeneous DLCM     & Pattern Adjusted  & 0  & 0  & 0  & 0  & 0  \\
Heterogeneous Data & Heterogeneous DLCM   & Bucket   & 0  & 30 & 1  & 0  & 0  \\
Heterogeneous Data & Heterogeneous DLCM   & Pattern Adjusted  & 0  & 69 & 0  & 0  & 0 \\
\multicolumn{3}{r}{Subtotal:}                  & \multicolumn{5}{c}{100 simulated datasets}            \\
\end{tabular}
\caption{From simulation studies with sample size$n=1000$. In $100$ simulated datasets, how many times was this model chosen as having the best WAIC?}
\label{table:simsWaicBest1}
\end{table}

\begin{table}[!htbp]
\begin{tabular}{lllrrrrrr}
\multicolumn{3}{l}{Times Chosen as min WAIC}           & \multicolumn{6}{l}{\makecell[l]{\\Sample Size}}            \\
Data               & Model              & Prior       & n=100 & 200 & 300 & 400 & 500 & 1000 \\\hline
Traditional Data   & Traditional LCM    &             & 36  & 31  & 32  & 32  & 41  & 33   \\
Traditional Data   & Homogeneous DLCM   & Bucket  & 6   & 7   & 9   & 6   & 11  & 11   \\
Traditional Data   & Homogeneous DLCM   & Pattern Adjusted & 29  & 33  & 34  & 39  & 26  & 25   \\
Traditional Data   & Heterogeneous DLCM & Bucket  & 11  & 15  & 7   & 9   & 6   & 5    \\
Traditional Data   & Heterogeneous DLCM & Pattern Adjusted & 18  & 14  & 18  & 14  & 16  & 26   \\
\multicolumn{3}{r}{Column Subtotal:}                  & \multicolumn{6}{c}{100 simulated datasets each}            \\
Homogeneous Data   & Traditional LCM    &             & 0   & 0   & 0   & 0   & 0   & 0    \\
Homogeneous Data   & Homogeneous DLCM   & Bucket  & 84  & 20  & 30  & 23  & 20  & 29   \\
Homogeneous Data   & Homogeneous DLCM   & Pattern Adjusted & 10  & 78  & 59  & 51  & 49  & 36   \\
Homogeneous Data   & Heterogeneous DLCM & Bucket  & 6   & 1   & 9   & 16  & 14  & 16   \\
Homogeneous Data   & Heterogeneous DLCM & Pattern Adjusted & 0   & 1   & 2   & 10  & 17  & 19   \\
\multicolumn{3}{r}{Column Subtotal:}       & \multicolumn{6}{c}{100 simulated datasets each}            \\
Heterogeneous Data & Traditional LCM    &             & 0   & 0   & 0   & 0   & 0   & 0    \\
Heterogeneous Data & Homogeneous DLCM   & Bucket  & 5   & 1   & 0   & 0   & 0   & 0    \\
Heterogeneous Data & Homogeneous DLCM   & Pattern Adjusted & 0   & 0   & 0   & 0   & 0   & 0    \\
Heterogeneous Data & Heterogeneous DLCM & Bucket  & 95  & 25  & 20  & 22  & 33  & 30   \\
Heterogeneous Data & Heterogeneous DLCM & Pattern Adjusted & 0   & 74  & 80  & 78  & 67  & 70   \\
\multicolumn{3}{r}{Column Subtotal:}          & \multicolumn{6}{c}{100 simulated datasets each}           
\end{tabular}
\caption{From simulation studies with $C=2$ classes. In $100$ simulated datasets, how many times was this model chosen as having the best WAIC?}
\label{table:simsWaicBest2}
\end{table}

\FloatBarrier
\subsection{Traditional Data Simulation}\label{appendix:simstrad}

For the simulations using Traditional Data we generated observations using $J=24$ items and $C=2$ equally sized classes ($\pi_{c}=0.5$). All items are conditionally independent. The true response probabilities can be found in Table~\ref{table:simtraditional-dgp}.

\begin{table}[!htbp]
\begin{tabular}{lrr}
  & \multicolumn{2}{c}{$P(X_{ij}=1|c_{i})$}     \\
Items  &  Class 0 & Class 1 \\
\hline
0,\ 6, 12, 18  & 0.2             & 0.8             \\
1,\ 7, 13, 19  & 0.8             & 0.2             \\
2,\ 8, 14, 20  & 0.2             & 0.5             \\
3,\ 9, 15, 21  & 0.8             & 0.8             \\
4, 10, 16, 22 & 0.2             & 0.2             \\
5, 11, 17, 23 & 0.8             & 0.5           
\end{tabular}
\caption{Traditional Data Simulation. True underlying conditional probabilities.}
\label{table:simtraditional-dgp}
\end{table}

\FloatBarrier
\subsection{Homogeneous Data Simulation}\label{appendix:simshomo}

For the simulations using Homogeneous Data we generated observations using $J=24$ items and $C=2$ equally sized classes ($\pi_{c}=0.5$). Both classes have the domains $\{Q0, Q1, Q2\}$, $\{Q3, Q4\}$ and $\{Q5, Q6\}$. The true response probabilities can be found in Tables \ref{table:simHomo-dgp} and \ref{table:simHomoDomain}.

\begin{table}[!htbp]
\begin{tabular}{lrr}
  & \multicolumn{2}{c}{$P(X_{ij}=1|c_{i})$}     \\
Items  &  Class 0 & Class 1 \\
\hline
Q1, Q2, Q3 & 0.64 & 0.36\\
Q3, Q4 ,Q5, Q6 & 0.65 & 0.35 \\
Q7, Q13, Q19 & 0.8 & 0.2 \\
Q8, Q14, Q20 & 0.2 & 0.8 \\
Q9, Q15, Q21 & 0.8 & 0.5 \\
Q10, Q16, Q22 & 0.2 & 0.2 \\
Q11, Q17, Q23 & 0.8 & 0.8 \\
Q12, Q18, Q24 & 0.2 & 0.5
\end{tabular}
\caption{True marginal probabilities for each item}\label{table:simHomo-dgp}
\end{table}

\begin{table}[!hbtp]
\begin{tabular}{lcccrr}
\multicolumn{6}{l}{Pattern Probabilities} \\
\multicolumn{6}{l}{Domain $\{Q0, Q1, Q2\}$} \\
 & Q0 & Q1 & Q2 & Class 0 & Class 1 \\\hline
 & 0 & 0 & 0 & 0.30 & 0.40 \\
 & 1 & 0 & 0 & 0.30 & 0.03 \\
 & 0 & 1 & 0 & 0.02 & 0.03 \\
 & 1 & 1 & 0 & 0.02 & 0.03 \\
 & 0 & 0 & 1 & 0.02 & 0.03 \\
 & 1 & 0 & 1 & 0.02 & 0.03 \\
 & 0 & 1 & 1 & 0.02 & 0.03 \\
 & 1 & 1 & 1 & 0.30 & 0.40 \\
\multicolumn{6}{l}{Domain   $\{Q3, Q4\}$} \\
 & Q3 & Q4 &  &  &  \\
 & 0 & 0 &  & 0.30 & 0.60 \\
 & 1 & 0 &  & 0.05 & 0.05 \\
 & 0 & 1 &  & 0.05 & 0.05 \\
 & 1 & 1 &  & 0.60 & 0.30 \\
\multicolumn{6}{l}{Domain   $\{Q5, Q6\}$} \\
 & Q5 & Q6 &  &  &  \\
 & 0 & 0 &  & 0.30 & 0.60 \\
 & 1 & 0 &  & 0.05 & 0.05 \\
 & 0 & 1 &  & 0.05 & 0.05 \\
 & 1 & 1 &  & 0.60 & 0.30
\end{tabular}
\caption{True conditional probabilities for domains $\{Q0, Q1, Q2\}$, $\{Q3, Q4\}$, and $\{Q5, Q6\}$.}
\label{table:simHomoDomain}
\end{table}

\FloatBarrier
\subsection{Heterogeneous Data Simulation}\label{appendix:simshet}

\begin{figure}[!htbp]
\includegraphics[width=14cm]{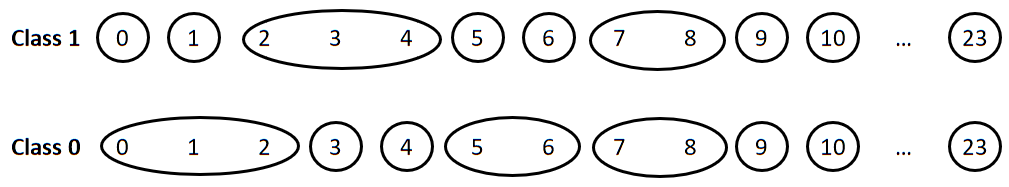}
\caption{Heterogeneous Data Simulation. True domain structure. Items in the same oval are part of the same domain.}
\label{fig:simHetDomains}
\end{figure}

For the simulations using Heterogeneous Data we generated observations using $J=24$ items and $C=2$ equally sized classes ($\pi_{c}=0.5$). The domains are described in Figure~\ref{fig:simHetDomains}. The response probabilities for each domain and item are given in Tables \ref{table:simHettrue1}, \ref{table:simHettrue2}, \ref{table:simHettrue3}, and \ref{table:simHettrue4}.

\begin{table}[!htbp]
\begin{tabular}{lrr}
  & \multicolumn{2}{c}{$P(X_{ij}=1|c_{i})$}     \\
Items  &  Class 0 & Class 1 \\\hline
0    & 0.36    & 0.80    \\
1    & 0.36    & 0.20    \\
2    & 0.36    & 0.36    \\
3    & 0.80    & 0.36    \\
4    & 0.20    & 0.36    \\
5    & 0.65    & 0.30    \\
6    & 0.65    & 0.30    \\
7    & 0.35    & 0.65    \\
8    & 0.35    & 0.65    \\
9    & 0.50    & 0.20    \\
10   & 0.20    & 0.80    \\
11   & 0.80    & 0.20    \\
12   & 0.50    & 0.80    \\
13   & 0.20    & 0.20    \\
14   & 0.80    & 0.80    \\
15   & 0.50    & 0.20    \\
16   & 0.20    & 0.80    \\
17   & 0.80    & 0.20    \\
18   & 0.50    & 0.80    \\
19   & 0.20    & 0.20    \\
20   & 0.80    & 0.80    \\
21   & 0.50    & 0.20    \\
22   & 0.20    & 0.80    \\
23   & 0.80    & 0.20   
\end{tabular}
\caption{Heterogeneous Simulation. True marginal probabilities for each item. }
\label{table:simHettrue1}
\end{table}

\begin{table}[!htbp]
\begin{tabular}{cccr}
\multicolumn{4}{l}{\makecell{
Domains:\\
Class 0 $\{Q0,Q1,Q2\}$\\
Class 1 $\{Q2,Q3,Q4\}$
}} \\
\makecell{1st Question\\ Value of $J_{(1)}(c,d)$} & \makecell{2nd Question\\ Value of $J_{(2)}(c,d)$} & \makecell{3rd Question\\ Value of $J_{(3)}(c,d)$} & Probability \\\hline
0 & 0 & 0 & 0.02 \\
1 & 0 & 0 & 0.30 \\
0 & 1 & 0 & 0.30 \\
0 & 0 & 1 & 0.30 \\
1 & 1 & 0 & 0.02 \\
1 & 0 & 1 & 0.02 \\
0 & 1 & 1 & 0.02 \\
1 & 1 & 1 & 0.02 \\
\multicolumn{3}{l}{Total:} & 1.00
\end{tabular}
\caption{Heterogeneous Simulation. True conditional probabilities for domains $J(c=0,d_{0})=\{Q0,Q1,Q2\}$ and $J(c=1,d_{1})=\{Q2,Q3,Q4\}$}
\label{table:simHettrue2}
\end{table}

\begin{table}[!htbp]
\begin{tabular}{ccr}
Q5 & Q6 & Class 0  \\\hline
0  & 0  & 0.30    \\
1  & 0  & 0.05    \\
0  & 1  & 0.05    \\
1  & 1  & 0.60    
\end{tabular}
\caption{Heterogeneous Simulation. True conditional probabilities for domain $\{Q5,Q6\}$}
\label{table:simHettrue3}
\end{table}

\begin{table}[!htbp]
\begin{tabular}{ccrr}
Q7 & Q8 & Class 0 & Class 1 \\\hline
0  & 0  & 0.60    & 0.30    \\
1  & 0  & 0.05    & 0.05    \\
0  & 1  & 0.05    & 0.05    \\
1  & 1  & 0.30    & 0.60   
\end{tabular}
\caption{Heterogeneous Simulation. True conditional probabilities for domain $\{Q7,Q8\}$}
\label{table:simHettrue4}
\end{table}

\FloatBarrier

\FloatBarrier
\subsection{Simulations under Bad Starting Conditions}\label{appendix:simsBadDomains}

In this subsection we deliberately seed simulations poorly to show that DLCMs can recover true domain structures even after reaching a poor state. 

\begin{figure}[!htbp]
\includegraphics[width=14cm]{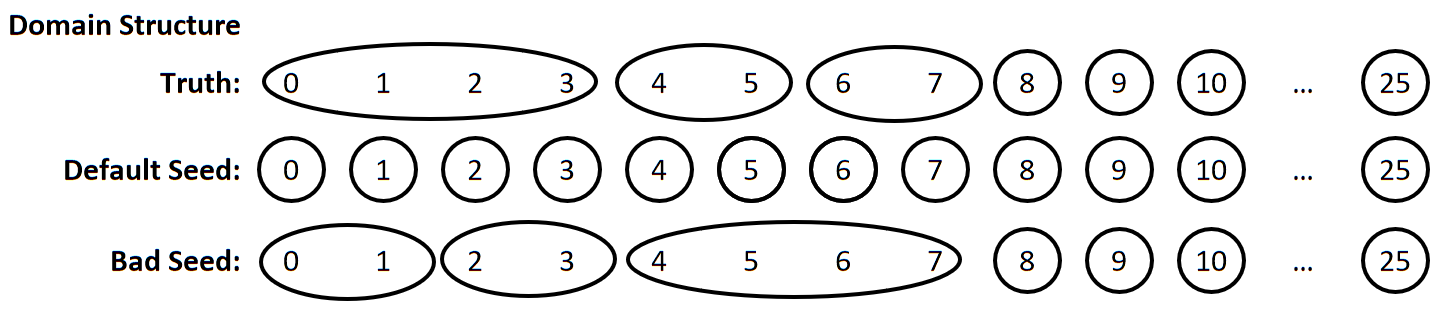}
\caption{Second Homogeneous Data Simulation. Items in the same oval are part of the same domain.}
\label{fig:simHomoBadDomains}
\end{figure}

We build a homogeneous dataset using $C=2$ classes and the domain structure given in Figure~\ref{fig:simHomoBadDomains}. Additional details can be found in Online Appendix~\ref{appendix:sims}. When moving to a correct domain structure there are two basic operations which need to be applied. First domains of related items need to be merged. Second domains containing unrelated items need to be split. We seed in a way that requires both of these operations (Figure~\ref{fig:simHomoBadDomains}). We also seed class membership using independent Bernouli variables. This introduces spurious local dependence caused by mixing the two true classes. This is intended to create a \quote{bad seed} which would serve as poor starting conditions.

In Table~\ref{table:badDomainsModeAccuracy}, we see that both the bad seed and default seeds perform similarly. Compared to Table~\ref{table:simModeAccuracy}, larger sample sizes are needed to recover the domain structure, especially for pattern adjusted priors. This is due to the 4-item domain used here. There is a strong bias against 4-item domains especially for the pattern corrected prior.

\begin{table}[]
\centerline{
\begin{tabular}{llllllllllll}
\multicolumn{3}{l}{Domain Structure Mode Accuracy}                                      & \multicolumn{6}{l}{\makecell[l]{\\Sample Size}}               \\
Data            & Model         & Domain      & Seed    & n=100 & 200   & 300  & 400   & 500   & 1000  & 2000  & 3000  \\\hline
Homogeneous \#2 & Homogeneous   & Bucket  & Default & 0\%   & 0\%   & 7\%  & 21\%  & 39\%  & 96\%  & 98\%  & 100\% \\
Homogeneous \#2 & Homogeneous   & Bucket  & Bad     & 0\%   & 0\%   & 7\%  & 22\%  & 40\%  & 97\%  & 98\%  & 100\% \\
Homogeneous \#2 & Homogeneous   & Pattern Correcting & Default & 0\%   & 0\%   & 0\%  & 0\%   & 0\%   & 0\%   & 65\%  & 100\% \\
Homogeneous \#2 & Homogeneous   & Pattern Correcting & Bad     & 0\%   & 0\%   & 0\%  & 0\%   & 0\%   & 0\%   & 63\%  & 100\% \\
\end{tabular}
}
\caption{Across 100 generated datasets, in what percent did the most common posterior domain structure match the truth?}
\label{table:badDomainsModeAccuracy}
\end{table}

\section{Real World Applications}

\FloatBarrier
\subsection{Marketing Example}\label{appendix:marketing}

We illustrate the DLCM procedure in a real world application.

\begin{example}\label{ex:marketing}

\subheader{Study} Five advertisements were sent to $n=2,240$ customers \citep{ParrRud2014}. For every customer, it was recorded whether they accepted (1) or ignored (0) the offer given in each of the five advertisements (\quote{ads}). The goal is to break these customers into groups (classes) and describe those groups.

The DLCM uses Markov chain Monte-Carlo simulations to approximate the parameter posterior distribution. In this example we simulated four parallel chains of 12,000 iterations each. In each iteration, the DLCM samples new parameter values corresponding to which customers belong to which class, which advertisements are grouped together into domains, and related probabilities.

A homogeneous DLCM fit well with two classes. There was one class (representing 94\% of customers) which was generally unresponsive. Among unresponsive customers, 84\% did not respond to any of the five advertisements. The second class (7\%) was highly responsive. Highly responsive customers responded to two advertisements on average.

We discard the first $2,000$ warmup iterations. Post-warmup, the majority of iterations ($>99\%$) chose to group the items into the same domains. Those domains are: \{Ad1, Ad5\}, \{Ad3, Ad4\}, \{Ad2\}. For the highly responsive group, advertisements 3 and 4 show strong conditional dependence. If you responded to ad 3 you generally did not respond to ad 4, and if you responded to ad 4 you generally did not respond to ad 3. Although we are not provided details on these advertisements, we might speculate for illustrative purposes. Perhaps these ads are two limited-time coupons which cannot both be applied to the same purchase. There is also some local dependence for ads 1 and 5. If you responded to one of the ads, you are more likely to have responded to the other. For more details see Table~\ref{table:illustrativeExample}.

\begin{table}[!htbp]
\textbf{Response Probabilities:}\\
\begin{tabular}{lrrrr}
\multicolumn{5}{l}{\textbf{Class 1: Unresponsive} (94\%)} \\
\multicolumn{1}{l}{\textbf{Ad 1 \textbackslash Ad 5}} & \textbf{1} & \textbf{0} & \textbf{Total} \\ \cline{1-4}
\multicolumn{1}{l}{\textbf{1}} & 1\% & 3\% & 4\% \\
\multicolumn{1}{l}{\textbf{0}} & 3\% & 93\% & 96\% \\
\multicolumn{1}{l}{\textbf{Total}} & 4\% & 96\% & 100\% \\
\end{tabular}
\hspace{0.5cm}
\begin{tabular}{lrrrr}
\multicolumn{5}{l}{\textbf{Class 1:   Highly Responsive} (6\%)} \\
\multicolumn{1}{l}{\textbf{Ad 1 \textbackslash Ad 5}} & \textbf{1} & \textbf{0} & \textbf{Total} \\ \cline{1-4}
\multicolumn{1}{l}{\textbf{1}} & 32\% & 13\% & 46\% \\
\multicolumn{1}{l}{\textbf{0}} & 23\% & 32\% & 54\% \\
\multicolumn{1}{l}{\textbf{Total}} & 55\% & 45\% & 100\% \\
\end{tabular}
\\\\\\
\begin{tabular}{lrrrr}
\multicolumn{5}{l}{\textbf{Class 0: Unresponsive}} \\
\multicolumn{1}{l}{\textbf{Ad 3 \textbackslash Ad 4}} & \textbf{1} & \textbf{0} & \textbf{Total} \\\cline{1-4}
\multicolumn{1}{l}{\textbf{1}} & 0\% & 6\% & 6\% \\
\multicolumn{1}{l}{\textbf{0}} & 3\% & 91\% & 94\% \\
\multicolumn{1}{l}{\textbf{Total}} & 3\% & 97\% & 100\% \\
\end{tabular}
\hspace{0.5cm}
\begin{tabular}{lrrrr}
 \multicolumn{5}{l}{\textbf{Class 1: Highly Responsive}} \\
\multicolumn{1}{l}{\textbf{Ad 3 \textbackslash Ad 4}} & \textbf{1} & \textbf{0} & \textbf{Total} \\\cline{1-4}
\multicolumn{1}{l}{\textbf{1}} & 1\% & 24\% & 25\% \\
\multicolumn{1}{l}{\textbf{0}} & 71\% & 4\% & 75\% \\
\multicolumn{1}{l}{\textbf{Total}} & 72\% & 28\% & 100\% \\
\end{tabular}
\\\\\\
\begin{tabular}{lrrrr}
 \multicolumn{5}{l}{\textbf{Class 0: Unresponsive}} \\
\textbf{Ad 2} & \textbf{1} & \textbf{0} & \textbf{Total} \\\cline{1-4}
\textbf{Probability} & $< 1\%$ & $> 99\%$ & 100\%
\end{tabular}
\hspace{0.5cm}
\begin{tabular}{lrrrrlrrrr}
\multicolumn{5}{l}{\textbf{Class 1:   Highly Responsive}} \\
\textbf{Ad 2} & \textbf{1} & \textbf{0} & \textbf{Total} \\\cline{1-4}\cline{6-10}
\textbf{Probability} & 21\% & 79\% & 100\%
\end{tabular}
\caption{Marketing Example. Conditional on class membership, these tables show how likely a customer is to respond to a pair of advertisements. This is the average probability across iterations and chains.}
\label{table:illustrativeExample}
\end{table}

In this way the DLCM searched and corrected for local dependence between items. Grouping items into a single domain is analogous to creating a new item describing each possible response to the grouped items. In other words, the DLCM found that the transformation given in Figure~\ref{fig:illustrativeExampleTransformation} was useful for correcting local dependence. If we fit a heterogeneous DLCM instead, the domains and corresponding transformation would vary from class to class.

\begin{figure}[!h]
\begin{align*}
\begin{blockarray}{llllll}
& \multicolumn{5}{c}{\text{Original Items}}\\
 & \text{Ad1} & \text{Ad2} & \text{Ad3} & \text{Ad4} & \text{Ad5} \\
 \begin{block}{l[lllll]}
\text{Customer1} & 0 & 0 & 0 & 1 & 0 \\
\text{Customer2} & 1 & 0 & 1 & 0 & 1 \\
\text{Customer3} & 0 & 1 & 0 & 1 & 1 \\
\text{Customer4} & 0 & 1 & 0 & 1 & 1 \\
\text{Customer5} & 1 & 1 & 0 & 1 & 0 \\
\end{block} \\
\end{blockarray}
\Rightarrow
\begin{blockarray}{ccc}
\multicolumn{3}{c}{\text{Transformed \quote{Patterns}}}\\
\text{Ad1 \& Ad5} & \text{Ad3 \& Ad4} & \text{Ad2} \\
\begin{block}{[ccc]}
0 & 2 & 0 \\
3 & 1 & 0 \\
2 & 2 & 1 \\
2 & 2 & 1 \\
1 & 2 & 1 \\
\end{block} \\
\end{blockarray}
\end{align*}
\caption{Marketing Example. Selected rows from transformed dataset.}
\label{fig:illustrativeExampleTransformation}
\end{figure}

\end{example}

\begin{figure}[!htbp]
  \includegraphics[width = 5in]{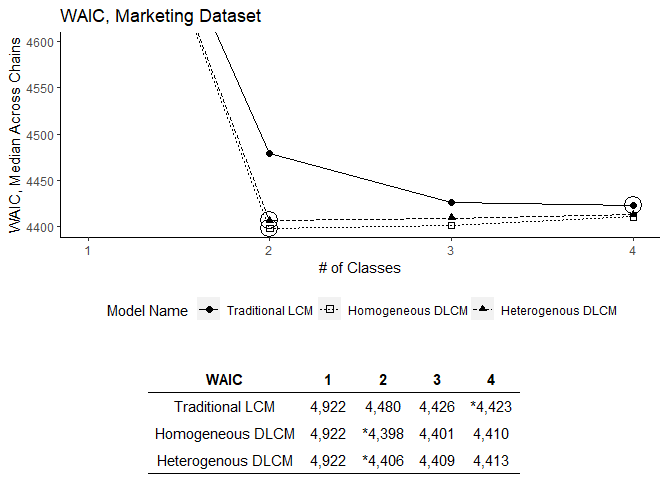}
  \caption{Marketing Example. Median goodness of fit of different models with permissive prior. Asterisks (*) indicate the top model by type.}\label{fig:marketing-waic}
\end{figure}

\begin{table}[!htbp]
\begin{tabular}{lrrrr}
Model Name        & \# of Classes & LPPD   & WAIC Penalty & WAIC   \\\hline
Traditional LCM & 4 & -2,198 & 13 & 4,423 \\
Homogeneous DLCM & 2 & -2,189 & 10 & 4,398\\
Heterogenous DLCM & 2 & -2,191 & 12 & 4,406
\end{tabular}
\caption{Education Application. Best models by type. Bucket prior only.}
\label{table:marketing-gof}
\end{table}

\FloatBarrier

\FloatBarrier
\subsection{Education Application}\label{appendix:education}

We provide some additional information about the education application described in Section~\ref{section:application-education}.

\subsubsection{Education Application: Domains}

Some pairs of questions were not put into the same domain. These are typically easier questions where participants are likely to answer correctly. When probabilities are close to one or zero, local independence can provide an adequate approximation even if items are linked. Suppose we have two perfectly dependent questions $A,B$. In the worst case of $P(A=B=1)=P(A=B=0)=0.5$ the local independence would provide a poor approximation. However if $P(A=B=1)=1-P(A=B=0)=0.99$ then independence may provide an adequate approximation. Under independence with $P(A=1)=P(B=1)=0.99$, we have $P(A=B)=98\%$ with only a $2\%$ `defect' rate. This gives some indication why easier questions were less likely to be grouped.

\begin{figure}[!htbp]
\includegraphics[height=6cm]{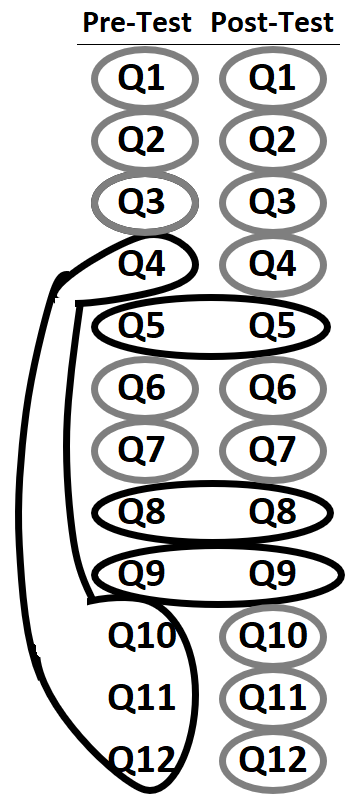}
\caption{Education Application. Visualization of most common homogeneous domain structure}
\label{fig:educationDomain}
\end{figure}

\FloatBarrier
\subsubsection{Education Application: Test Questions}

Test questions from Anselmi, et al. (2010).
\begin{enumerate}
\item[b101:] A box contains 30 marbles in the following colors: 8 red, 10 black, 12 yellow. What is the probability that a randomly drawn marble is yellow? (Correct: 0.40)
\item[b102:] A bag contains 5-cent, 10-cent, and 20-cent coins. The probability of drawing a 5-cent coin is 0.35, that of drawing a 10-cent coin is 0.25, and that of drawing a 20-cent coin is 0.40. What is the probability that the coin randomly drawn is not a 5-cent coin? (0.65)
\item[b103:] A bag contains 5-cent, 10-cent, and 20-cent coins. The probability of drawing a 5-cent coin is 0.20, that of drawing a 10-cent coin is 0.45, and that of drawing a 20-cent coin is 0.35. What is the probability that the coin randomly drawn is a 5-cent coin or a 20-cent coin? (0.55)
\item[b104:] In a school, 40\% of the pupils are boys and 80\% of the pupils are right-handed. Suppose that gender and handedness are independent. What is the probability of randomly selecting a right-handed boy? (0.32)
\item[b105:] Given a standard deck containing 32 different cards, what is the probability of not drawing a heart? (0.75)
\item[b106:] A box contains 20 marbles in the following colors: 4 white, 14 green, 2 red. What is the probability that a randomly drawn marble is not white? (0.80)
\item[b107:] A box contains 10 marbles in the following colors: 2 yellow, 5 blue, 3 red. What is the probability that a randomly drawn marble is yellow or blue? (0.70)
\item[b108:] What is the probability of obtaining an even number by throwing a dice? (0.50)
\item[b109:] Given a standard deck containing 32 different cards, what is the probability of drawing a 4 in a black suit? (Responses that round to 0.06 were considered correct.)
\item[b110:] A box contains marbles that are red or yellow, small or large. The probability of drawing a red marble is 0.70 (lab: 0.30), the probability of drawing a small marble is 0.40. Suppose that the color of the marbles is independent of their size. What is the probability of randomly drawing a small marble that is not red? (0.12, lab: 0.28)
\item[b111:] In a garage there are 50 cars. 20 are black and 10 are diesel powered. Suppose that the color of the cars is independent of the kind of fuel. What is the probability that a randomly selected car is not black and it is diesel powered? (0.12)
\item[b112:] A box contains 20 marbles. 10 marbles are red, 6 are yellow and 4 are black. 12 marbles are small and 8 are large. Suppose that the color of the marbles is independent of their size. What is the probability of randomly drawing a small marble that is yellow or red? (0.48)
\item[b201:] A box contains 30 marbles in the following colors: 10 red, 14 yellow, 6 green. What is the probability that a randomly drawn marble is green? (0.20)
\item[b202:] A bag contains 5-cent, 10-cent, and 20-cent coins. The probability of drawing a 5-cent coin is 0.25, that of drawing a 10-cent coin is 0.60, and that of drawing a 20-cent coin is 0.15. What is the probability that the coin randomly drawn is not a 5-cent coin? (0.75)
\item[b203:] A bag contains 5-cent, 10-cent, and 20-cent coins. The probability of drawing a 5-cent coin is 0.35, that of drawing a 10-cent coin is 0.20, and that of drawing a 20-cent coin is 0.45. What is the probability that the coin randomly drawn is a 5-cent coin or a 20-cent coin? (0.80)
\item[b204:] In a school, 70\% of the pupils are girls and 10\% of the pupils are left-handed. Suppose that gender and handedness are independent. What is the probability of randomly selecting a left-handed girl? (0.07)
\item[b205:] Given a standard deck containing 32 different cards, what is the probability of not drawing a club? (0.75)
\item[b206:] A box contains 20 marbles in the following colors: 6 yellow, 10 red, 4 green. What is the probability that a randomly drawn marble is not yellow? (0.70)
\item[b207:] A box contains 10 marbles in the following colors: 5 blue, 3 red, 2 green. What is the probability that a randomly drawn marble is blue or red? (0.80)
\item[b208:] What is the probability of obtaining an odd number by throwing a dice? (0.50)
\item[b209:] Given a standard deck containing 32 different cards, what is the probability of drawing a 10 in a red suit? (Responses that round to 0.06 were considered correct.)
\item[b210:] A box contains marbles that are green or red, large or small The probability of drawing a green marble is 0.40, the probability of drawing a large marble is 0.20. Suppose that the color of the marbles is independent of their size. What is the probability of randomly drawing a large marble that is not green? (0.12)
\item[b211:] In a garage there are 50 cars. 15 are white and 20 are diesel powered. Suppose that the color of the cars is independent of the kind of fuel. What is the probability that a randomly selected car is not white and it is diesel powered? (0.28)
\item[b212:] A box contains 20 marbles. 8 marbles are white, 4 are green and 8 are red. 15 marbles are small and 5 are large. Suppose that the color of the marbles is independent of their size. What is the probability of randomly drawing a large marble that is white or green? (0.15)
\end{enumerate}

\FloatBarrier
\subsubsection{Education Application: Goodness of Fit}

\begin{figure}[!htbp]
  \includegraphics[width = 5in]{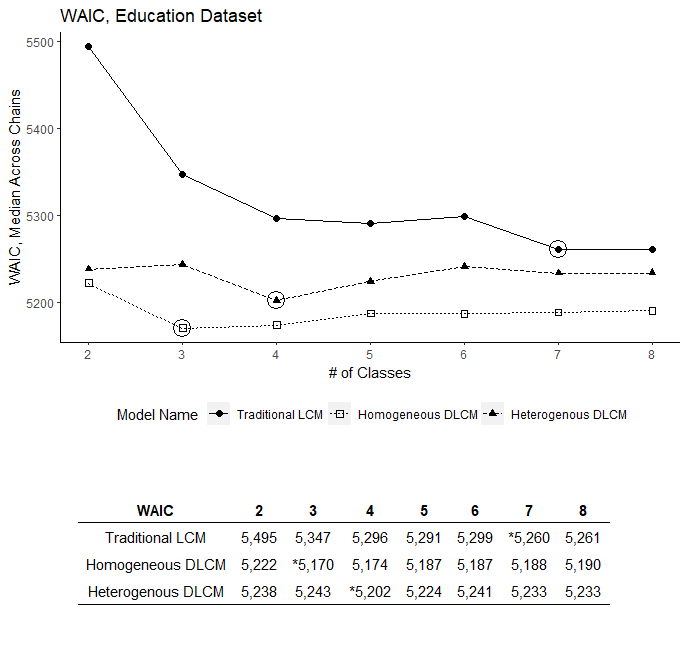}
  \caption{Education Application. Median goodness of fit of different models. Permissive domains priors only. Asterisks (*) indicate the top model by type.}\label{fig:education-waic}
\end{figure}

\begin{table}[!htbp]
\begin{tabular}{llrrrrrrrr}
WAIC              &             &   \multicolumn{8}{l}{\makecell[l]{\\Number of Classes}}       \\
Model             & Prior       & C=1   & 2     & 3      & 4      & 5      & 6     & 7      & 8     \\\hline
Traditional LCM   &             & 6,378 & 5,495 & 5,347  & 5,296  & 5,291  & 5,299 & *5,260 & 5,261 \\
Homogeneous DLCM  & permissive  & 5,527 & 5,222 & *5,170 & 5,174  & 5,187  & 5,187 & 5,188  & 5,190 \\
Homogeneous DLCM  & restrictive & 5,562 & 5,305 & 5,250  & 5,234  & *5,218 & 5,239 & 5,226  & 5,245 \\
Heterogenous DLCM & permissive  & 5,526 & 5,238 & 5,243  & *5,202 & 5,224  & 5,241 & 5,233  & 5,233 \\
Heterogenous DLCM & restrictive & 5,565 & 5,309 & 5,254  & 5,235  & *5,226 & 5,235 & 5,234  & 5,237
\end{tabular}
\caption{Education Application. Median goodness of fit of different models.}\label{fig:education-waic-full}
\end{table}

\FloatBarrier
\subsubsection{Education Application: Response Probabilities}

\begin{table}[!htbp]
\scalebox{1}{
\begin{tabular}{rrrrrrr}
& \multicolumn{2}{l|}{\makecell{Class 0 \\ \quote{Proficient}}}
& \multicolumn{2}{l|}{\makecell{Class 1 \\ \quote{Pre \textgreater Post}}}
& \multicolumn{2}{l}{\makecell{Class 2 \\ \quote{Beginners}}} \\
Question & Pre-Test    & Post-Test & \multicolumn{1}{|r}{Pre}         & Post & \multicolumn{1}{|r}{Pre}         & Post \\\hline
01 & 94\% & 97\% & 74\% & 41\% & 69\% & 91\% \\
02 & 98\% & 99\% & 81\% & 26\% & 89\% & 97\% \\
03 & 96\% & 98\% & 60\% & 10\% & 81\% & 81\% \\
04 & 89\% & 98\% & 39\% & 9\% & 42\% & 45\% \\
05 & 88\% & 90\% & 71\% & 38\% & 62\% & 69\% \\
06 & 97\% & 97\% & 60\% & 25\% & 73\% & 83\% \\
07 & 93\% & 99\% & 76\% & 26\% & 79\% & 90\% \\
08 & 95\% & 96\% & 74\% & 45\% & 85\% & 88\% \\
09 & 79\% & 85\% & 39\% & 17\% & 45\% & 48\% \\
10 & 75\% & 95\% & 35\% & 10\% & 26\% & 28\% \\
11 & 67\% & 84\% & 35\% & 15\% & 27\% & 26\% \\
12 & 78\% & 91\% & 37\% & 17\% & 28\% & 31\% \\
Average: & 87\% & 94\% & 57\% & 23\% & 59\% & 65\% \\
\end{tabular}
}
\caption{Education Application. Marginal probabilities of each item under a homogeneous DLCM.}
\label{table:education-homomarginals}
\end{table}

\begin{figure}[!htbp]
  \includegraphics[width = 5in]{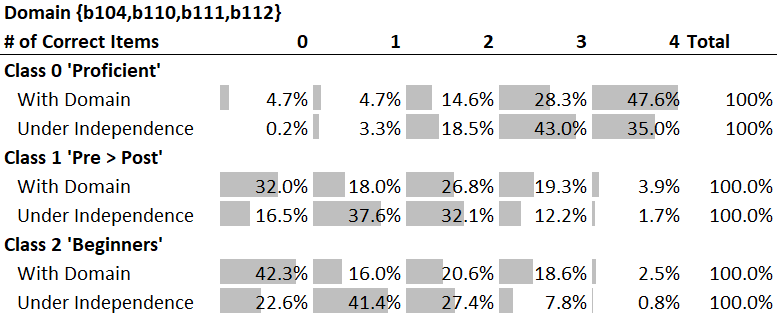}
  \caption{Education application with homogeneous DLCM. Conditionally independent and conditionally dependent probabilities are compared. Independence probabilities are found by marginalizing the probability of each item and taking the product.}
  \label{fig:education-bigdomaintable}
\end{figure}

\begin{table}[!htbp]
\scalebox{1}{
\begin{tabular}{llrrr}
\multicolumn{2}{l}{Conditional   Probability} & \makecell{Class 0\\'Proficient'} & \makecell{Class 1\\'Pre \textgreater Post'} & \makecell{Class 2\\'Beginners'} \\\hline
\multicolumn{2}{l}{Domain   \{B105,B205\}:} &  &  \\
& B205 Correct $\vert$ B105 Correct & 94\% & 43\% & 83\% \\
& B205 Correct $\vert$ B105 Mistake & 61\% & 25\% & 45\% \\
\multicolumn{2}{l}{Domain   \{B108,B208\}:} &  &  \\
& B208 Correct $\vert$ B108 Correct & 99\% & 45\% & 90\% \\
& B208 Correct $\vert$ B108 Mistake & 45\% & 47\% & 75\% \\
\multicolumn{2}{l}{Domain   \{B109,B209\}:} &  &  \\
& B209 Correct $\vert$ B109 Correct & 96\% & 25\% & 69\% \\
& B209 Correct $\vert$ B109 Mistake & 43\% & 12\% & 31\%
\end{tabular}
}
\caption{Education Application. Conditional probability of paired items.}
\label{table:education-homodomainprobs}
\end{table}

\begin{table}[!htbp]
\scalebox{1}{
\begin{tabular}{llllrrr}
\multicolumn{4}{l}{Domain: $\{b104,b110,b111,b112\}$}
& \multicolumn{3}{c}{Pattern Probabilities}
\\
B104 & B110 & B111 & B112 & \makecell{Class 0\\'Proficient'} & \makecell{Class 1\\'Pre \textgreater Post'} & \makecell{Class 2\\'Beginners'} \\\hline
Mistake & Mistake & Mistake & Mistake & 4.7\% & 32.0\% & 42.3\% \\
Correct & Mistake & Mistake & Mistake & 2.8\% & 3.9\% & 9.1\% \\
Mistake & Correct & Mistake & Mistake & 0.3\% & 3.8\% & 2.8\% \\
Correct & Correct & Mistake & Mistake & 4.3\% & 3.8\% & 4.6\% \\
Mistake & Mistake & Correct & Mistake & 0.5\% & 3.8\% & 2.1\% \\
Correct & Mistake & Correct & Mistake & 2.4\% & 3.9\% & 4.1\% \\
Mistake & Correct & Correct & Mistake & 0.8\% & 3.8\% & 2.2\% \\
Correct & Correct & Correct & Mistake & 6.2\% & 7.8\% & 5.2\% \\
Mistake & Mistake & Mistake & Correct & 1.0\% & 6.5\% & 2.0\% \\
Correct & Mistake & Mistake & Correct & 6.3\% & 7.6\% & 4.9\% \\
Mistake & Correct & Mistake & Correct & 0.4\% & 3.8\% & 2.5\% \\
Correct & Correct & Mistake & Correct & 13.4\% & 3.8\% & 5.0\% \\
Mistake & Mistake & Correct & Correct & 0.4\% & 3.9\% & 2.4\% \\
Correct & Mistake & Correct & Correct & 6.5\% & 3.8\% & 6.8\% \\
Mistake & Correct & Correct & Correct & 2.3\% & 3.8\% & 1.6\% \\
Correct & Correct & Correct & Correct & 47.6\% & 3.9\% & 2.5\% \\
\multicolumn{4}{l}{Total} & 100\% & 100\% & 100\%
\end{tabular}
}
\caption{Education Application. Conditional probabilities of four-item domain.}
\label{table:education-homobigdomainprobs}
\end{table}

\FloatBarrier
\subsubsection{Education Application: Class and Treatments}

\begin{table}[!htbp]\label{table:education-treatments}
\begin{tabular}{lll}
Homogeneous DLCM          &       &          \\
Class                     & Basic & Enhanced \\\hline
0 \quote{Proficient}            & 132   & 148      \\
1 \quote{Pre \textgreater Post} & 5    & 5       \\
2 \quote{Beginners} & 30     & 25        \\
\end{tabular}
\caption{Education Application. Treatment by latent class. The mode class of each subject is considered. The \quote{enhanced} treatment provided additional instruction with examples, and the \quote{basic} treatment provided instruction without examples.}
\end{table}

\subsection{Medical Application}\label{appendix:medical}

We provide some additional information about the medical application described in Section~\ref{section:application-medical}.

\FloatBarrier
\subsubsection{Medical Application: Goodness of Fit}

\begin{figure}[!htbp]
  \includegraphics[width = 5in]{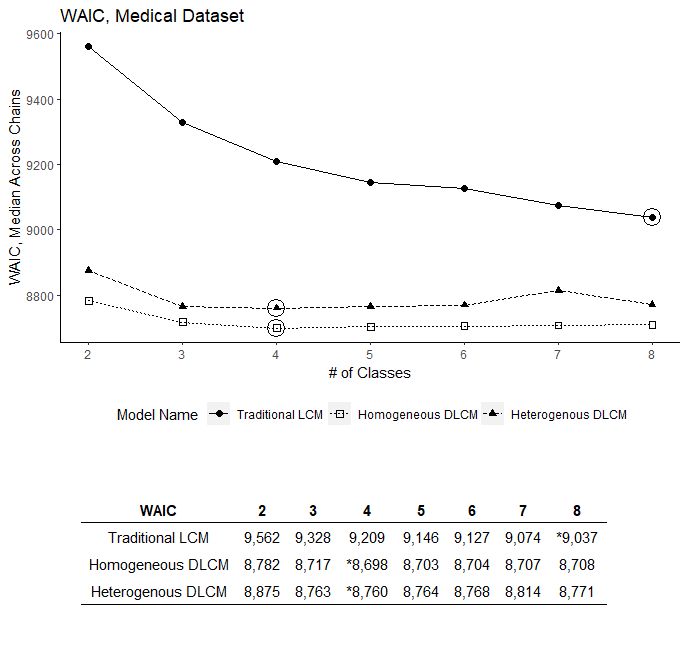}
\caption{Medical Application. Goodness of fit for evaluated models. Bucket domain prior only. Asterisks (*) indicate the top model by type.}\label{fig:medical-waic}
\end{figure}

Figure~\ref{fig:medical-waic} shows the goodness of fit under different models and number of classes.

\begin{table}[!htbp]
\begin{tabular}{llrrrrrrrr}
WAIC              &             &   \multicolumn{8}{l}{\makecell[l]{\\Number of Classes}}       \\
Model             & Prior       & C=1   & 2     & 3      & 4      & 5      & 6     & 7      & 8     \\\hline
Traditional LCM &  & 10,321 & 9,562 & 9,328 & 9,209 & 9,146 & 9,127 & 9,074 & * 9,037 \\
Homogeneous DLCM & permissive & 8,887 & 8,782 & 8,717 & *8,698 & 8,703 & 8,704 & 8,707 & 8,708 \\
Homogeneous DLCM & restrictive & 8,999 & 8,823 & 8,731 & *8,723 & 8,761 & 8,799 & 8,842 & 8,848 \\
Heterogenous DLCM & permissive & 8,887 & 8,875 & 8,763 & *8,760 & 8,764 & 8,768 & 8,814 & 8,771 \\
Heterogenous DLCM & restrictive & 9,000 & 8,918 & *8,789 & 8,796 & 8,817 & 8,823 & 8,794 & 8,831 \\
\end{tabular}
\caption{Medical Application. Median goodness of fit of different models.}\label{fig:medical-waic-full}
\end{table}

\FloatBarrier
\subsubsection{Medical Application: Domains}
\begin{figure}
\includegraphics[width=8cm]{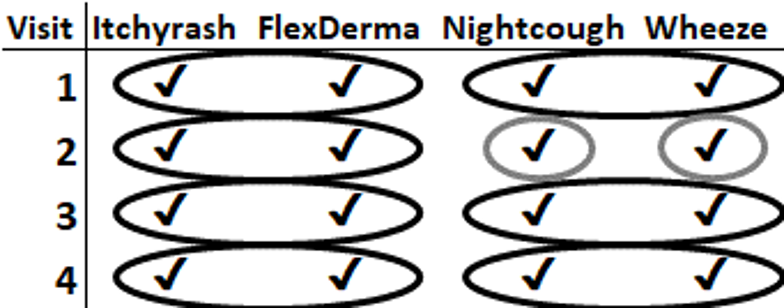}
\caption{Medical Application. Visualization of most common domain structure.}
\label{fig:medicalDomain}
\end{figure}

The most common domains structure is visualized in Figure~\ref{fig:medicalDomain}.

\FloatBarrier
\subsubsection{Medical Application: Response Probabilities}
The tables and figures in this section describe the fitted homogeneous response probabilities for the medical dataset.

\begin{table}[!htbp]
\begin{tabular}{llrrrr}
\multicolumn{2}{l}{\makecell{Symptom\\Time}}           & \makecell{Class 0\\\quote{Good All}} & \makecell{Class 1\\\quote{Bad All}} & \makecell{Class 2\\\quote{Bad Lungs}} & \makecell{Class 3\\\quote{Bad Skin}} \\\hline
\multicolumn{2}{l}{FlexDerma}  &                    &                  &                   &                   \\
 & 1 (Time) & 15\% & 51\% & 15\% & 76\% \\
 & 2 & 14\% & 58\% & 11\% & 66\% \\
 & 3 & 10\% & 43\% & 4\% & 44\% \\
 & 4 & 7\% & 41\% & 6\% & 31\% \\
\multicolumn{2}{l}{Itchyrash}  &                    &                  &                   &                   \\
 & 1 & 24\% & 76\% & 23\% & 83\% \\
 & 2 & 18\% & 73\% & 25\% & 83\% \\
 & 3 & 14\% & 71\% & 5\% & 64\% \\
 & 4 & 14\% & 61\% & 7\% & 59\% \\
\multicolumn{2}{l}{Nightcough} &                    &                  &                   &                   \\
 & 1 & 35\% & 74\% & 76\% & 50\% \\
 & 2 & 39\% & 87\% & 90\% & 48\% \\
 & 3 & 31\% & 65\% & 59\% & 42\% \\
 & 4 & 18\% & 63\% & 54\% & 48\% \\
\multicolumn{2}{l}{Wheeze}     &                    &                  &                   &                   \\
 & 1 & 25\% & 66\% & 53\% & 30\% \\
 & 2 & 13\% & 83\% & 62\% & 15\% \\
 & 3 & 9\% & 54\% & 37\% & 17\% \\
 & 4 & 2\% & 46\% & 28\% & 22\% \\
\end{tabular}
\caption{Medical Application. Marginal probabilities for homogeneous DLCM}
\label{table:medical-classes}
\end{table}

{
\spacingset{1.1}
\begin{table}[!htbp]
\begin{tabular}{llrrrrr}
\multicolumn{2}{l}{Question 1}                 & Yes  & Yes  & No   & No   & \multirow{2}{*}{Total} \\
\multicolumn{2}{l}{Question 2} & Yes & No & Yes & No &  \\\hline
\multicolumn{2}{l}{Nightcough.1; Wheeze.1} &  &  &  &  &  \\
 & Class 0: Good All & 14\% & 21\% & 11\% & 54\% & 100\% \\
 & Class 1: Bad All & 56\% & 18\% & 10\% & 15\% & 100\% \\
 & Class 2: Bad Lungs & 47\% & 28\% & 6\% & 19\% & 100\% \\
 & Class 3: Bad Skin & 19\% & 31\% & 11\% & 39\% & 100\% \\
\multicolumn{2}{l}{Nightcough.3; Wheeze.3} &  &  &  &  &  \\
 & Class 0: Good All & 8\% & 23\% & 2\% & 67\% & 100\% \\
 & Class 1: Bad All & 46\% & 19\% & 8\% & 27\% & 100\% \\
 & Class 2: Bad Lungs & 26\% & 33\% & 11\% & 30\% & 100\% \\
 & Class 3: Bad Skin & 13\% & 29\% & 5\% & 53\% & 100\% \\
\multicolumn{2}{l}{Nightcough.4; Wheeze.4} &  &  &  &  &  \\
 & Class 0: Good All & 2\% & 17\% & 1\% & 81\% & 100\% \\
 & Class 1: Bad All & 35\% & 28\% & 10\% & 27\% & 100\% \\
 & Class 2: Bad Lungs & 24\% & 29\% & 4\% & 43\% & 100\% \\
 & Class 3: Bad Skin & 17\% & 30\% & 5\% & 47\% & 100\% \\
\multicolumn{2}{l}{Itchyrash.1; FlexDerma.1} &  &  &  &  &  \\
 & Class 0: Good All & 13\% & 10\% & 1\% & 75\% & 100\% \\
 & Class 1: Bad All & 48\% & 28\% & 3\% & 21\% & 100\% \\
 & Class 2: Bad Lungs & 12\% & 11\% & 3\% & 74\% & 100\% \\
 & Class 3: Bad Skin & 72\% & 11\% & 4\% & 14\% & 100\% \\
\multicolumn{2}{l}{Itchyrash.2; FlexDerma.2} &  &  &  &  &  \\
 & Class 0: Good All & 11\% & 7\% & 3\% & 78\% & 100\% \\
 & Class 1: Bad All & 57\% & 16\% & 1\% & 26\% & 100\% \\
 & Class 2: Bad Lungs & 10\% & 16\% & 1\% & 73\% & 100\% \\
 & Class 3: Bad Skin & 62\% & 21\% & 4\% & 13\% & 100\% \\
\multicolumn{2}{l}{Itchyrash.3; FlexDerma.3} &  &  &  &  &  \\
 & Class 0: Good All & 7\% & 7\% & 3\% & 83\% & 100\% \\
 & Class 1: Bad All & 41\% & 30\% & 2\% & 27\% & 100\% \\
 & Class 2: Bad Lungs & 2\% & 3\% & 1\% & 94\% & 100\% \\
 & Class 3: Bad Skin & 40\% & 24\% & 4\% & 32\% & 100\% \\
\multicolumn{2}{l}{Itchyrash.4; FlexDerma.4} &  &  &  &  &  \\
 & Class 0: Good All & 5\% & 9\% & 2\% & 84\% & 100\% \\
 & Class 1: Bad All & 36\% & 26\% & 6\% & 33\% & 100\% \\
 & Class 2: Bad Lungs & 2\% & 5\% & 3\% & 89\% & 100\% \\
 & Class 3: Bad Skin & 29\% & 30\% & 1\% & 40\% & 100\% \\
\end{tabular}
\caption{Medical Application. Domain probabilities for homogeneous DLCM.}
\label{table:medical-domains}
\end{table}
}

\FloatBarrier

\subsection{Sociology Application}\label{appendix:sociology}

We provide some additional information about the sociological application described in Section~\ref{section:appSociology}.

\subsubsection{Sociology Application: Survey Questions}

Survey Questions from \citet{cdc2017}.
\begin{itemize}
\item Q19: Have you ever been physically forced to have sexual intercourse when you did not want to?
\item Q20: During the past 12 months, how many times did anyone force you to do sexual things that you did not want to do?
\item Q21: During the past 12 months, how many times did someone you were dating or going out with force you to do sexual things that you did not want to do?
\item Q22: During the past 12 months, how many times did someone you were dating or going out with physically hurt you on purpose?
\item Q59: Have you ever had sexual intercourse?
\item Q60: How old were you when you had sexual intercourse for the first time?
\item Q61: During your life, with how many people have you had sexual intercourse?
\item Q62: During the past 3 months, with how many people did you have sexual intercourse?
\item Q63: Did you drink alcohol or use drugs before you had sexual intercourse the last time?
\item Q64: The last time you had sexual intercourse, did you or your partner use a condom?
\item Q65: The last time you had sexual intercourse, what one method did you or your partner use to prevent pregnancy?
\item Q66: During your life, with whom have you had sexual contact?
\item Q85: Have you ever been tested for HIV, the virus that causes AIDS?
\end{itemize}

\FloatBarrier
\subsubsection{Sociology Application: Goodness of Fit}
\begin{figure}[!htbp]
  \includegraphics[width = 5in]{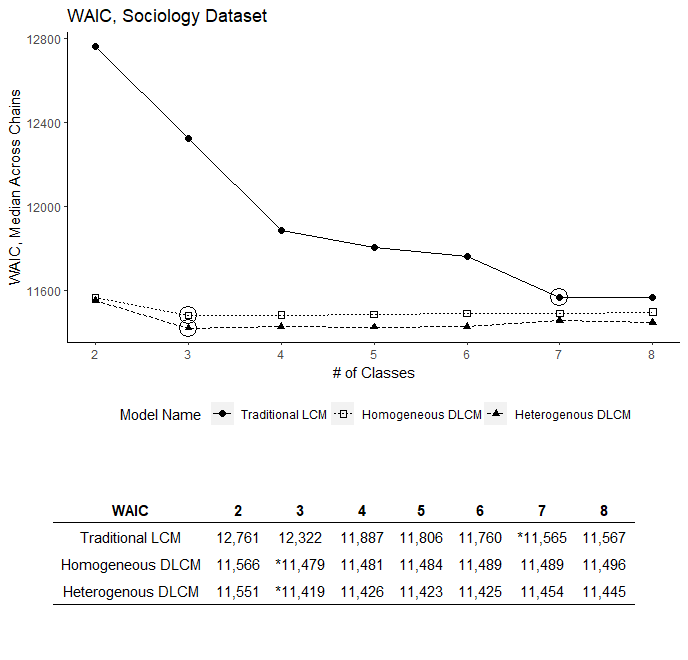}
  \caption{Sociology Application. Goodness of fit of evaluated models. Asterisks (*) indicate the top model by type.}
  \label{fig:sociology-waic}
\end{figure}

Figure~\ref{fig:sociology-waic} shows the goodness of fit under different models and number of classes.

\begin{table}[!htbp]
\begin{tabular}{llrrrrrrrr}
WAIC              &             &   \multicolumn{8}{l}{\makecell[l]{\\Number of Classes}}       \\
Model             & Prior       & C=1   & 2     & 3      & 4      & 5      & 6     & 7      & 8     \\\hline
Traditional LCM &               & 21,170 & 12,761 & 12,322  & 11,887 & 11,806  & 11,760 & *11,565 & 11,567 \\
Homogeneous DLCM & permissive   & 13,712 & 11,566 & *11,479 & 11,481 & 11,484  & 11,489 & 11,489  & 11,496 \\
Homogeneous DLCM & restrictive  & 13,711 & 11,741 & *11,479 & 11,559 & 11,490  & 11,489 & 11,495  & 11,489 \\
Heterogenous DLCM & permissive  & 13,712 & 11,551 & *11,419 & 11,426 & 11,423  & 11,425 & 11,454  & 11,445 \\
Heterogenous DLCM & restrictive & 13,712 & 11,551 & 11,471  & 11,488 & *11,456 & 11,516 & 11,519  & 11,527
\end{tabular}
\caption{Medical Application. Median goodness of fit of different models.}\label{fig:sociology-waic-full}
\end{table}

\FloatBarrier
\subsubsection{Sociology Application: Response Probabilities}
The tables and figures in this section describe the fitted homogeneous response probabilities for the sociology dataset.

{
\spacingset{1.1}
\begin{table}[!htbp]
\begin{tabular}{lllll}
Question & Value & \makecell{Class 0\\Sexually Active} & \makecell{Class 1\\Not Sexually Active} & \makecell{Class 2\\At Risk} \\\hline
Q19 & No & 93.8\% & 96.6\% & 58.3\% \\
Q19 & Yes & 6.2\% & 3.4\% & 41.7\% \\
Q20 & 0 & 87.2\% & 90.4\% & 54.7\% \\
Q20 & 1 & 8.4\% & 5.7\% & 19.8\% \\
Q20 & 2+ & 4.4\% & 3.9\% & 25.6\% \\
Q21 & 0 & 92.3\% & 94.7\% & 70.8\% \\
Q21 & 1 & 5.4\% & 2.6\% & 15.9\% \\
Q21 & 2+ & 2.3\% & 2.7\% & 13.4\% \\
Q22 & 0 & 93.7\% & 98.4\% & 77.9\% \\
Q22 & 1 & 3.6\% & 0.7\% & 8.0\% \\
Q22 & 2+ & 2.7\% & 1.0\% & 14.1\% \\
Q59 & No & 0.3\% & 99.9\% & 0.5\% \\
Q59 & Yes & 99.7\% & 0.1\% & 99.5\% \\
Q60 & Never & 0.3\% & 99.7\% & 0.5\% \\
Q60 & <=13 Years & 1.2\% & 0.1\% & 25.5\% \\
Q60 & 14+ years & 98.5\% & 0.1\% & 74.0\% \\
Q61 & Never & 0.3\% & 99.7\% & 0.5\% \\
Q61 & 1 & 66.3\% & 0.1\% & 13.8\% \\
Q61 & 2+ & 33.4\% & 0.1\% & 85.7\% \\
Q62 & 0 & 22.6\% & 99.7\% & 24.4\% \\
Q62 & 1 & 76.2\% & 0.1\% & 50.2\% \\
Q62 & 2+ & 1.2\% & 0.1\% & 25.5\% \\
Q63 & No & 92.6\% & 99.9\% & 72.6\% \\
Q63 & Yes & 7.4\% & 0.1\% & 27.4\% \\
Q64 & No & 39.7\% & 99.9\% & 60.6\% \\
Q64 & Yes & 60.3\% & 0.1\% & 39.4\% \\
Q65 & Never had sex & 0.6\% & 99.5\% & 0.9\% \\
Q65 & No method & 12.1\% & 0.1\% & 19.7\% \\
Q65 & Birth control pills & 19.8\% & 0.1\% & 18.9\% \\
Q65 & Condoms & 49.0\% & 0.1\% & 26.4\% \\
Q65 & Other & 18.5\% & 0.1\% & 34.0\% \\
Q66 & I have never had sexual contact & 0.3\% & 76.7\% & 0.5\% \\
Q66 & Females & 3.5\% & 1.4\% & 3.5\% \\
Q66 & Males & 91.0\% & 17.7\% & 59.4\% \\
Q66 & Females and males & 5.2\% & 4.2\% & 36.7\% \\
Q85 & No & 84.1\% & 96.6\% & 69.9\% \\
Q85 & Yes & 15.9\% & 3.4\% & 30.1\% \\
\end{tabular}
\caption{Marginal probabilities for sociology dataset}
\label{table:sociologyProbabilities}
\end{table}
}

{
\spacingset{1.5}
\setlength\dashlinedash{2.0pt}
\setlength\dashlinegap{1.5pt}
\setlength\arrayrulewidth{0.3pt}
\begin{table}[!htbp]
\begin{tabular}{ll|lrrr}
\multicolumn{1}{l}{} & & \multicolumn{4}{l}{\makecell{Q21: During the past 12 months,\\ how many times did someone you were {\ul \textbf{dating}} \\ or going out with force you to do sexual things \\ that you did not want to do?}} \\\hline 
\multirow{13}{2cm}{Q20: During the past 12 months, how many times did {\ul \textbf{anyone}} force you to do sexual things\\that you did not want to do?} 
 && \multicolumn{4}{l}{Class 1 \quote{Not Sexually Active}:} \\\cline{3-6}
 & & & 0 & 1 & 2+ \\
&& 0 & 90.1\% & \multicolumn{1}{:r}{0.1\%} & 0.1\% \\\cdashline{5-5}
&& 1 & 3.4\% & 2.2\% & \multicolumn{1}{:r}{0.1\%} \\\cdashline{6-6}
&& 2+ & 1.2\% & 0.3\% & 2.4\%\\\cline{3-6}
 && \multicolumn{4}{l}{Class 0 \quote{Sexually Active}:} \\\cline{3-6}
 & & & 0 & 1 & 2+ \\
&& 0 & 86.6\% & \multicolumn{1}{:r}{0.3\%} & 0.3\%\\\cdashline{5-5}
&& 1 & 4.9\% & 3.3\% & \multicolumn{1}{:r}{0.3\%}\\\cdashline{6-6}
&& 2+ & 0.9\% & 1.8\% & 1.7\%\\\cline{3-6}
 && \multicolumn{3}{l}{Class 2 \quote{At Risk}, conditioned on:} & \\
 && \multicolumn{4}{l}{Q19. Ever physically forced to have sex = No} \\\cline{3-6}
 & & & 0 & 1 & 2+ \\
&& 0 & 72.5\% & \multicolumn{1}{:r}{0.8\%} & 0.8\%\\\cdashline{5-5}
&& 1 & 6.9\% & 8.1\% & \multicolumn{1}{:r}{0.8\%}\\\cdashline{6-6}
&& 2+ & 2.9\% & 0.8\% & 6.4\%\\\cline{3-6}
 && \multicolumn{4}{l}{Class 2 \quote{At Risk}, conditioned on:} \\
 && \multicolumn{4}{l}{Q19. Ever physically forced to have sex = Yes} \\\cline{3-6}
 & & & 0 & 1 & 2+ \\
&& 0  & 25.6\% & \multicolumn{1}{:r}{1.0\%}  & 1.1\%  \\\cdashline{5-5}
&& 1  & 10.4\% & 13.8\% & \multicolumn{1}{:r}{1.0\%}  \\\cdashline{6-6}
&& 2+ & 18.6\% & 9.6\%  & 18.9\%
\end{tabular}
\caption{Sociology Application. Response probabilities for domains \{Q20,Q21\} and \{Q20,Q21,Q19\}.}
\label{table:sociology-Q20Q21}
\end{table}
}

\begin{table}[!htbp]
\begin{tabular}{llrrrr}
& & \multicolumn{4}{c}{\makecell{Q64: The last time you had \\ sexual intercourse, did \\ your partner use a {\ul \textbf{condom}}?}} \\
                    && \multicolumn{2}{l}{\makecell{Class 2\\At Risk}} & \multicolumn{2}{|l}{\makecell{Class 0\\Sexually Active}} \\
                    & & No                & \multicolumn{1}{r|}{Yes}              & No                    & Yes                  \\\hline
\multirow{5}{3.5cm}{Q65: The last time you had sexual intercourse, what {\ul \textbf{one method}} did you or your partner use to prevent pregnancy?}
& $\bullet$ Never had sex       & 0\%               & 0\%              & 0\%                   & 0\%                  \\
& $\bullet$ Condoms             & 0\%               & 26\%             & 0\%                   & 49\%                 \\
& $\bullet$ Birth control pills & 13\%              & 5\%              & 12\%                  & 8\%                  \\
& $\bullet$ No method           & 19\%              & 0\%              & 12\%                  & 0\%                  \\
& $\bullet$ Other               & 27\%              & 7\%              & 15\%                  & 3\%                  \\\cline{2-6}
& Total               & 61\%              & 39\%             & 40\%                  & 60\%                
\end{tabular}
\\
\caption{Sociology Application. Response probabilities for domain \{Q64,Q65\}.}
\label{table:sociology-Q64Q65}
\end{table}

\section{End} 

\end{document}